\documentclass[lettersize,journal]{IEEEtran}
\usepackage{amsmath,amsfonts}
\usepackage{algorithmic}
\usepackage{algorithm}
\usepackage{array}
\usepackage[caption=false,font=normalsize,labelfont=sf,textfont=sf]{subfig}
\usepackage{textcomp}
\usepackage{stfloats}
\usepackage{url}
\usepackage{verbatim}
\usepackage{graphicx}
\usepackage{cite}
\usepackage{xcolor}
\begin{document}

\title{DIDLM: A SLAM Dataset for Difficult Scenarios Featuring Infrared, Depth Cameras, LIDAR, 4D Radar, and Others under Adverse Weather, Low Light Conditions, and Rough Roads}

\author{Weisheng Gong, Chen He,~\IEEEmembership{Member,~IEEE,} Kaijie Su, Qingyong Li,  Tong Wu and Z. Jane Wang,~\IEEEmembership{Fellow,~IEEE}
\thanks{This work was supported in part by the Shaanxi Innovative Team under Grant 2023-CX-TD-04, the Shaanxi Natural Science Foundation under Grant 2025JC-JCQN-088, the National Natural Science Foundation of China under Grant 62271392 and U24B20130, and the Xi’an  Science and Technology Project under Grant 25ZQRC00001. Corresponding author: Chen He (email: chenhe@nwu.edu.cn).

Weisheng Gong, Chen He, Kaijie Su, Qingyong Li and Tong Wu are with the School of Information Science and Technology, Northwest University, Xi'an 710069, China.

Z. Jane Wang is with the Department of Electrical and Computer Engineering, The University of British Columbia, Vancouver, BC V6T 1Z4, Canada.}% <-this % stops a space
\thanks{}}



    \maketitle

\begin{abstract}
Adverse weather conditions, low-light environments, and bumpy road surfaces pose significant challenges to SLAM in robotic navigation and autonomous driving. Existing datasets in this field predominantly rely on single sensors or combinations of LiDAR, cameras, and IMUs. However, 4D millimeter-wave radar demonstrates robustness in adverse weather, infrared cameras excel in capturing details under low-light conditions, and depth images provide richer spatial information. Multi-sensor fusion methods also show potential for better adaptation to bumpy roads. Despite some SLAM studies incorporating these sensors and conditions, there remains a lack of comprehensive datasets addressing low-light environments and bumpy road conditions, or featuring a sufficiently diverse range of sensor data.
In this study, we introduce a multi-sensor dataset covering challenging scenarios such as snowy weather, rainy weather, nighttime conditions, speed bumps, and rough terrains. The dataset includes rarely utilized sensors for extreme conditions, such as 4D millimeter-wave radar, infrared cameras, and depth cameras, alongside 3D LiDAR, RGB cameras, GPS, and IMU. It supports both autonomous driving and ground robot applications and provides reliable GPS/INS ground truth data, covering structured and semi-structured terrains.
We evaluated various SLAM algorithms using this dataset, including RGB images, infrared images, depth images, LiDAR, and 4D millimeter-wave radar. The dataset spans a total of 18.5 km, 69 minutes, and approximately 660 GB, offering a valuable resource for advancing SLAM research under complex and extreme conditions.
Our dataset is available at https://gongweisheng.github.io/DIDLM.github.io/
\end{abstract}

    
\section{Introduction}
\IEEEPARstart{S}{imultaneous} Localization and Mapping (SLAM) is a pivotal technology in both autonomous driving and robotic self-mapping, orchestrating the construction of environmental maps alongside precise robot localization \cite{Pritsker1984}. Its indispensable role in navigation, exploration, and collaborative tasks is self-evident. In uncharted geographic territories, SLAM ensures the accurate positioning of robots, a critical element for search and rescue operations as well as scientific expeditions. The fidelity of the constructed map data is essential for tasks such as environmental monitoring. In environments devoid of GPS signals, SLAM emerges as a dependable guiding principle, assuring high-precision navigation. Its transformative impact extends to the realms of augmented and virtual reality. The profound influence of SLAM has propelled advancements in computer vision \cite{Singandhupe2019}. At its core, SLAM's essence lies in its remarkable capacity to enhance perception, navigation, and cooperation, thereby solidifying its stature as a foundational technology \cite{Cadena2016}.

One challenge for outdoor SLAM is that laser SLAM often struggles with performance in adverse weather conditions such as rainy or snowy environments, while visual SLAM experiences significant accuracy degradation in nights or low-light conditions. Fortunately, the widely used 4D millimeter-wave radar offers superior penetration through rain and snow, while infrared cameras enable reliable imaging in complete darkness, which is expected to enhance SLAM performance in these environments. Another challenge is that in practice, LiDAR, 4D millimeter-wave radar, RGB cameras, infrared cameras, and IMUs all exhibit reduced robustness on bumpy terrains or surfaces with speed bumps.

To address the above chanllenges, we need to develop improved SLAM algorithms which require datasets of multiple types of sensors in adverse weather and road conditions.  However, to our knowledge, there is currently no such dataset available. To enrich this field, we introduce a variety of sensors and meticulously curate datasets from these extreme environments. This dataset provides support for professionals dedicated to researching multi-sensor fusion SLAM algorithms in these extreme conditions. 

Our primary contributions are as follows:

\begin{enumerate}
    \item Our dataset includes 4D millimeter-wave radar, infrared cameras, depth cameras, 3D LiDAR, RGB cameras, GPS, and IMU data. To our knowledge, there has been no previous work that includes all seven types of sensors and nine different types of information simultaneously.
    \item The collected scenarios encompass complex conditions such as snowy weather, nighttime, rough terrains, and roads with speed bumps. To the best of our knowledge, this is one of the few datasets featuring seven types of sensor information under diverse weather, lighting, and road conditions for SLAM.
    \item Our trajectories cover both structured and semi-structured terrains and include looped scenes to facilitate loop closure detection for SLAM systems. Reliable GPS/INS ground truth data is provided alongside raw data and processed TUM-format data.
    \item The dataset includes data collected from both ground robots and vehicles, catering to SLAM research for both robot navigation and autonomous driving. It spans large-scale and small-scale scenarios, with a total of 18.5 km, 69 minutes, and approximately 660 GB.
    \item We evaluated the performance of pure LiDAR SLAM, pure visual SLAM, pure 4D millimeter-wave radar SLAM, LiDAR\_IMU SLAM, visual\_IMU SLAM, LiDAR\_visual\_IMU SLAM, and Gaussian SLAM under various adverse environmental conditions. By comparing the ATE, RPE, and trajectory plots, we analyzed the challenges faced by different sensor combinations in various scenarios.
\end{enumerate}

\section{RELATED WORK}
\begin{table*}[htbp]
\caption{The table illustrates the scenarios and types of devices included in our latest dataset.}\vspace{-10pt}
\begin{center}
\begin{tabular}{l|ccccccccccccc|ccc} \hline
\textbf{Category}&\multicolumn{6}{|c|}{Road Conditions} & \multicolumn{7}{|c|}{Sensor Types}& \multicolumn{3}{c}{Devices}\\ \hline 
Dataset& Day& Night& Rain&Snow& \multicolumn{2}{c|}{Bump} & LiDAR& IMU& RGB& Stereo& Infrared & 4D Radar&GPS & Robot& {Car}\\ \hline 

KITTI&\checkmark& \checkmark&  && \multicolumn{2}{c|}{} & \checkmark& \checkmark& \checkmark& \checkmark& & &\checkmark& &{\checkmark}\\
 TUM&\checkmark& &  && \multicolumn{2}{c|}{} & & \checkmark& \checkmark& \checkmark& & && \checkmark&{}\\
 RSRD& \checkmark& & & & \multicolumn{2}{c|}{\checkmark} & \checkmark& \checkmark& \checkmark& & & &\checkmark& & {\checkmark}\\  
 M2DGR& \checkmark& \checkmark&  && \multicolumn{2}{c|}{} & \checkmark& \checkmark& \checkmark& & & &\checkmark& \checkmark& {}\\ 
 Urbanloco& \checkmark& \checkmark&  && \multicolumn{2}{c|}{\checkmark} & \checkmark& \checkmark& & & \checkmark& &\checkmark& & {\checkmark}\\ 
 RobotCar & \checkmark& \checkmark& \checkmark &\checkmark& \multicolumn{2}{c|}{\checkmark} & & \checkmark& \checkmark& & & &\checkmark& & {\checkmark}\\ 
 4Seasons& \checkmark& \checkmark&  \checkmark&\checkmark& \multicolumn{2}{c|}{} & \checkmark& \checkmark& \checkmark& & & &\checkmark& & {\checkmark}\\ 
 Brno Urban& \checkmark& \checkmark& \checkmark && \multicolumn{2}{c|}{\checkmark} & \checkmark& \checkmark& \checkmark& & \checkmark& &\checkmark& & {\checkmark}\\ 
 NTU4DRadLM& \checkmark& \checkmark& \checkmark && \multicolumn{2}{c|}{} & \checkmark& \checkmark& \checkmark& & \checkmark& \checkmark&\checkmark&\checkmark  & {\checkmark}\\  
 Goose& \checkmark& \checkmark& \checkmark& \checkmark& \multicolumn{2}{c|}{\checkmark} & \checkmark& \checkmark& \checkmark& & \checkmark& &\checkmark& & {\checkmark}\\  
 KAIST& \checkmark& \checkmark& & & \multicolumn{2}{c|}{}& \checkmark& \checkmark& \checkmark& & \checkmark& &\checkmark& & {\checkmark}\\ 
 NuScenes& \checkmark& \checkmark& \checkmark& & \multicolumn{2}{c|}{}& \checkmark& \checkmark& \checkmark& & & 3D&\checkmark& & {\checkmark}\\  
  K-Radar& \checkmark& \checkmark& \checkmark& \checkmark& \multicolumn{2}{c|}{}& \checkmark& \checkmark& \checkmark& \checkmark& & \checkmark&\checkmark& & {\checkmark}\\
 View of Delft& \checkmark& & & & \multicolumn{2}{c|}{}& \checkmark& \checkmark& \checkmark& & & \checkmark& \checkmark& & {\checkmark}\\ 
 Dual Radar & \checkmark& \checkmark& & & \multicolumn{2}{c|}{}& \checkmark& & \checkmark& & & \checkmark&\checkmark& & {\checkmark}\\ 
 Boreas& \checkmark& \checkmark& \checkmark& \checkmark& \multicolumn{2}{c|}{\checkmark} & \checkmark& \checkmark& \checkmark& & & 3D&\checkmark& & {\checkmark}\\ 
 OORD& \checkmark& \checkmark& \checkmark& \checkmark& \multicolumn{2}{c|}{\checkmark} & & \checkmark& & & & 3D&\checkmark& & {\checkmark}\\
 RADIATE& \checkmark& \checkmark& \checkmark& \checkmark& \multicolumn{2}{c|}{}& \checkmark& \checkmark& & \checkmark& & 3D& \checkmark& & &\\\hline
 DIDLM(ours)& \checkmark& \checkmark& \checkmark &\checkmark&\multicolumn{2}{c|}{\checkmark} & \checkmark& \checkmark& \checkmark& \checkmark& \checkmark& \checkmark&\checkmark& \checkmark&{\checkmark}\\\hline\end{tabular}
\vspace{-20pt}
\end{center}
\end{table*} 

Published in 2012, the KITTI dataset~\cite{Geiger2012} comprises data recorded in urban, rural, and highway environments. Published in 2016, the Oxford RobotCar dataset~\cite{oxford} contains synchronized LiDAR and camera recordings. Introduced in 2018, the TUM VI dataset~\cite{Schubert2018} offers both indoor and outdoor trajectories acquired with a handheld rig that integrates visual sensors and inertial measurement units. While these established datasets include LiDAR, IMU, and image data under standard operating conditions, they were not designed to address adverse environments. In contrast, our dataset is specifically collected to encompass a broad spectrum of challenging conditions—such as sunny, rainy, and snowy weather; day and night illumination; uneven road surfaces; and routes featuring frequent speed bumps—thereby providing extensive coverage for evaluating autonomous driving and robotic navigation systems under adverse scenarios.
\begin{figure}
\setlength{\fboxsep}{0pt}%
\setlength{\fboxrule}{0pt}%
\begin{center}
    \includegraphics[width=1\linewidth]{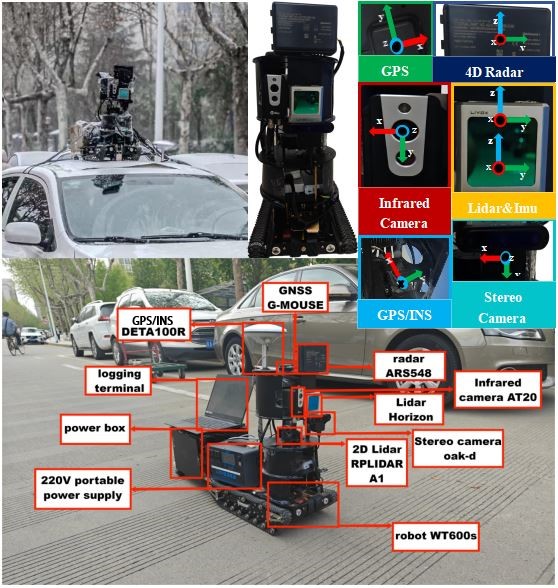}
\end{center}\vspace{-15pt}
\caption{The figure illustrates the actual ground robot and the vehicle equipped with sensors, with detailed labels indicating the function of each component. Additionally, the coordinates of each sensor are marked, with red representing the x-axis, green representing the y-axis, and blue representing the z-axis.}\vspace{-20pt}
\end{figure}

Published in 2020, the 4Seasons dataset~\cite{Wenzel2021b} covers challenging scenarios such as snowy conditions. Released in 2021, the M2DGR dataset~\cite{Yin2021} provides a wide range of indoor and outdoor sequences, diverse illumination, and unique settings such as elevators. Published in 2023 and 2024, the RSRD dataset~\cite{Zhao2023,Zhao2024} targets complex road surfaces and integrates LiDAR and visual data. Published in 2020, the UrbanLoco dataset~\cite{urban} captures dynamic objects and other challenging factors in urban scenes. Published in 2020, the Brno Urban dataset~\cite{brno} incorporates infrared cameras and adverse-weather sequences, yet lacks 4-D millimetre-wave radar and systematic coverage of compound challenges. Released in 2023, the GOOSE dataset~\cite{goose} also employs infrared sensors but targets unstructured environments without further adverse scenarios. Although these datasets all include adverse environments, they lack compound scenarios that simultaneously present multiple challenges. In contrast, our dataset simultaneously incorporates multiple adverse conditions—such as driving on bumpy roads at night in snowfall—which is essential for studying SLAM under compound adverse environments. Moreover, the aforementioned datasets do not employ sensors specialized for severe weather, while our dataset leverages 4D millimetre-wave radar in snow and infrared cameras at night.

Introduced in 2018, the KAIST Multi-Spectral Day/Night dataset~\cite{kaist} delivers synchronized recordings from RGB/thermal cameras, stereo RGB, 3-D LiDAR, and a high-precision GPS/IMU unit. Traversing urban, campus, and residential areas, the collection spans both daytime and nighttime, yet it omits adverse weather and rugged intersections. Critically, the sensor suite does not include 4-D millimetre-wave radar, the modality best suited for rain or snow. Our dataset fills these gaps by adding dedicated all-weather sensors and extensive sequences captured in harsh environments.

Published in 2019, nuScenes\cite{nuScenes} delivers 1000 driving scenes recorded in Boston and Singapore, although the majority were captured under conventional conditions. Collected from February 2019 to February 2020, RADIATE\cite{Sheeny} includes data under diverse weather conditions but omits bumpy-road scenarios. Introduced in 2020, Boreas\cite{Burnett} spans multiple seasons and adverse weather, while the large-scale OORD dataset\cite{OORD}, released in 2024, focuses on on-road driving in off-road environments. Existing datasets that feature millimetre-wave radar rely on 3-D units, which cannot accurately measure object height, suffer from low resolution, and are prone to motion blur. In contrast, 4-D millimetre-wave radar provides elevation information, robust penetration through precipitation, precise localisation, and cost-effective deployment, positioning it as a promising LiDAR alternative. 

Introduced in 2022, the View-of-Delft (VoD) dataset~\cite{view} delivers synchronized 4-D millimetre-wave radar, 64-beam LiDAR and stereo imagery, yet its recordings are confined to fair-weather urban traffic and lack adverse-condition scenarios. Also released in 2022, the Dual Radar dataset~\cite{dual} provides dual 4-D millimetre-wave radars, but similarly omits challenging weather or terrain. Published the same year, the K-Radar dataset~\cite{kradar} spans diverse weather and lighting, yet it was not collected on bumpy or uneven road surfaces. All three datasets include night-time sequences, yet none supplements RGB cameras with infrared sensors for superior nocturnal vision. In contrast, our dataset not only unifies multiple adverse environments—rain, snow, night, and rough roads—but also integrates infrared cameras to deliver high-quality night imagery, enhancing the robustness of multi-sensor-fusion SLAM systems under extreme conditions.

The work closest to ours is NTU4DRadLM~\cite{ntu}, introduced in 2023. This dataset uniquely integrates 4-D millimetre-wave radar, thermal cameras, IMU, 3-D LiDAR, RGB cameras, and RTK-GPS into one fully synchronized multi-sensor suite. However, it omits snowy sequences—the very conditions that best showcase the radar’s all-weather depth and the thermal camera’s night-vision advantages. Furthermore, its trajectories avoid speed-bump and uneven-road sections, and no stereo or depth imagery is provided. In contrast, our dataset retains the same multi-modal sensor suite while adding dedicated recordings in snow, snow-at-night, and bumpy-road scenarios, together with stereo depth to enhance night-time visual cues, enabling a more rigorous evaluation of multi-sensor-fusion SLAM under compound adverse environments. For comprehensive details regarding all datasets, please refer to Table I.

\begin{figure}[tbp]
\begin{center}
\includegraphics[width=0.5\linewidth]{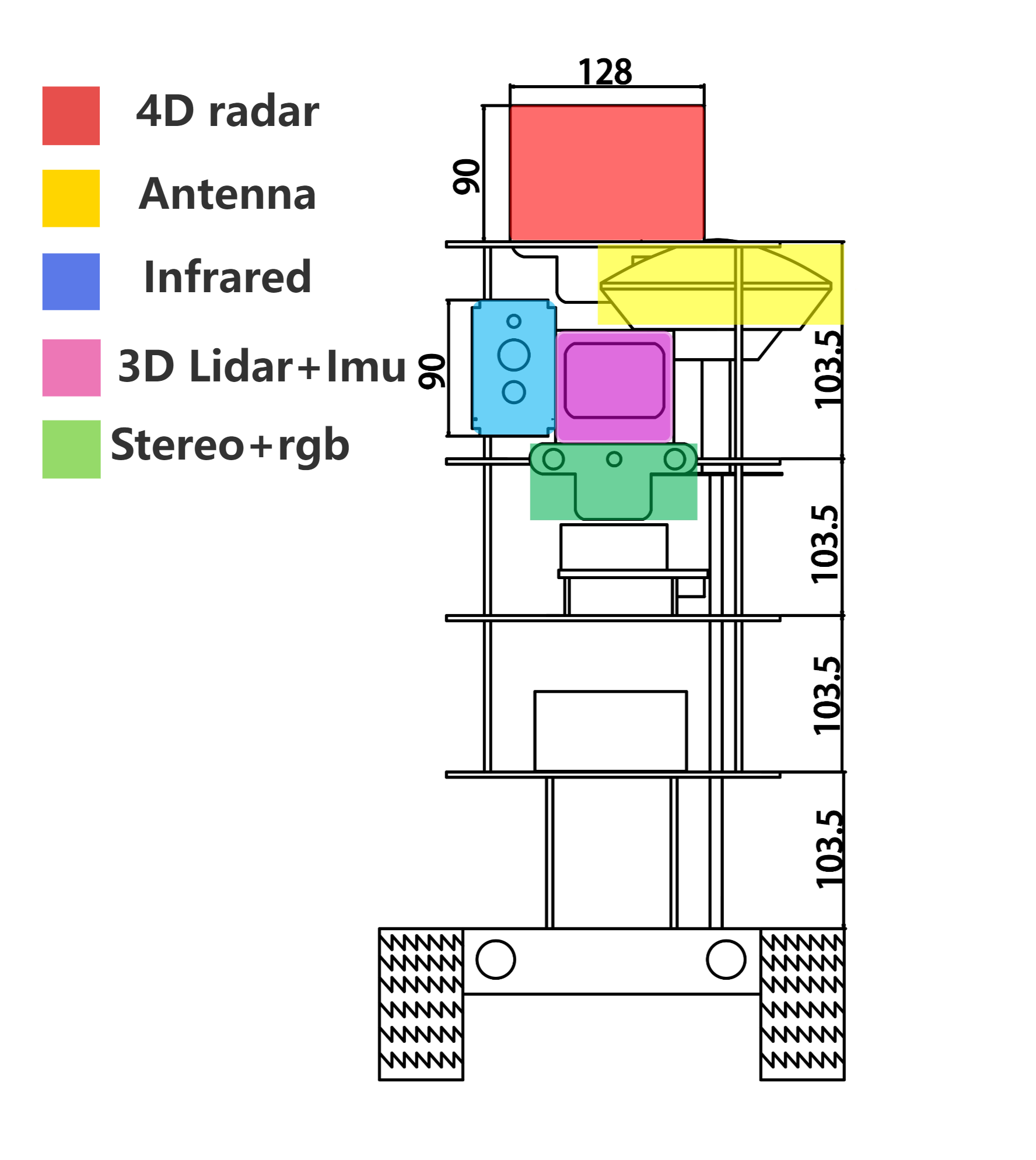}
\includegraphics[width=0.4\linewidth]{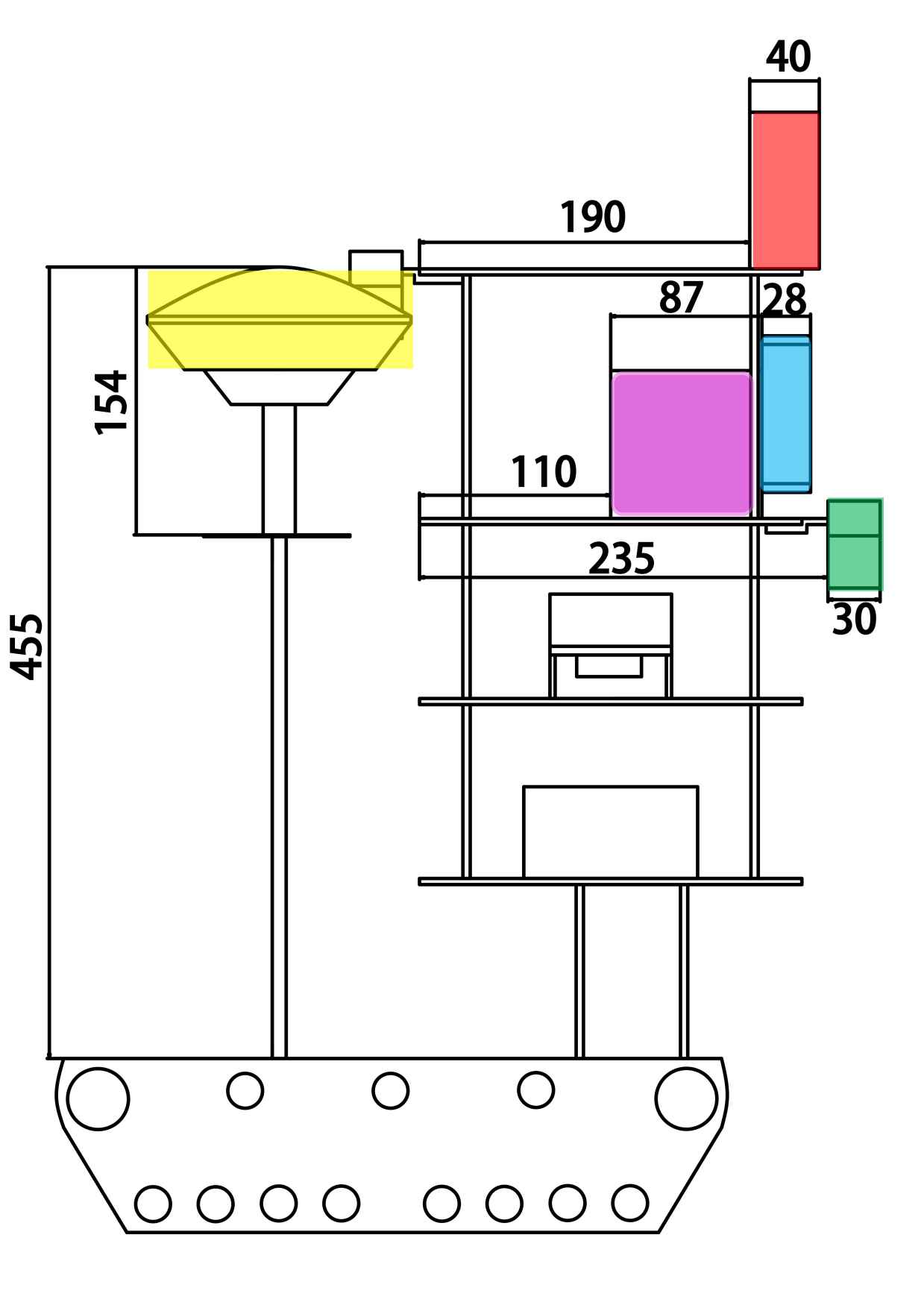}
\end{center}\vspace{-15pt}
\caption{Two-view diagram of the ground robot.}\vspace{-15pt}
\end{figure}

\section{The DIDLM Dataset}

DIDLM, which stands for “Difficult Scenarios Infrared Depth LiDAR Millimetre-wave,” fuses infrared cameras, depth cameras, 3-D LiDAR and 4-D millimetre-wave radar to enable robust localisation and mapping in challenging environments. We deploy a ground robot that is rigidly attached to a vehicle platform via a quick-release mount. The robot’s mechanical stack comprises five tiers: the upper three hold the sensors, while the lower two accommodate power supplies, interfaces, cabling and onboard computers. An overview of the collected data types is given in Fig. 1, and front and side views of the robot assembly are shown in Fig. 2.

\subsection{Sensors setup}
To achieve high-precision real-time localization, we employ a dot-matrix Livox system to construct three-dimensional laser point clouds. Additionally, for comprehensive data acquisition, we integrate a depth camera with an RGB camera to capture depth and image information from multiple perspectives. To address challenges posed by adverse weather conditions and low-light environments, we augment the setup with infrared cameras, ensuring accurate SLAM establishment under extreme conditions. Furthermore, we incorporate the latest 4D millimeter-wave radar sensors to enhance point cloud map construction and mitigate the impact of inclement weather on localization map development.

To complement ground-level robotic data collection, all sensor units are mounted on a vehicle roof platform to simulate high-speed data acquisition characteristic of autonomous driving scenarios. The sensors employed in this study, along with their technical specifications, are detailed in Tables II.
\begin{table*}[htbp]
\caption{Details of the sensors we used are provided. The stereo's baseline is 0.1136(m). }\vspace{-10pt}
\begin{center}
\begin{tabular}{llllllll} 
\hline 
\textbf{Category}   &\textbf{Manufacturer}&  \textbf{Type} &   \textbf{Distance range}&\textbf{FOV (H × V × D)}&\textbf{Resolution}&\textbf{Special Spec.}& \textbf{Freq.} \\ 
\hline 
Lidar  &Livox& Horizon &   260m(±2cm)& 81.7°×25.1°×0°&-&-& 10HZ\\
 imu& livox& BMI088&   -& -&-&6-axis&200HZ\\  
4D Radar&Continental & ARS548 &   0.2-301 m(±0.15 m)& 50°×20°×0°±0.1° … ±0.5°&0.22m&-& 20Hz \\ 
RGB &Luxonis& IMX378 &   -& 69°×55°×81°&1920*1080(pixels)&-& 30Hz\\ 
stereo &Luxonis& OV9282 &   0.7-12m(±0.12 m)& 72°×50°×82°&640*400(pixels)&-& 30Hz\\ 
Infrared &InfiRay& AT20 &   -& 56°×42°×0°&1024×768(pixels)&-20℃-550℃& 25Hz\\  
GPS&wheeltec& DETA100R &   -& -&-&1m & 200Hz \\
IMU &wheeltec& DETA100R &   -& -&-&9-axis & 400Hz \\ 
\hline
\end{tabular}
\label{tab3}\vspace{-15pt}
\end{center}
\setlength{\abovecaptionskip}{0cm} 
\setlength{\belowcaptionskip}{-0.2cm}
\small
\end{table*}

\subsection{File format and topic name}
All data was packaged into ROSBAG files. The ground truth format has been converted into the easily usable tum format, and the original format of GPS/INS has also been included in the dataset for everyone's use.\begin{table}[htbp]\vspace{-10pt}
\caption{Our topics, message types, and sampling frequencies are provided below.} \vspace{-10pt}
\begin{center}
\begin{tabular}{llll} \hline 
\textbf{Sensor}  &\textbf{Topics}  &\textbf{Message type} &\textbf{Rate}\\ \hline 
LiDAR &/livox/lidar&livox\_ros\_driver/CustomMsg&10Hz\\
 &/livox/imu&sensor\_msgs/Imu&200Hz\\ 
Camera&/color/compressed&sensor\_msgs/CompressedIm&30Hz\\
 & & age&\\
 &/left/compressed&sensor\_msgs/CompressedIm&30Hz\\
 & & age&\\
 &/right/compressed&sensor\_msgs/CompressedIm&30Hz\\
 & & age&\\
 &/pointcloud/raw &sensor\_msgs/PointCloud2&30Hz\\
 & /depth& sensor\_msgs/CompressedIm&30Hz\\
 & & age&\\ 
Radar &/ars548.detectionl&-&20Hz\\
 & ist& &\\
 &/ars548.objectList&-&20Hz\\ 
Infrared&/infrared  &sensor\_msgs/Image&25Hz\\ 
GPS/INS &-&-&200Hz\\ 
GT(tum)&/timestamps  &-&-\\ 
&/x, /y, /z  &-&-\\ 
&/qx, /qy, /qz, /qw  &-&-\\ \hline 
\end{tabular}
\label{tab4}\vspace{-20pt}
\end{center}
\setlength{\abovecaptionskip}{0cm} 
\setlength{\belowcaptionskip}{-0.2cm}
\small
\end{table} Detailed specifications for all topics can be found in Table III. The data storage format is as follows:

\texttt{data name.zip:}

\texttt{--color/compressed/Timestamps.png}

\texttt{--left/compressed/Timestamps.png}

\texttt{--right/compressed/Timestamps.png}

\texttt{--depth/Timestamps\_depth.png}

\texttt{--infrared/Timestamps.png}

\texttt{--livox/lidar/Timestamps.pcd}

\texttt{--radar/Timestamps.pcd}

\texttt{--pointcloud/raw/Timestamps.pcd}

\texttt{--livox/imu/imu.txt}

\texttt{--GT.txt}

\texttt{--GTorig.txt}

\subsection{Sensor Calibration}

Our dataset comprises 7 types of sensors and 9 data modalities. To ensure consistency and facilitate multi-sensor fusion, we have designated the LiDAR coordinate system as the primary reference frame. All other sensors, including a RGB camera, left/right cameras, an infrared camera, a 4D millimeter-wave radar, and an IMU have been calibrated with respect to this LiDAR coordinate system, with their extrinsic parameters precisely calculated.

 Furthermore, we complete intrinsic calibration for all optical sensors (RGB, left/right, and infrared cameras) as well as the IMU.

\begin{figure}[htbp]
\setlength{\fboxsep}{0pt}%
\setlength{\fboxrule}{0pt}%
\begin{center}
    \includegraphics[width=4cm,height=2.5cm]{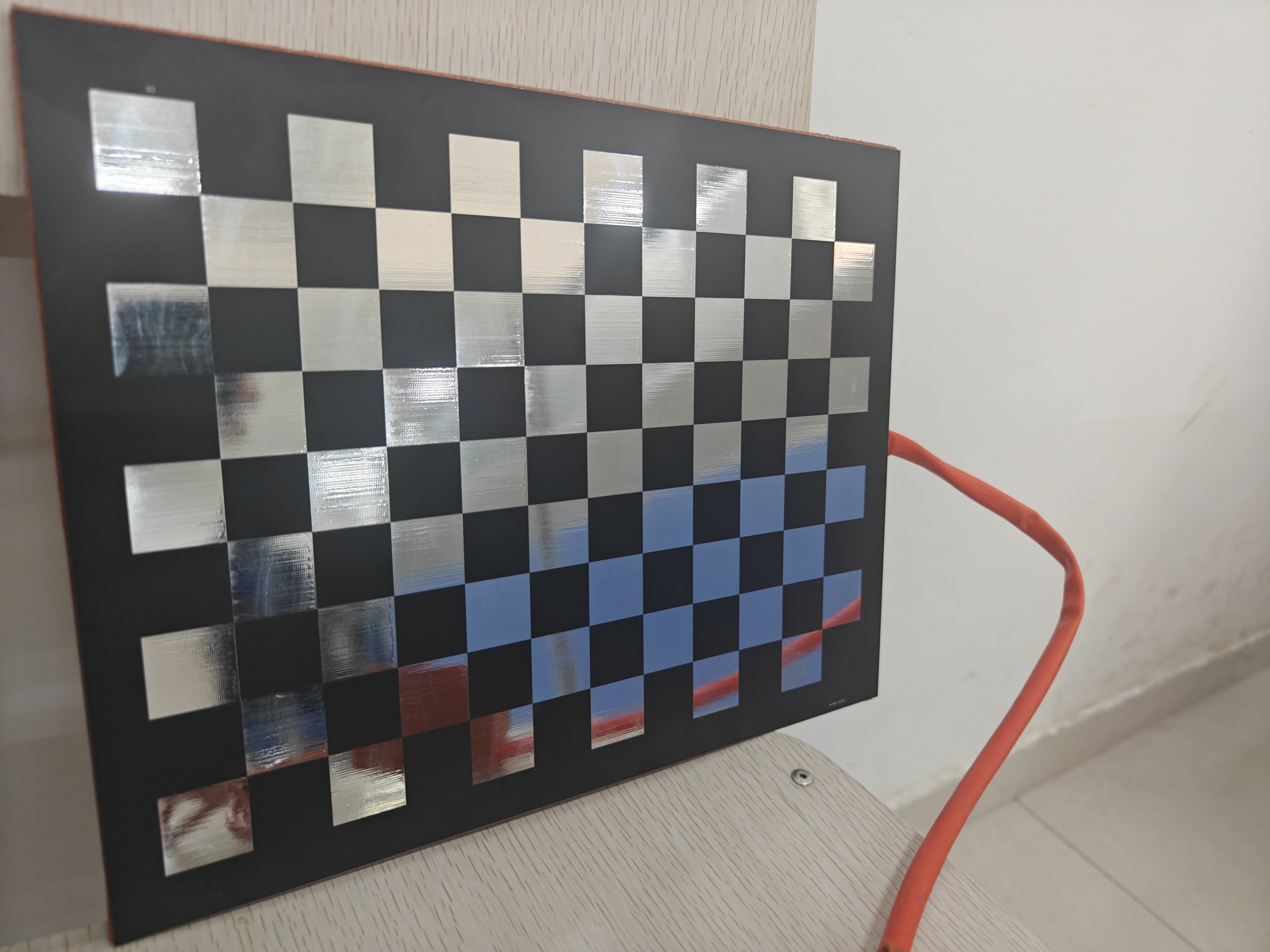}\includegraphics[width=4cm,height=2.5cm]{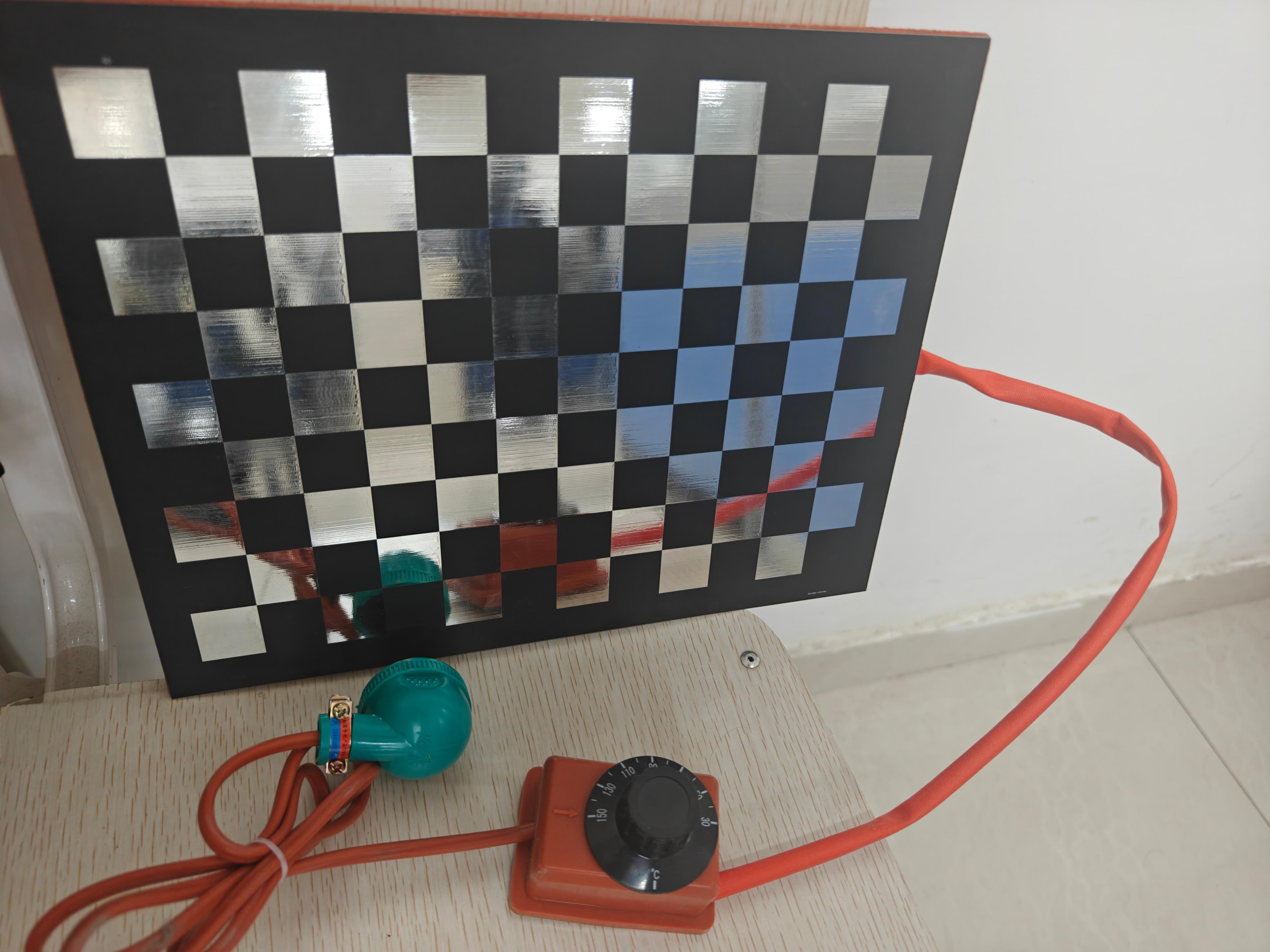}
    \includegraphics[width=4cm,height=2.5cm]{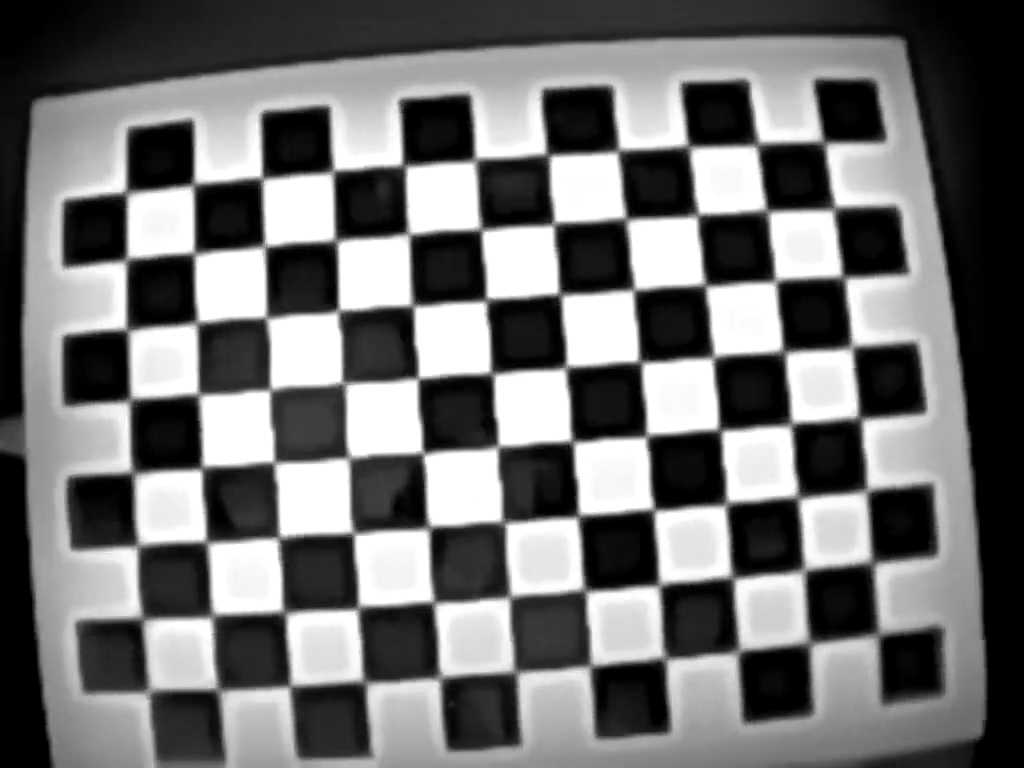}\includegraphics[width=4cm,height=2.5cm]{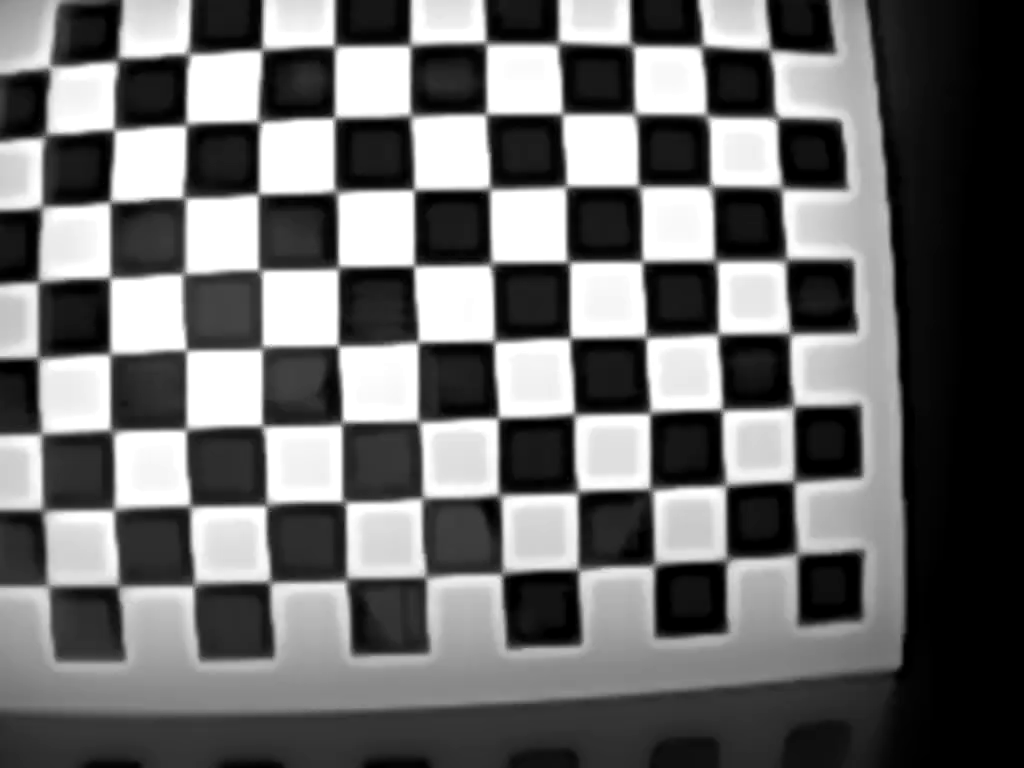}
    \includegraphics[width=4cm,height=2.5cm]{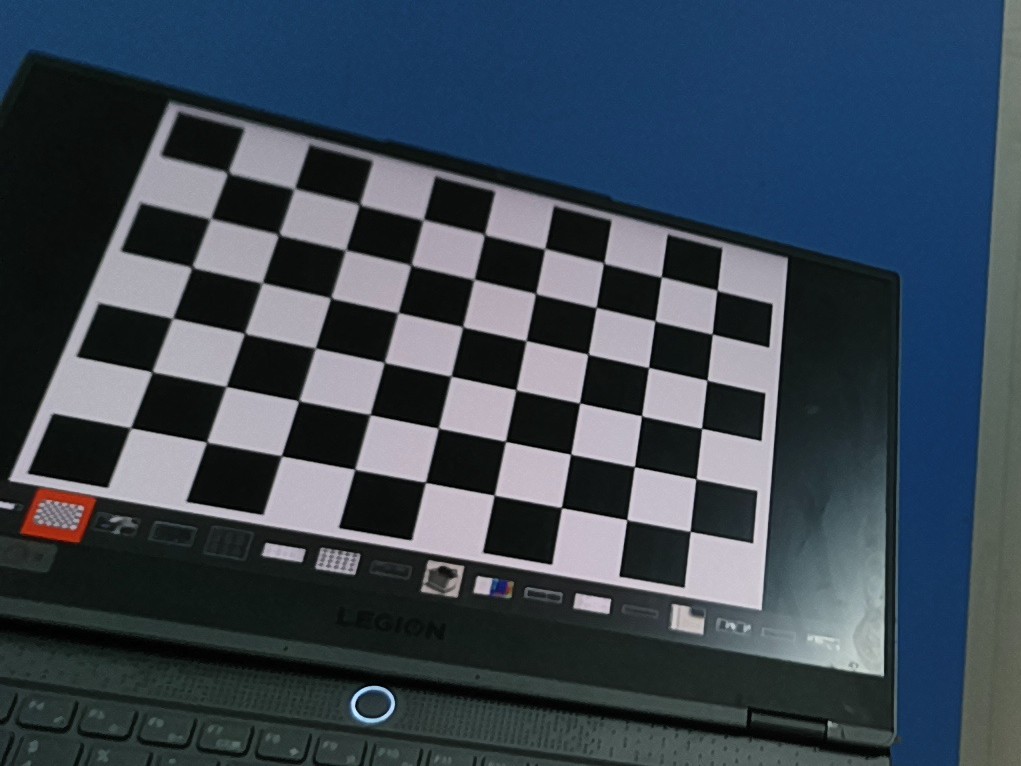}\includegraphics[width=4cm,height=2.5cm]{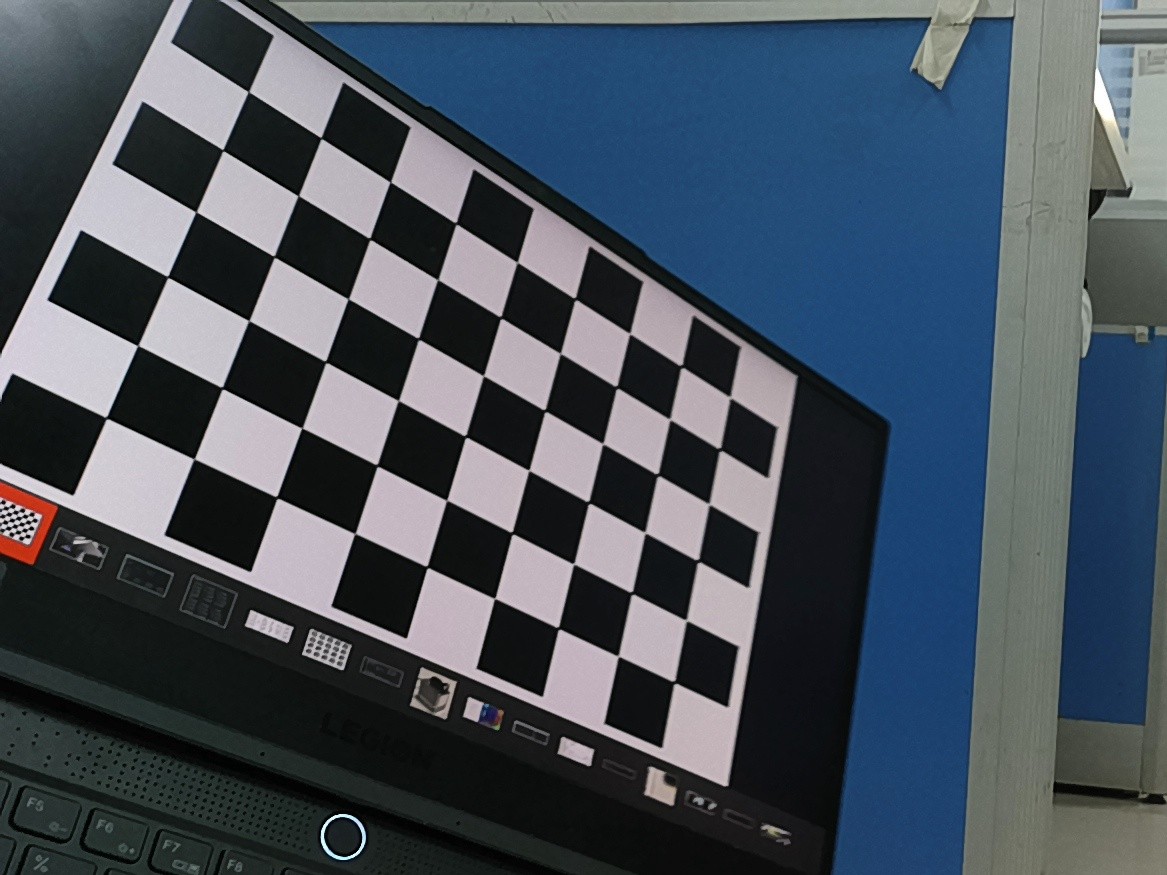}
    \includegraphics[width=4cm,height=2.5cm]{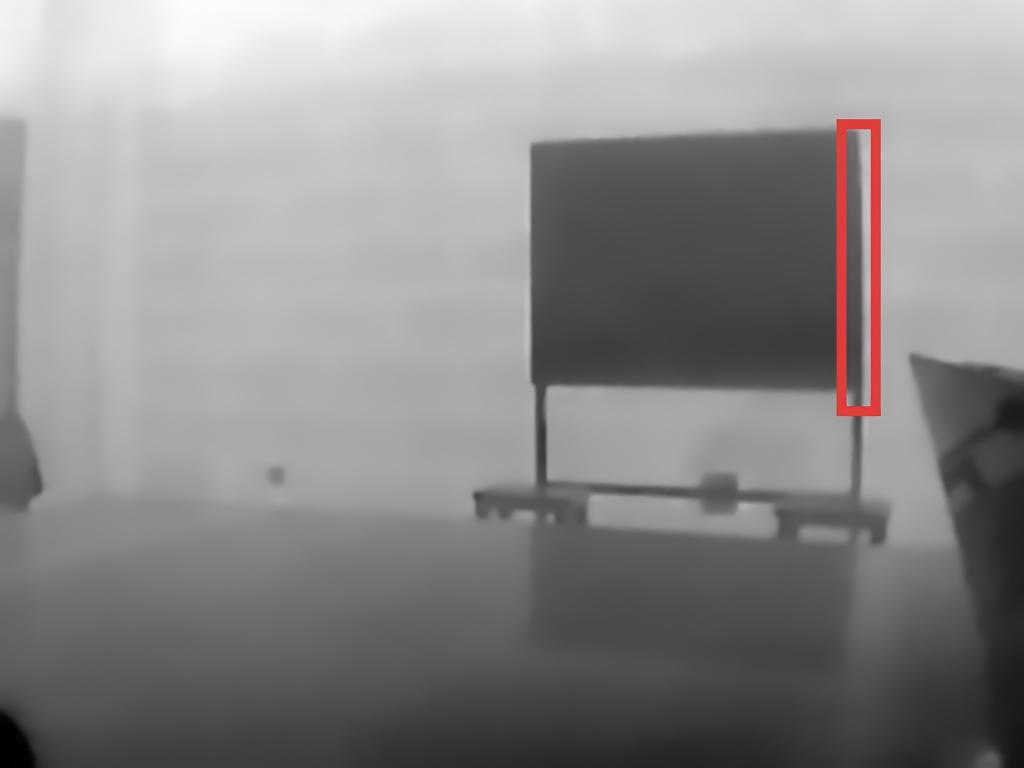}\includegraphics[width=4cm,height=2.5cm]{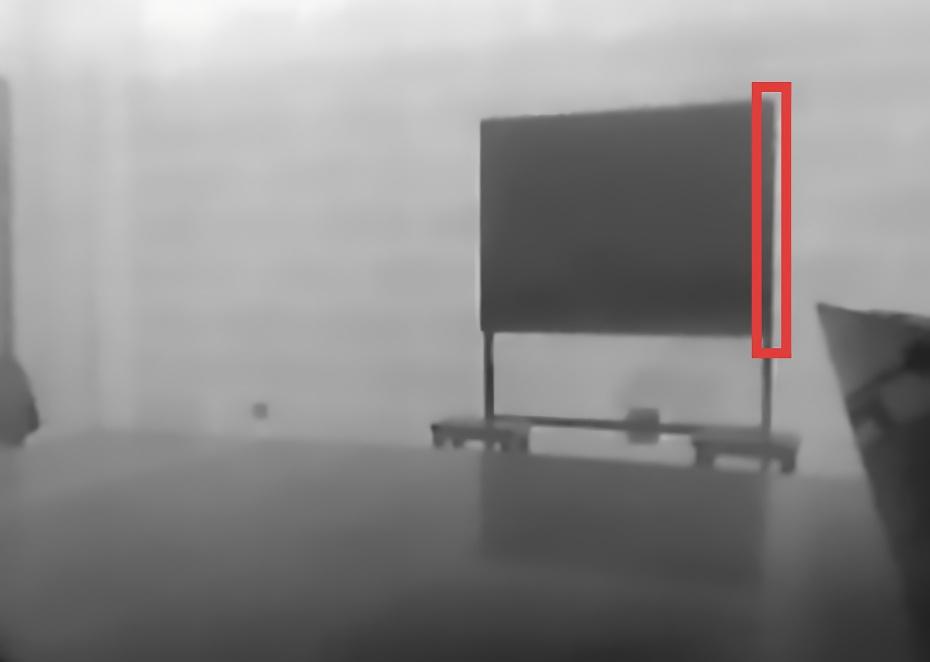}
    \includegraphics[width=4cm,height=2.5cm]{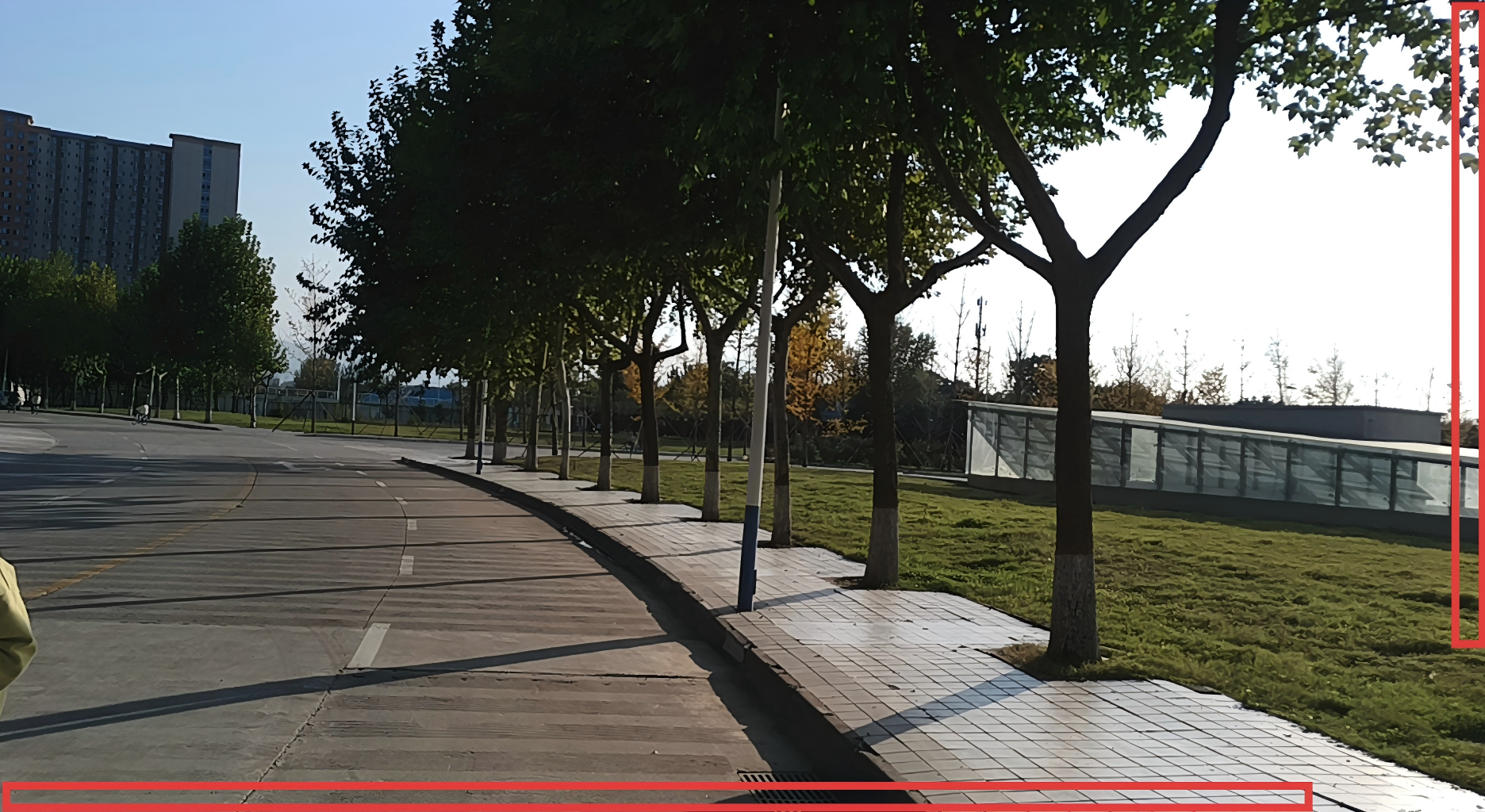}\includegraphics[width=4cm,height=2.5cm]{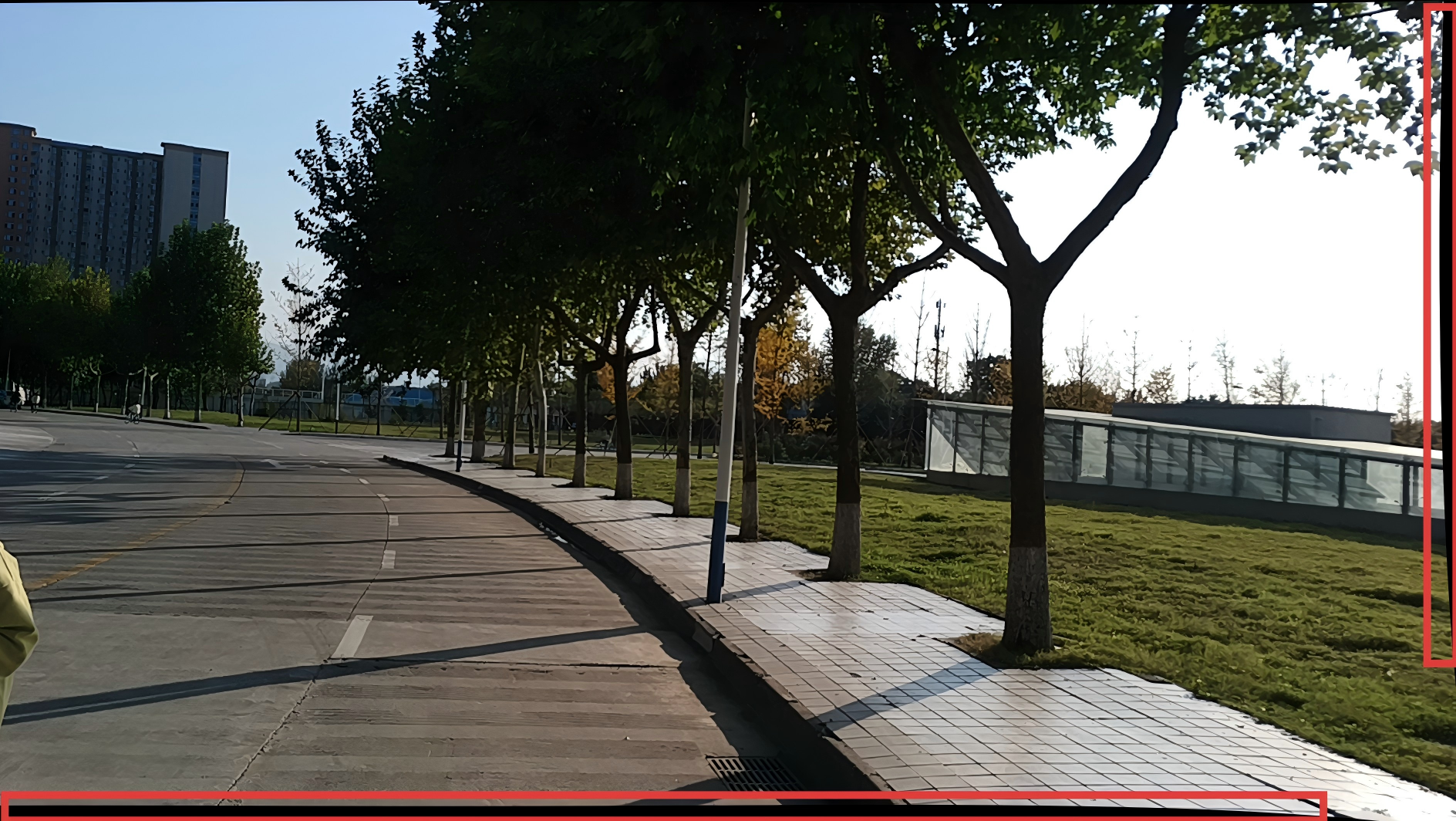}
\end{center}\vspace{-13pt}
\caption{The diagram illustrates the intrinsic calibration of the infrared camera and RGB camera using a chessboard pattern. The fourth and fifth rows show the infrared and RGB images before and after intrinsic calibration, respectively. The corrected areas in the fifth row are highlighted with red boxes, showing  that the image distortion has been rectified using the intrinsic parameters.}\vspace{-13pt}
\end{figure}
\subsubsection{imu intrinsic}
We employ the open-source algorithm imu-utils\cite{imu} to calibrate the IMU. The calibrated parameters obtain through Allan variance analysis are as follows: Accelerometer Noise Density (AND) is $0.0198\ \text{m/s}^2/\sqrt{\text{Hz}}$, Accelerometer Random Walk (ARW) is $0.0005\ \text{m/s}^2/\sqrt{\text{Hz}}$; Gyroscope Noise Density (GND) is $0.0025\ \text{rad/s}/\sqrt{\text{Hz}}$, and Gyroscope Random Walk (GRW) is $0.00003077\ \text{rad/s}/\sqrt{\text{Hz}}$. These low noise levels indicate that the IMU is suitable for high-precision applications.

\subsubsection{camera intrinsic}
\begin{figure*}[tbp]
\begin{center}   
    \includegraphics[width=4.25cm,height=2.5cm]{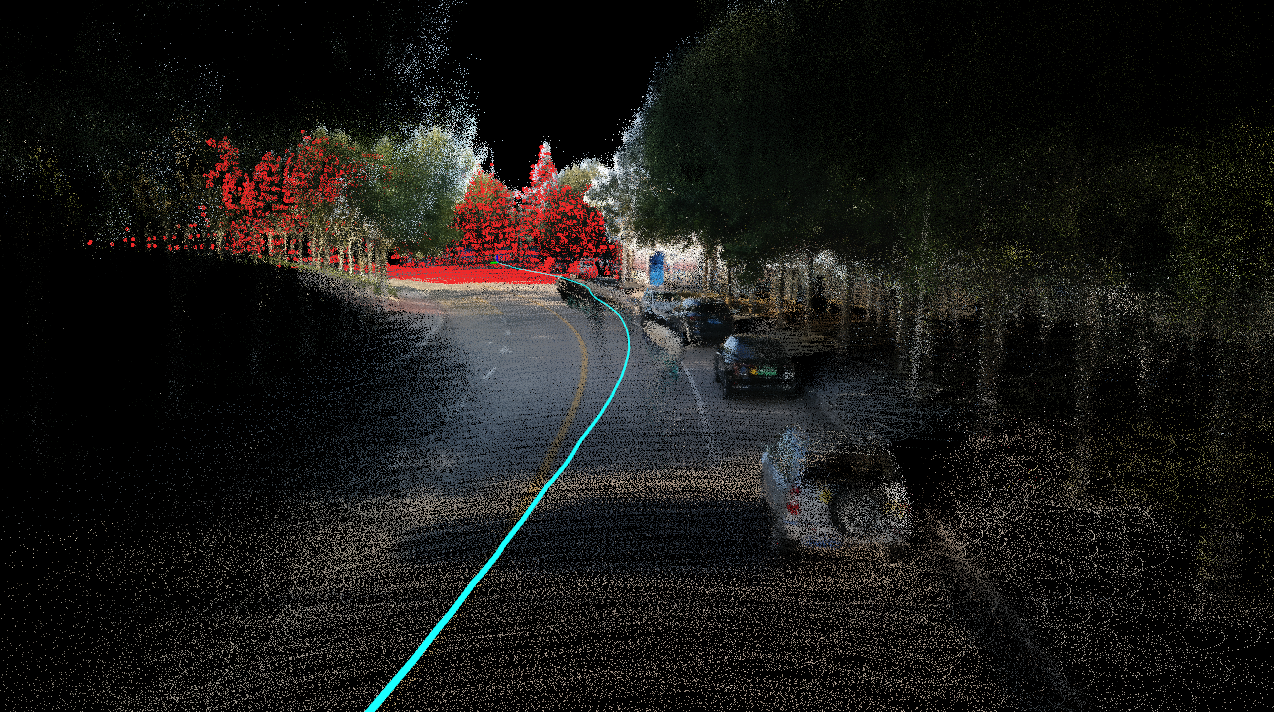}\includegraphics[width=4.25cm,height=2.5cm]{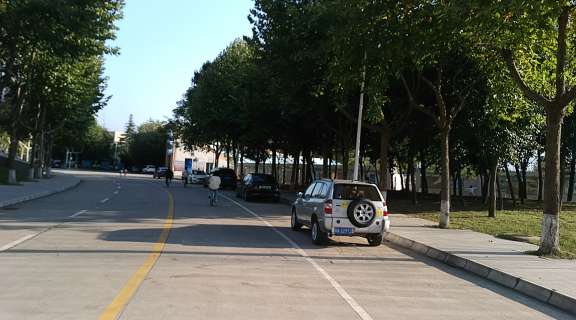}\includegraphics[width=4.25cm,height=2.5cm]{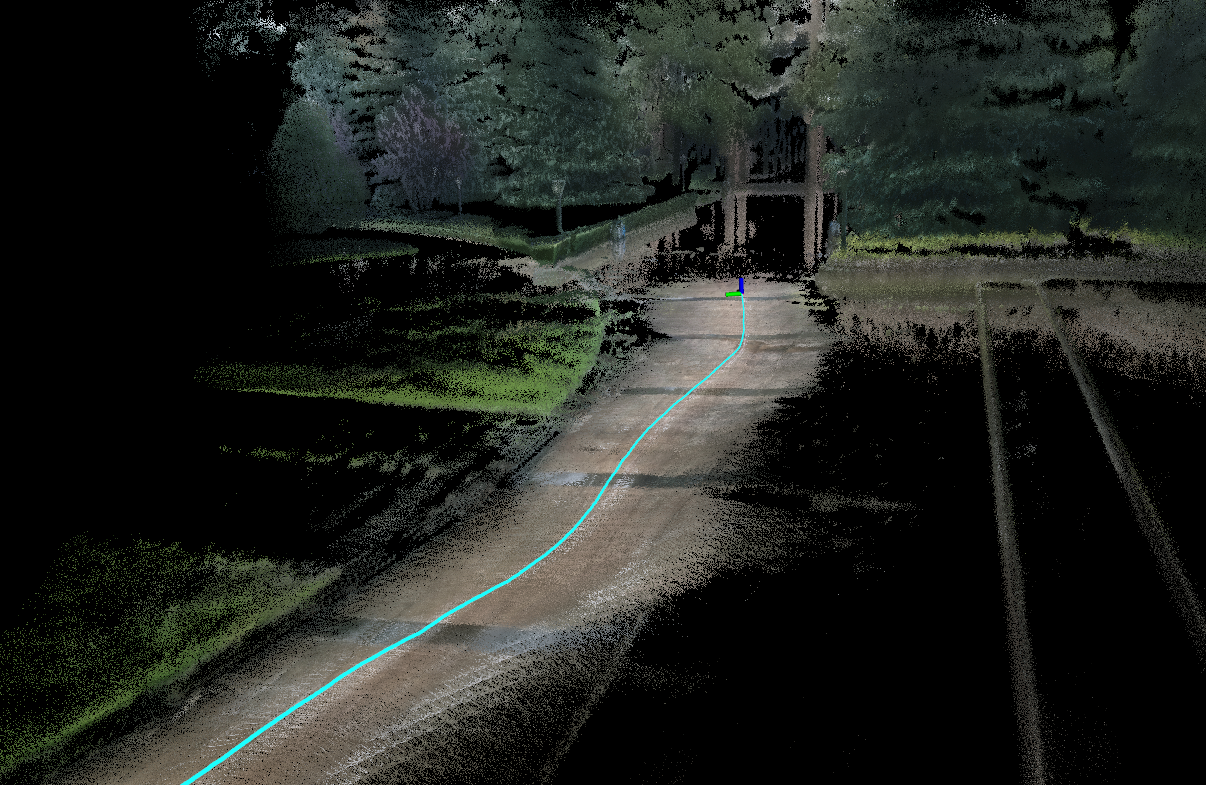}\includegraphics[width=4.25cm,height=2.5cm]{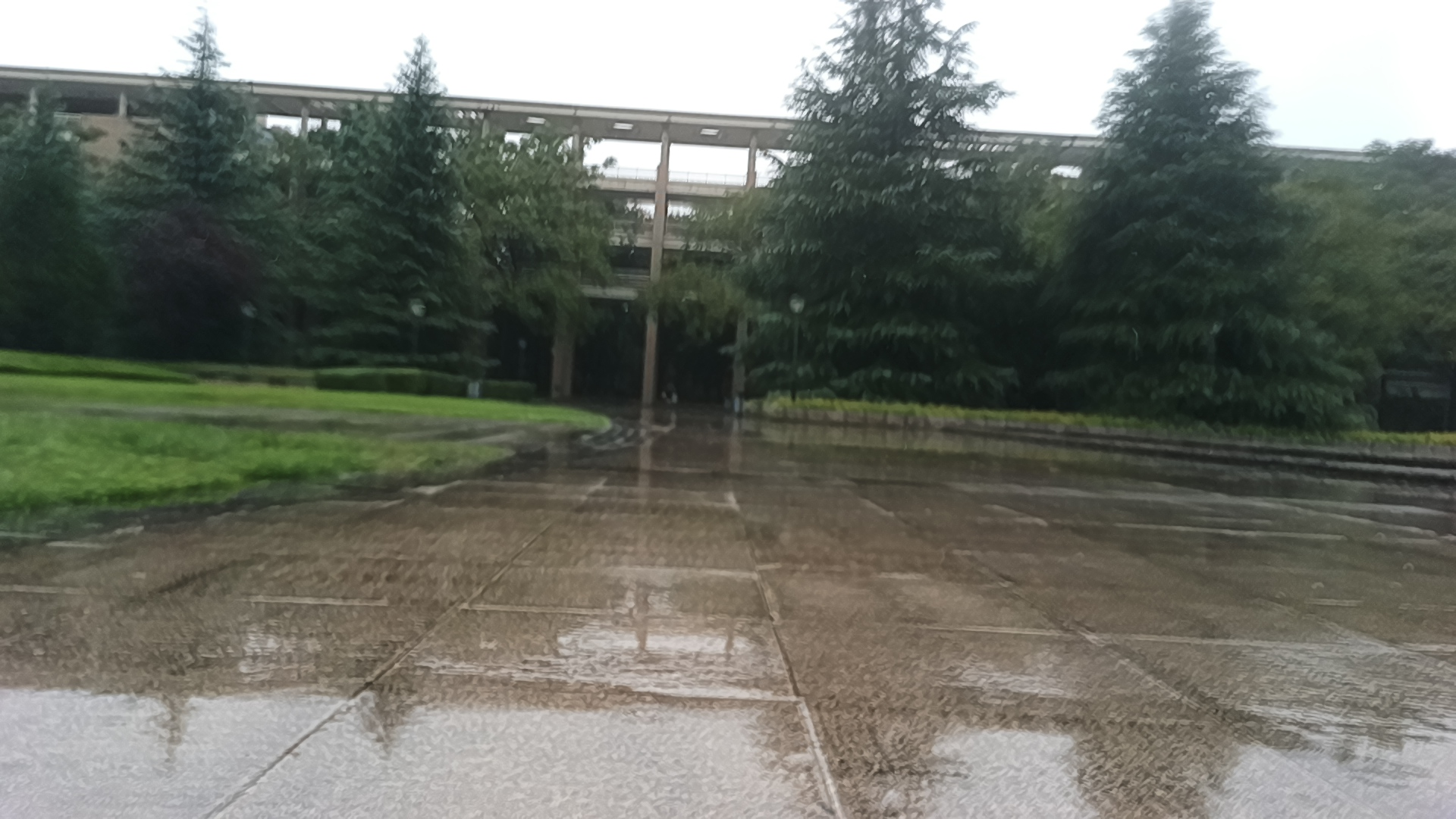}
    \includegraphics[width=4.25cm,height=2.5cm]{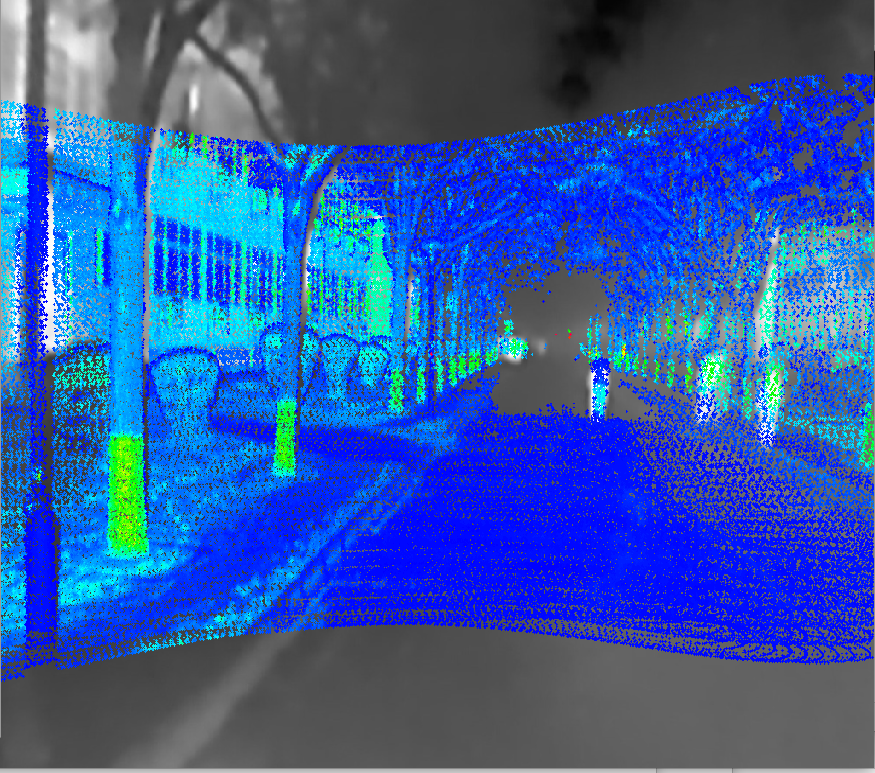}\includegraphics[width=4.25cm,height=2.5cm]{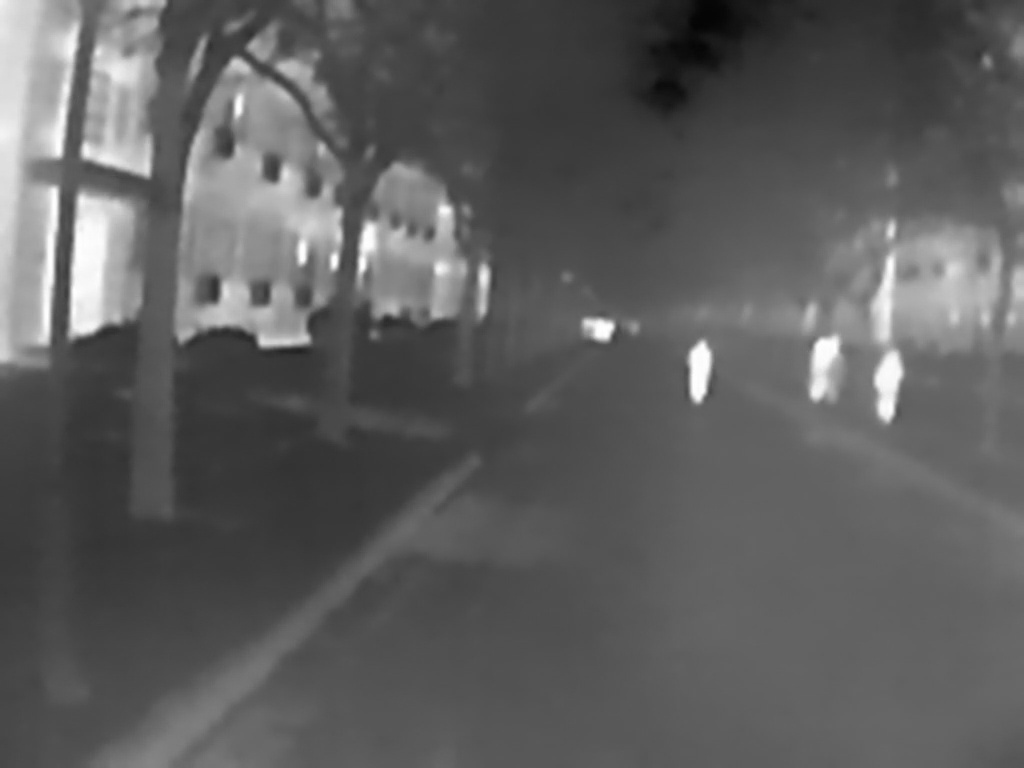}\includegraphics[width=4.25cm,height=2.5cm]{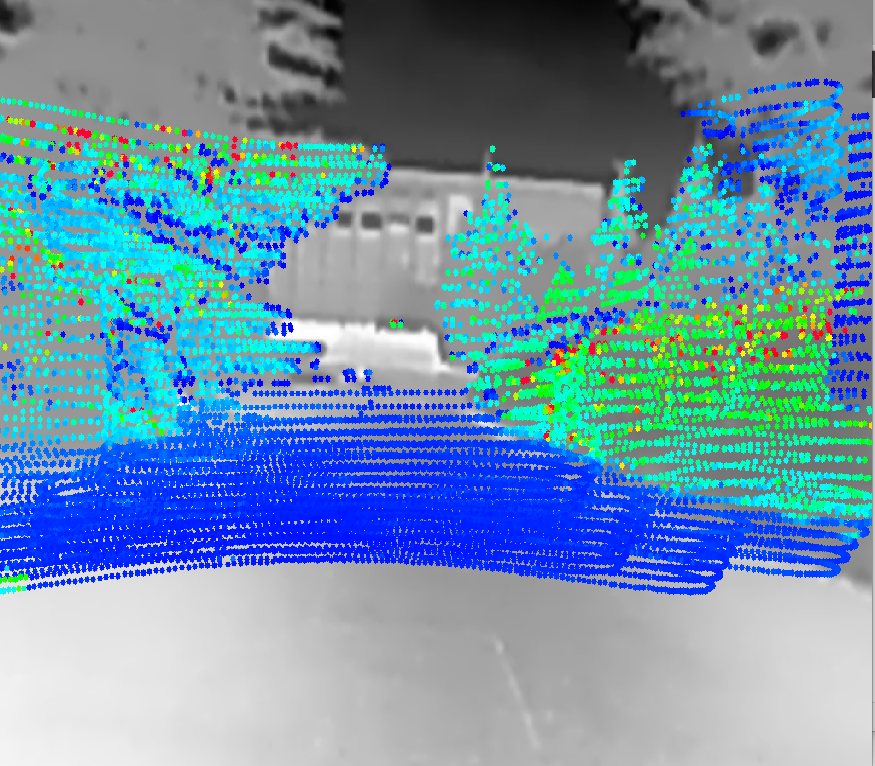}\includegraphics[width=4.25cm,height=2.5cm]{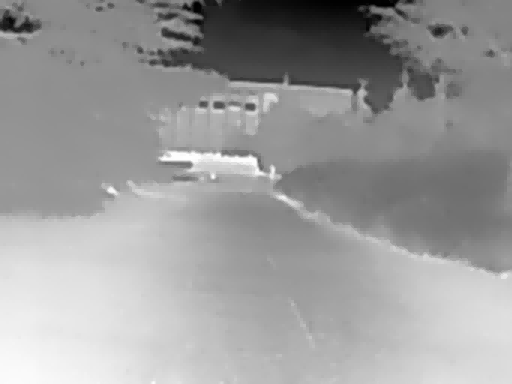}
    \includegraphics[width=4.25cm,height=2.5cm]{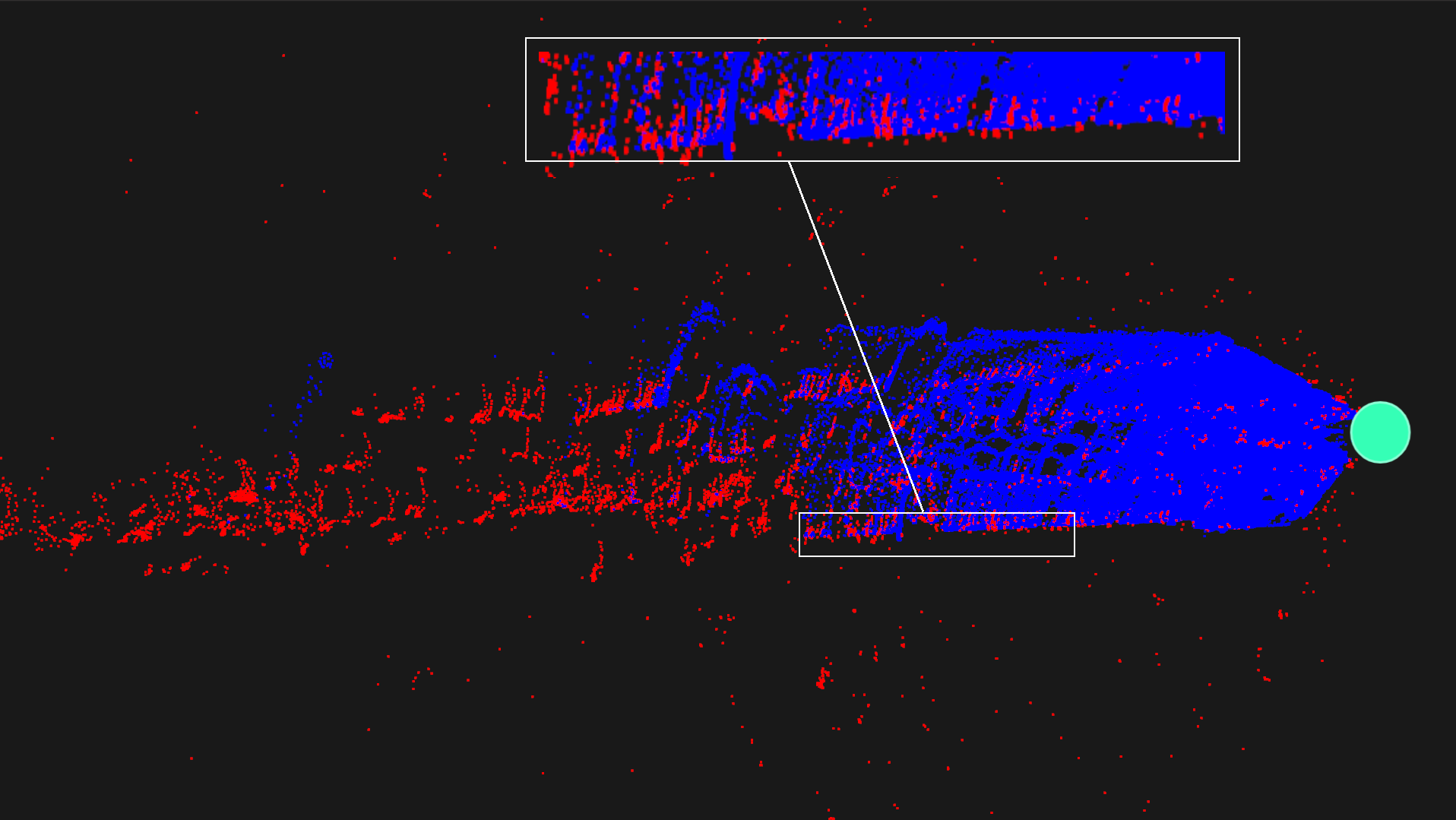}\includegraphics[width=4.25cm,height=2.5cm]{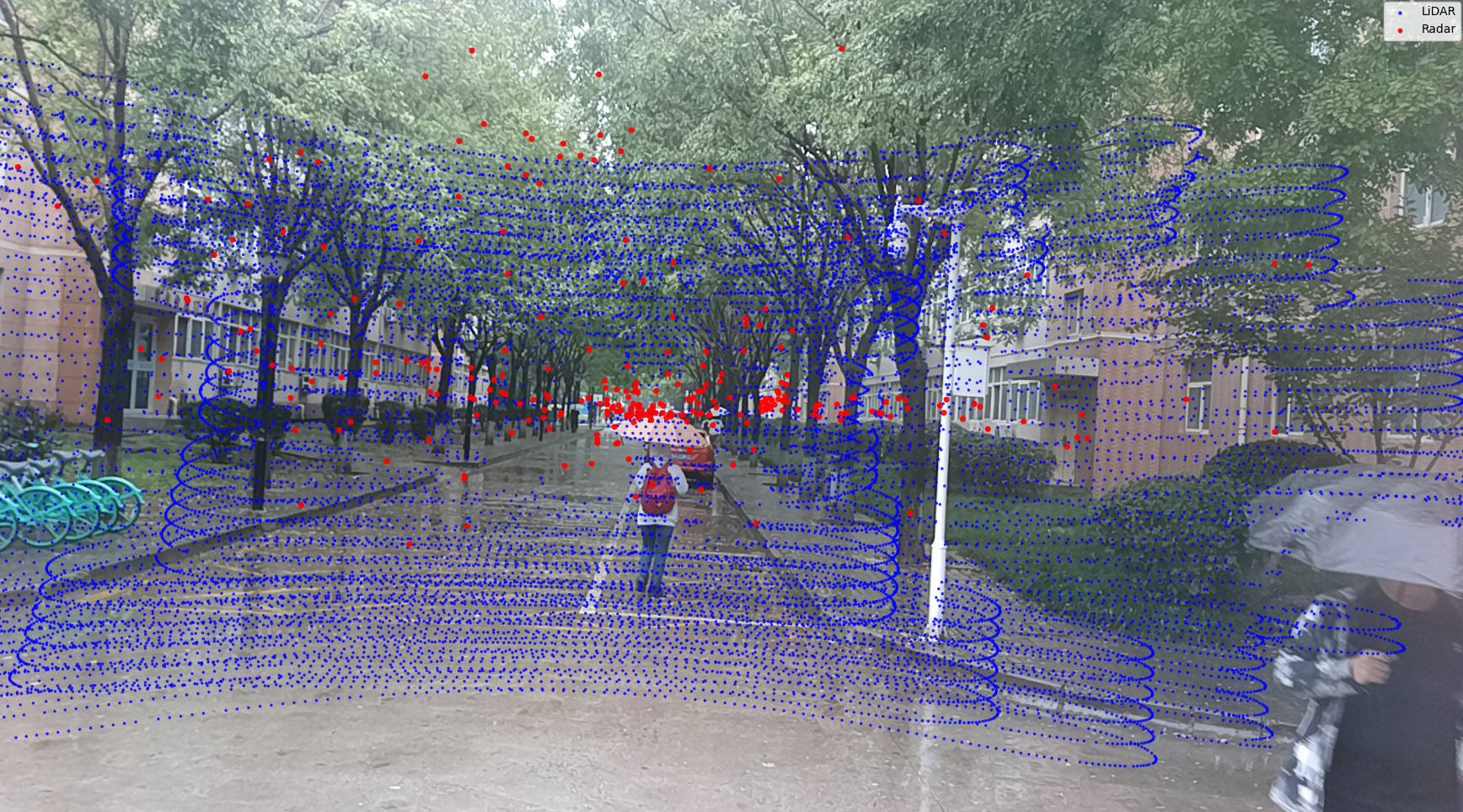}\includegraphics[width=4.25cm,height=2.5cm]{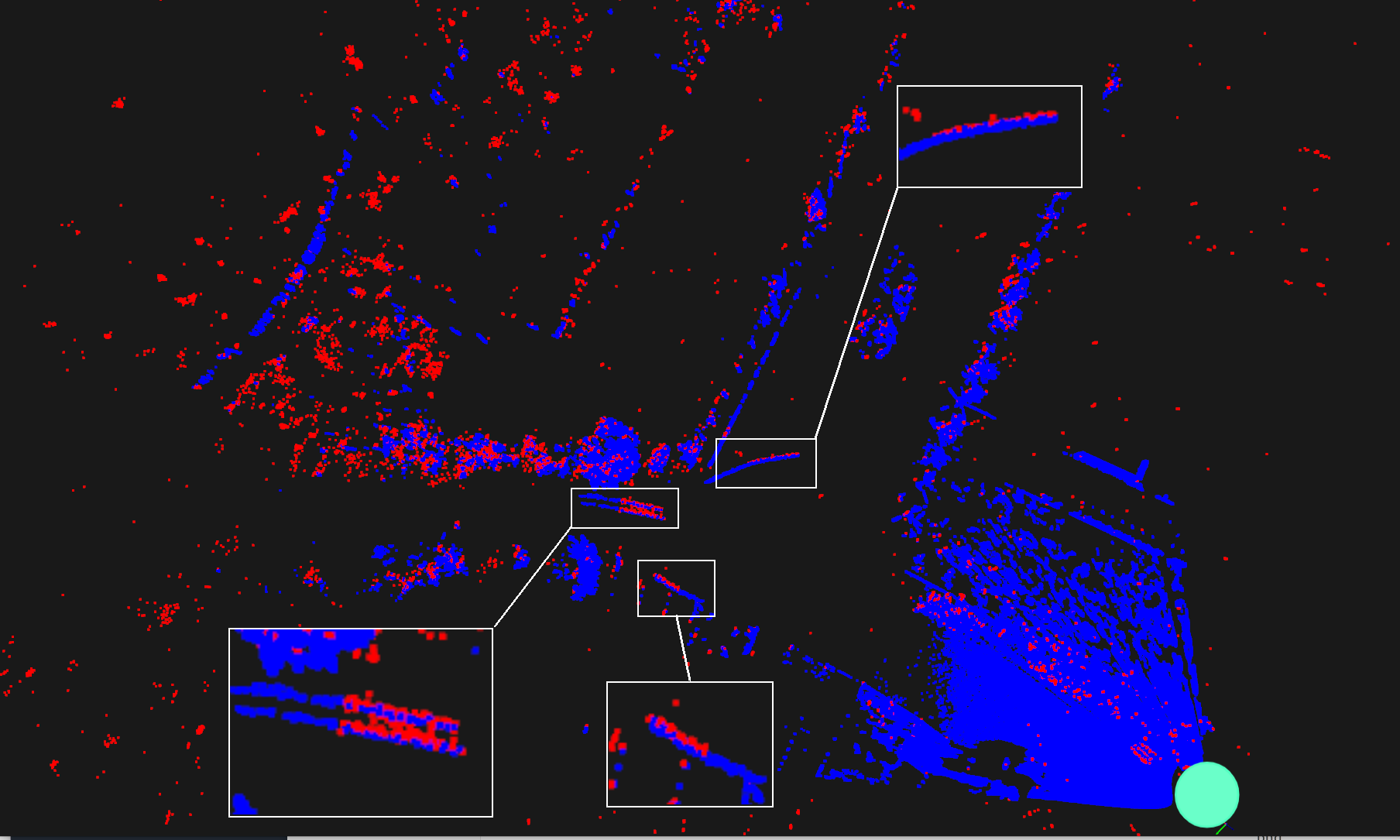}\includegraphics[width=4.25cm,height=2.5cm]{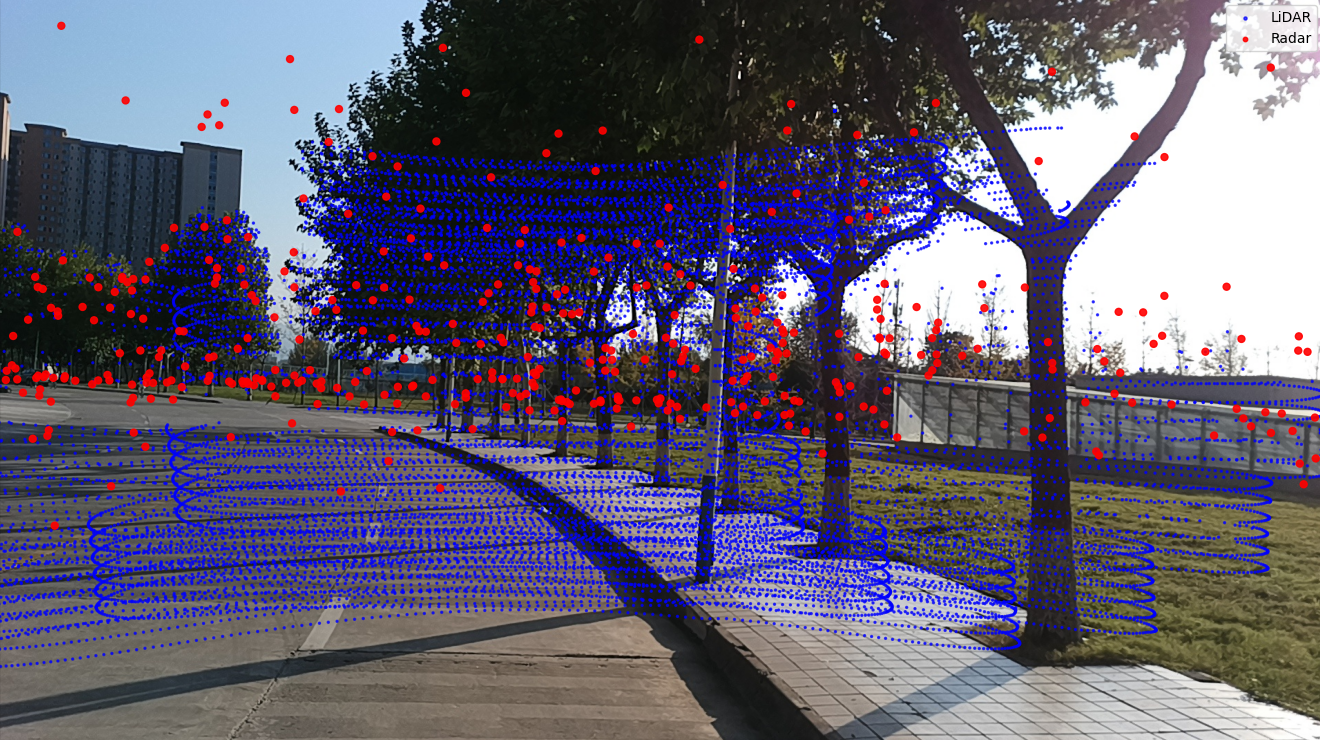}
\end{center}\vspace{-10pt}
\caption{The first row shows the LiDAR-camera extrinsic calibration using FastLIVO2\cite{fastlivo2}. The second row shows the LiDAR-infrared image calibration. The third row shows the LiDAR-radar calibration. Left: Top view of point cloud alignment (blue: LiDAR, red: radar, green: origin). Right: Lidar-Radar-image alignment.}\vspace{-15pt}
\end{figure*}
We employ the calibration method livox\_camera\_lidar\_calibration\cite{livox} provided by LIVOX to calibrate the intrinsic parameters of the RGB camera. This method employs feature point matching and geometric constraints to calibrate the intrinsic parameters. During the calibration process, we use a traditional checkerboard pattern. To ensure the flatness of the checkerboard, we display it on a zero-curvature computer screen and captured 20 sets of images for calibration. The checkerboard have a size of 24 mm and a configuration of 6×9. The overall average error for the 20 sets of images is \(0.100751\) pixels.

Considering the unique imaging principles of the infrared camera, we adopt a custom electrified metal checkerboard to collect higher-precision intrinsic parameter information. The metal checkerboard have a size of 25 mm and a configuration of 8×11, and we capture 23 images for calibration. The average calibration error for the 23 images is 0.158631 pixels, which is significantly less than one pixel, indicating sufficiently high precision.

The calibration process and equipment used for these cameras are shown in Fig. 3. From the error values, we can quantitatively analyze that the calibration results have high precision.

\subsubsection{camera\_imu extrinsic}

We calibrate the extrinsic parameters between the IMU, RGB camera, and infrared camera using the Kalibr algorithm \cite{kar}. Kalibr is based on the principle of jointly optimizing the camera and IMU parameters by minimizing the reprojection errors of the camera and the inertial measurement errors. During the calibration process, we synchronously collect 2 minutes of IMU and camera data while fully exciting motion along the x, y, and z axes. Our calibration results show reprojection error is 0.25 pixels. Subsequently, we use the VINS-Mono\cite{vin} system to individually refine the extrinsic parameters for each dataset. 

\subsubsection{lidar\_imu extrinsic}

We employ lidar\_imu\_init\cite{lii} to calibrate the extrinsic parameters between the LiDAR and IMU. Additionally, we validate these parameters using FastLIO2\cite{fastlio2}. Our experiments demonstrate that the integration of the LiDAR and IMU performs well.

\subsubsection{lidar\_camera extrinsic}
We employ the livox\_camera\_lidar\_calibration\cite{livox} algorithm for the extrinsic calibration between the LiDAR and multiple cameras. This algorithm is based on 2D-3D correspondences and is provided by Livox. It is primarily used for the extrinsic calibration between the LiDAR and cameras.

In our experiments, the average calibration error between the LiDAR and the RGB camera is 1.72177 pixels in the vertical direction and 1.39296 pixels in the horizontal direction. The average calibration error between the LiDAR and the infrared camera is 2.90876 pixels in the vertical direction and 2.0405 pixels in the horizontal direction.

 To qualitatively analyze the calibration results, we input the calibrated RGB images and LiDAR data into FastLIVO2\cite{fastlivo2}, as shown in Fig. 4. We also input the calibrated infrared images and LiDAR data into the visualization interface. The results demonstrate that our calibration has achieved sufficiently high precision. To further enhance the accuracy, we perform fine-tuning for all scenarios and provid the generated YAML files. 

\subsubsection{lidar\_radar extrinsic}

\begin{table}[htbp]\vspace{-10pt}
\small\sf\centering
\caption{The types of data we collected include specific details such as distance traveled, recording duration, traveling speed, and data size.}\vspace{-5pt}
\begin{tabular}{lccc}
\hline
Sequence & Distance (m)& Duration (s) & Speed (m/s) \\\hline
\texttt{L-B-SU-D}& 161& 233& 0.8\\
\texttt{L-B-SU-N}& 217& 260& 0.9\\
\texttt{L-B-R-N}& 257& 270& 1.0\\
\texttt{L-NB-SU-D}& 377& 585& 0.7\\
\texttt{L-NB-SU-N}& 331& 352& 0.9\\
 \texttt{L-NB-R-D}& 336& 352& 0.9\\
\texttt{H-B-SU-D}& 158& 29& 5.5\\
\texttt{H-B-R-D}& 190& 31& 6.2\\
\texttt{H-B-SN-D}& 607& 128& 5.0\\
\texttt{H-B-SU-N}& 194& 40& 5\\
\texttt{H-B-R-N}& 193& 38& 5.0\\
\texttt{H-B-SN-N}& 188& 39& 5.0\\
\texttt{H-BS-SU-D}& 2386& 314& 2-8\\
\texttt{H-BS-R-D}& 2409& 288& 6.0-10.0\\
\texttt{H-BS-SN-D}& 2324& 362& 5.0-8.0\\
\texttt{H-BS-SU-N}& 2319& 349& 5.0-8.0\\
\texttt{H-BS-R-N}& 2352& 320& 6.0-9.0\\
\texttt{H-BS-SN-N}& 2337& 347& 5.0-8.0\\
\texttt{H-BS-R-N2}& 1476& 130& 12\\
\texttt{total}& 18476& 4115& -\\\hline
\end{tabular}\\\vspace{1pt}\raggedright\footnotesize{\hspace{0em} L: low speed, H: high speed (10-30km/h), B: bumpy road, NB: non bumpy road, BS: bumps, SU: sunny, R: rainy, SN: snowy, D: day, N: night.} 
\end{table} 
To address the lack of open-source solutions for extrinsic calibration between LiDAR and 4D millimeter-wave radar, we propose a targetless calibration method based on an improved Iterative Closest Point (ICP) algorithm. The method first achieves millisecond-level temporal synchronization via timestamp alignment and interpolation compensation, followed by point cloud preprocessing using voxel filtering and statistical outlier removal. Guided by initial extrinsic parameters, a multistage ICP optimization strategy is developed: it starts with coarse registration using larger matching thresholds, then progressively refines alignment with tighter thresholds. The method incorporates point-to-plane error metrics and Huber loss functions to enhance robustness. The entire calibration process can be completed within 5 minutes. Experimental results demonstrate a calibration accuracy of 0.142 meters RMSE, surpassing the 4D radar's inherent error limit of 0.15 meters. The reliability of the calibration is further validated through the alignment of characteristic features such as building contours and road edges, meeting the precision requirements for multi-sensor fusion in autonomous driving systems. The calibration results are shown in Fig. 4.

\begin{figure}[t!]
\setlength{\fboxsep}{0pt}%
\setlength{\fboxrule}{0pt}%
\begin{center}
    \includegraphics[width=0.65\linewidth]{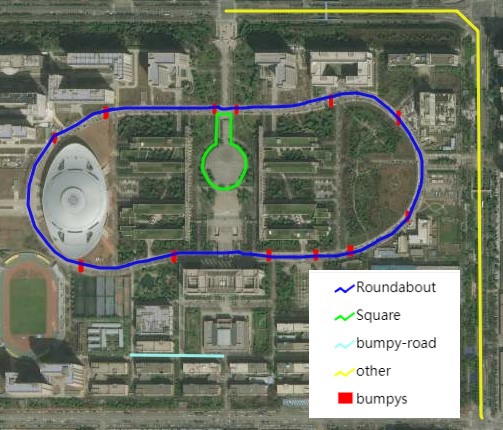}
\end{center}\vspace{-12pt}
\caption{Data Trajectory Top View.}\vspace{-12pt}
\end{figure}
\subsection{Time synchronization }
We use a unified system clock source, combining hardware synchronization and software-based temporal calibration to align all sensors with a 200 Hz IMU.

We use the Precision Time Protocol (PTP) to synchronize the LiDAR with the system time. The LiDAR and IMU are hardware-synchronized with a timing error better than 1 millisecond.

The four types of image data(rgb, left, right and depth) from the OAK-D achieve time synchronization through their internal hardware synchronization mechanisms. We employ the VINS-Mono\cite{vin} to calibrate the inherent time delay between the OAK-D images and the IMU, and the time delay is approximately 30 ms.

Similarly, we calibrate the time offset between the infrared images and the IMU for each sequence. However, due to the loss of feature points in infrared images in scenarios with blurred temperature boundaries, we also conduct time offset calibration in the laboratory. We collect ten sets of data and take the average value of 140 ms as the laboratory time offset calibration value, with a root mean square error (RMSE) of 6.455 ms.

We employ the iKalibr\cite{ika} to calibrate the timing offset for the 4D millimeter-wave radar and the IMU, and the time delay is approximately 20 ms.

We calculate the time offset between the GPS/INS and the IMU for each sequence in our dataset using ikalibr. The time offset is approximately 50 ms.

After compensating for the time offsets between each sensor and the IMU, the inter-sensor time offsets fall within the range of 0–25 ms based on the sampling frequency.

We test Fast-LIO2\cite{fastlio2}, LIO-SAM\cite{liosam}, VINS-Mono\cite{vin}, FAST-LIVO\cite{fastlivo}, FAST-LIVO2\cite{fastlivo2}, and R3live\cite{r3live} on our dataset. The above algorithms can be successfully executed on our dataset. 

\begin{figure}[t!]
\setlength{\fboxsep}{0pt}%
\setlength{\fboxrule}{0pt}%
\begin{center}
    \includegraphics[width=0.475\linewidth]{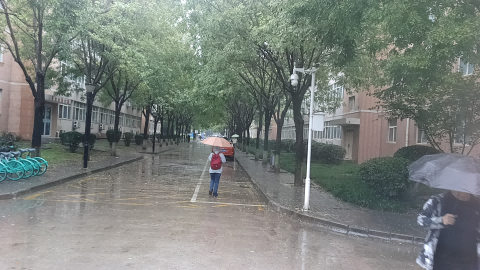}\includegraphics[width=0.475\linewidth]{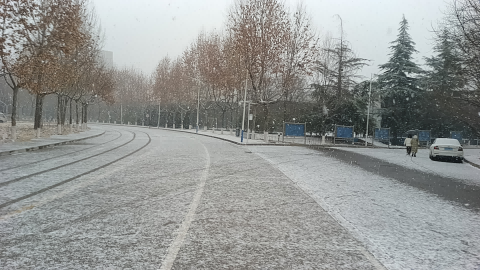}
    \includegraphics[width=0.475\linewidth]{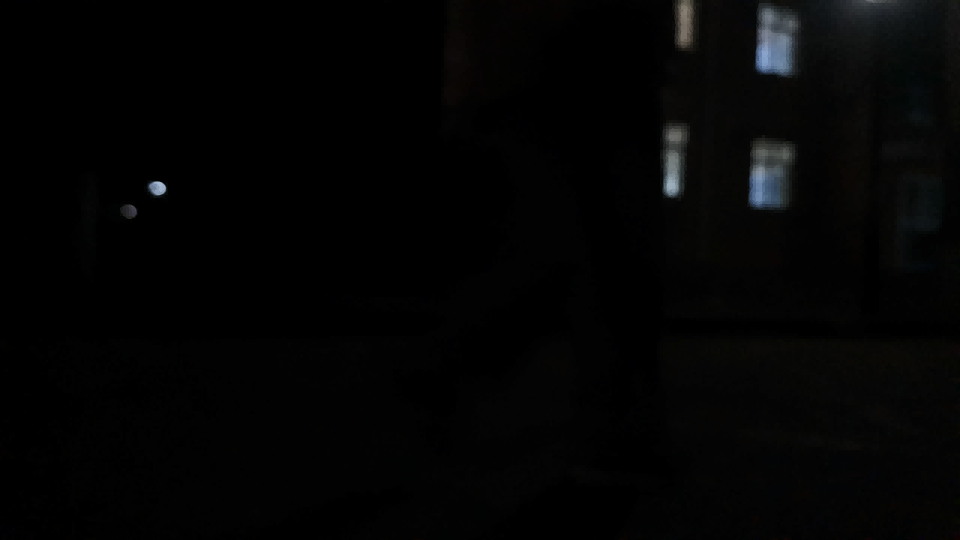}\includegraphics[width=0.475\linewidth]{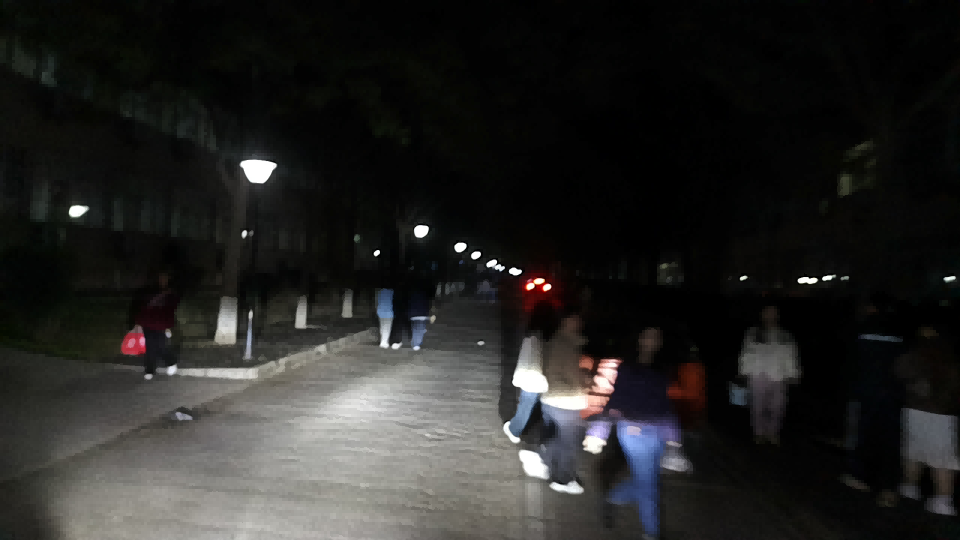}
    \includegraphics[width=0.475\linewidth]{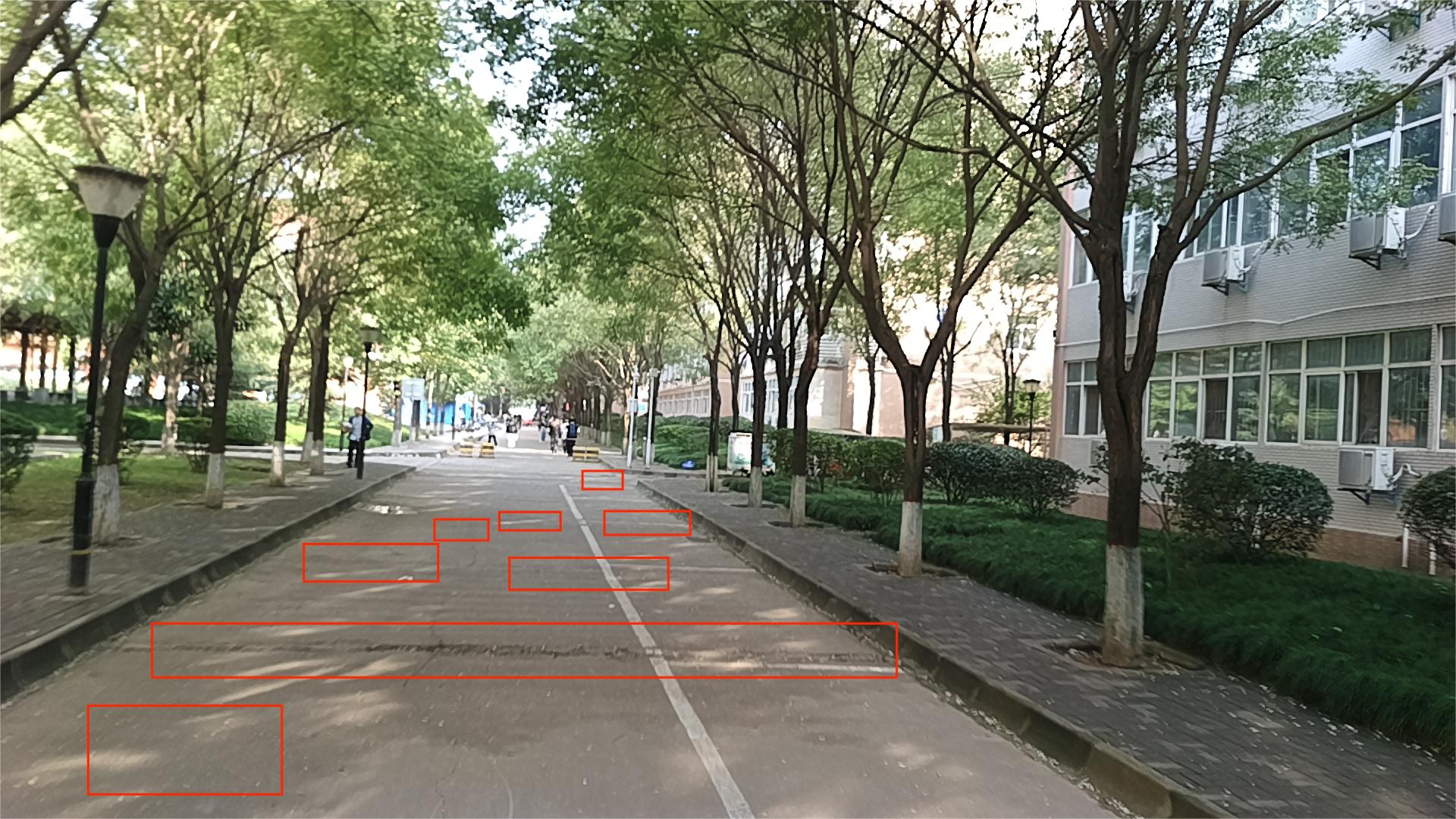}\includegraphics[width=0.475\linewidth]{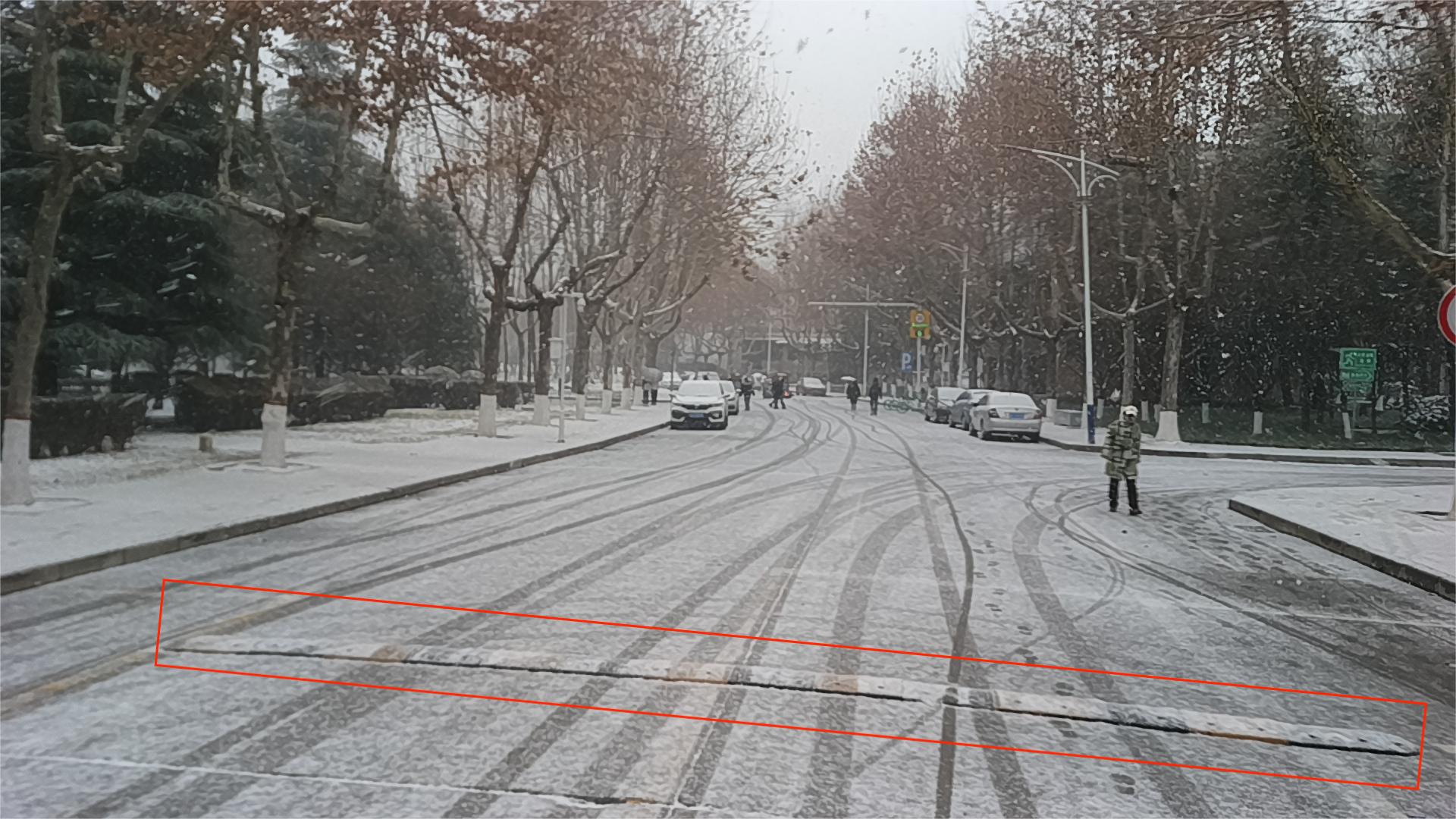}
\end{center}\vspace{-12pt}
\caption{The figure illustrates a variety of adverse environmental scenarios, including rainy and snowy conditions, nighttime with and without headlights, continuous bumpy roads, and roads with multiple speed bumps.}\vspace{-12pt}
\end{figure}
\subsection{GPS/INS }
We use the DETA100R GPS/INS system, which comprises a high-precision GPS and a 9-axis IMU. The system employs the Unscented Kalman Filter (UKF) for data processing. This setup is capable of sensitively reflecting changes in position and orientation, thereby significantly enhancing positioning accuracy.

\subsection{Data Collection }

To ensure the diversity of our dataset, we employ both low-speed and high-speed data collection methods. Our planned routes include plazas, bumpy roads, campus loops, and areas outside the campus. Most of these routes are covered under three weather conditions: sunny, rainy, and snowy, as well as two lighting conditions: daytime and nighttime. Our dataset includes continuous flat roads, continuous bumpy roads, and roads with multiple speed bumps. In summary, our dataset is deliberately designed with varying speeds, different road conditions, diverse weather, and lighting conditions, and features a rich set of sensors. The detail can be seen in Fig. 5 and 6.

We record approximately 1500 GB of image and point cloud data, which is then converted to about 660 GB in .rosbag format and upload to the data website. It should be noted that conditions such as low light, strong turbulence, rain, snow, and others can affect the imaging quality of cameras. We have compiled a list of the imaging quality for each sequence's RGB, left camera, right camera, and infrared camera, which is available on the data download website. Users can select camera topics with good imaging quality for algorithm testing based on the provided list. Overall data information is provided in Table IV.

\section{Evaluation}
In this section, we present the origin and mechanism of the algorithms we used, and analyze how different sensors and algorithms may affect SLAM through ATE and RPE data. The specific numerical values are presented in Tables V and VI. Due to space limits, trajectory plots and related experimental analyses are in the appendix.
\begin{figure}[htbp]
\setlength{\fboxsep}{0pt}%
\setlength{\fboxrule}{0pt}%
\begin{center}
    \includegraphics[width=4cm,height=2.5cm]{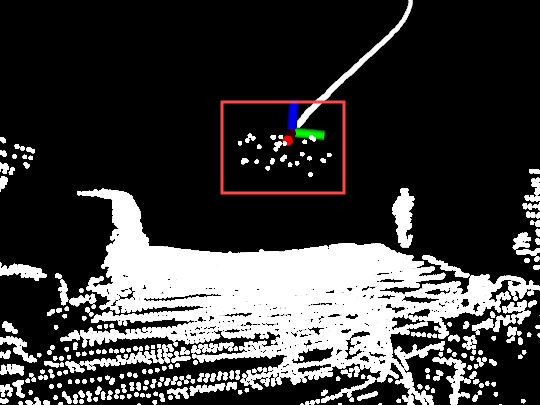}\includegraphics[width=4cm,height=2.5cm]{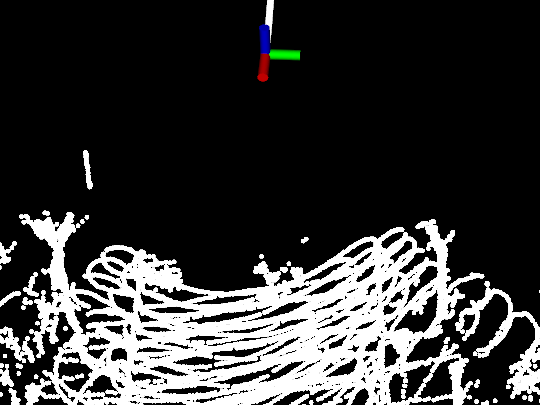}
    \includegraphics[width=4cm,height=2.5cm]{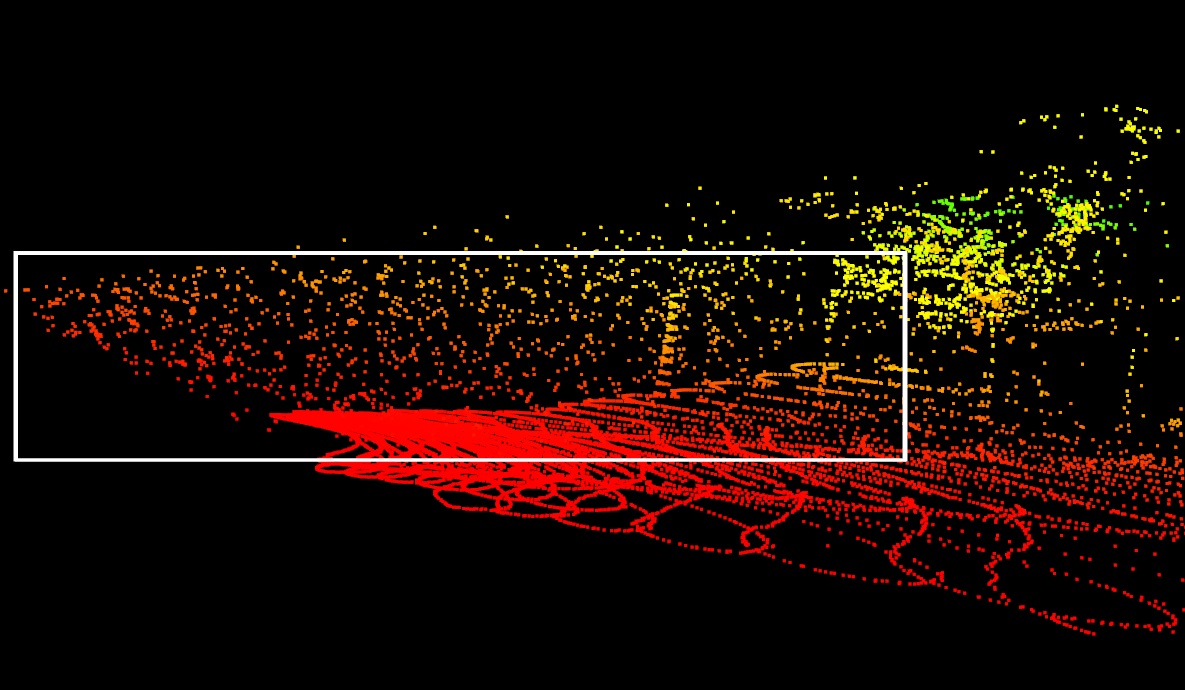}\includegraphics[width=4cm,height=2.5cm]{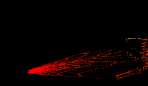}
\end{center}\vspace{-13pt}
\caption{The left image shows a LiDAR scan in a snowy scene with a large number of snowflakes obstructing the lens. The right images are comparisons with sunny weather, respectively.}\vspace{-13pt}
\end{figure}

\subsection{Lidar SLAM}
LOAM-Livox (2020) \cite{livoxslam} represents an early open-source SLAM solution optimized for Livox LiDARs, supporting both pure LiDAR and LiDAR-IMU fusion configurations. CT-ICP (2022) \cite{cticp} enhances this framework by integrating loop closure functionality. Our comparative experiments with these algorithms under identical conditions demonstrate two consistent observations: (i) single-LiDAR localization accuracy degrades more severely in snowfall compared to clear/rainy weather, and (ii) speed bump traversals induce measurable position drift even with loop closure activated.

FAST-LIO2 (2021) \cite{fastlio2} is a LiDAR-IMU fusion SLAM algorithm based on a tightly-coupled iterative Kalman filter.
LIO-SAM (2020) \cite{liosam} achieves tight coupling of LiDAR and IMU through factor graph optimization, supporting various relative and absolute measurements (including loop closures) as factors in the system. It also utilizes IMU preintegration to estimate motion, undistorts point clouds, and provides initial guesses for LiDAR odometry optimization. Through experiments with these two algorithms, compared to the previous pure LiDAR SLAM, the SLAM systems with tightly-coupled IMU information show significantly stronger robustness against speed bumps. However, according to the trajectory plots, there are still noticeable fluctuations when passing over speed bumps. And the algorithm with sensor fusion shows a significant improvement in positioning accuracy in snowy conditions. The detail can be seen in Fig. 7.
\begin{figure}[htbp]
\setlength{\fboxsep}{0pt}%
\setlength{\fboxrule}{0pt}%
\begin{center}
    \includegraphics[width=4cm,height=2.5cm]{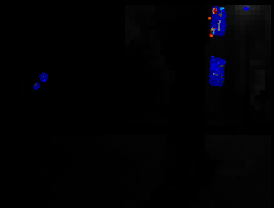}\includegraphics[width=4cm,height=2.5cm]{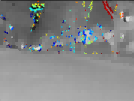}
\end{center}\vspace{-15pt}
\caption{The left and right figures show DSO-SLAM performance under low-speed-bumpy-sunny-night  scenarios using RGB and infrared images, respectively. It is evident that infrared images generally provide better visibility and richer point clouds than RGB images at night.}\vspace{-15pt}
\end{figure}

\subsection{Visual/Infrared SLAM}
ORB-SLAM3 (2021) \cite{orb} is known for its high accuracy and robustness. However, when running our data with a monocular camera, keyframe loss frequently occurs. This issue is particularly prominent when traversing speed bumps or traveling on bumpy roads. Additionally, in low-light conditions, vehicle headlights are often mistakenly detected as feature points, leading to feature point loss.
\begin{table*}[h!]\vspace{-10pt}
\small\sf\centering
\caption{Different SLAM algorithms exhibit varying Absolute Trajectory Errors (ATE)  (unit: meters). }\vspace{-5pt}
\renewcommand{\arraystretch}{1.1}\begin{tabular}{lccccccccccccccccccccc}
\hline
Sequence &      {Livox-}&{4DRT-}& {CT-}& {ORB-}&{FAST-}& {LIO-}& {VINS-}& {FAST-} & {FAST-} &   {R3}&{GS-}\\
 &      {SLAM}&{SLAM}& {ICP}& {SLAM3}&{LIO2}& {SAM}& {MONO}& {LIVO}& {LIVO2}& {live}& {SLAM}\\\hline
 sensors&      {3D}&{4D }& {3D}& {mono}&{LiDAR}& {LiDAR}& {RGB}& {LiDAR}& {LiDAR}&   {LiDAR}&{RGB}\\
 & {LiDAR}& {Radar}& {LiDAR}& {RGB}& {IMU}& {IMU}& {IMU}& {IMU RGB}& {IMU RGB}& {IMU RGB}& {Depth}\\
\hline
\texttt{L-NB-SU-D}&      {1.20}&{11.72}& {1.66}& {1.81}&{\textbf{1.13}}& {1.37}& {14.98}& {2.01}& {2.37}&   {3.79}&{38.80}\\
\texttt{L-NB-SU-N}&      {3.13}&{12.07}& {3.29}& {x}&{\textbf{2.13}}& {3.62}& {x }& {2.39}& {2.19}&   {2.19}&{42.23}\\
\texttt{L-B-SU-D}&      {2.14}&{6.95}& {1.51}& {x}&{2.09}& {1.93}& {4.96}& {\textbf{1.86}}& {2.12}&   {2.42}&{32.41}\\
\texttt{L-B-SU-N}&      {2.28}&{\textbf{0.47}}& {2.22}& {x}&{2.35}& {2.30}& {x}& {2.26}& {2.34}&   {2.31}&{37.79}\\
\texttt{H-BS-SU-D}&      {33.77}&{45.84}& {126.48}& {46.72}&{16.95}& {18.63}& {21.88}& {14.75}& {\textbf{11.59}}&   {13.95}&{249.36}\\
\texttt{H-BS-R-D}&      {24.74}&{24.87}& {24.94}& {x}&{13.17}& {57.49}& {x}& {23.09}& {\textbf{9.96}}&   {213.85}&{272.21}\\
\texttt{H-BS-SN-D}&      {44.71}&{76.16}& {55.85}& {39.32}&{\textbf{20.95}}& {33.08}& {59.57}& {36.32}& {21.09}&   {21.88}&{325.10}\\
\texttt{H-BS-SN-N}&      {39.44}&{36.67}& {68.02}& {287.82}&{27.65}& {28.16}& {89.06}& {28.19}& {\textbf{25.03}}&   {26.67}&{322.11}\\\hline
\end{tabular}\vspace{1pt}\renewcommand{\arraystretch}{1}\\\raggedright\footnotesize{\hspace{1em}x denotes failed experiments. }\vspace{-5pt}
\end{table*}

VINS-Mono (2018) \cite{vin} is a highly robust algorithm with camera-IMU extrinsic calibration and IMU bias correction. When running our data, frame loss due to bumps no longer occurs. Compared to monocular SLAM, the positioning is more accurate. However, in heavy snow conditions, enabling real-time calibration mode results in larger errors. We analyzed that this is because the snowflakes in snowy weather affect the normal feature point selection, causing the extrinsic parameters undergoing real-time calibration to have larger errors, which in turn affect the positioning accuracy.

Both of these algorithms are based on RGB images for SLAM and perform poorly in night scenes, being unable to extract a sufficient number of feature points in completely dark environments. However, the infrared images are clearly visible and rich in feature points as shown in Fig. 8. DSO\cite{dso} is a direct method SLAM. It can be used to better locate feature points for infrared SLAM. However, due to the significant differences between infrared and RGB images, DSO still cannot effectively complete infrared SLAM. Therefore, we do not calculate its ATE and RPE.
\begin{figure}[htbp]
\begin{center}\vspace{-10pt}
    \includegraphics[width=9cm]{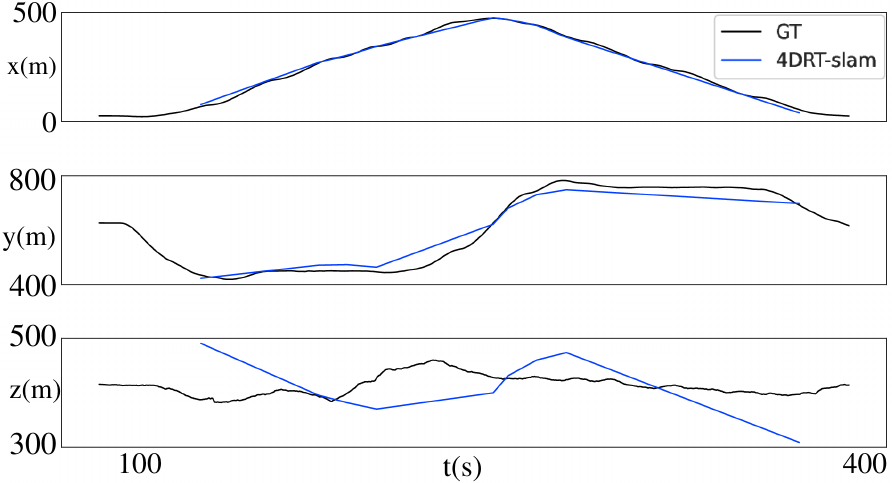}
    \includegraphics[width=9cm]{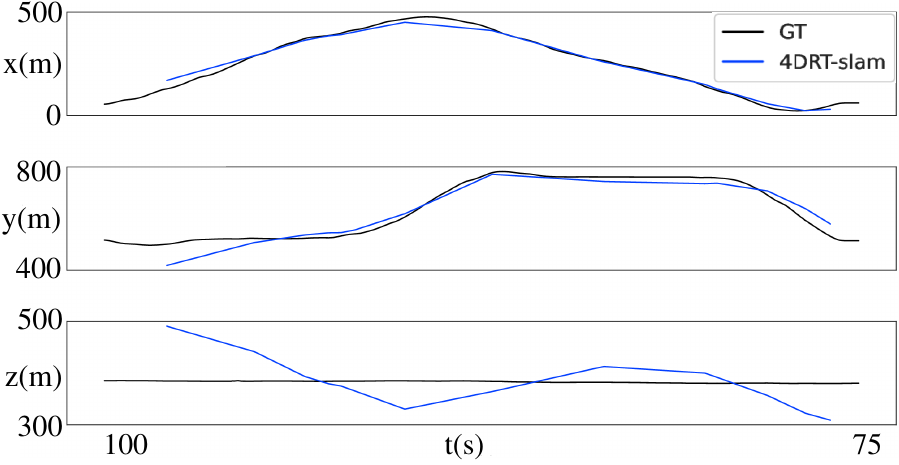}
\end{center}
\caption{The figure presents the variation of x, y, and z coordinates over time in two closed-loop scenarios. The blue line represents the 4D millimeter-wave SLAM, while the gray line represents the ground truth (GT).}
\end{figure}

\subsection{4D millimeter-wave radar SLAM}
We evaluate our data using the 4DRadarSLAM\cite{4dslam} algorithm introduced in 2023. As we can see from Tables 5 and 6, pure 4D millimeter-wave radar SLAM exhibits inferior performance compared to both LiDAR SLAM and the LiDAR+IMU+RGB SLAM framework under normal weather conditions (sunny day) and even some adverse weather conditions. However, under adverse weather conditions (snowy day, rainy day, etc ), 4D millimeter-wave radar can assist LiDAR and RGB to achieve enhanced performance, and the reason is given as follows. Under normal weather conditions, LiDAR and RGB images achieve excellent mapping performance due to their dense point clouds and rich semantic information. However, in adverse weather conditions(snowy day, rainy day, etc), LiDAR point clouds are severely affected by snowflakes(or raindrops) due to their limited penetration, while RGB images exhibit pronounced snow-induced noise. In contrast, although 4D millimeter-wave radar point clouds are comparatively sparse, they demonstrate superior penetration. Therefore, integrating 4D millimeter-wave radar with LiDAR and RGB for SLAM effectively mitigates snowflake-induced noise in both LiDAR and images, thereby enhancing SLAM accuracy. Furthermore, in loop-closure scenarios, SLAM systems relying on sparse millimeter-wave point clouds consistently exhibit particularly large vertical errors, as Fig. 9 clearly shows. 

\begin{figure}
\begin{center}
    \includegraphics[width=4cm,height=2.5cm]{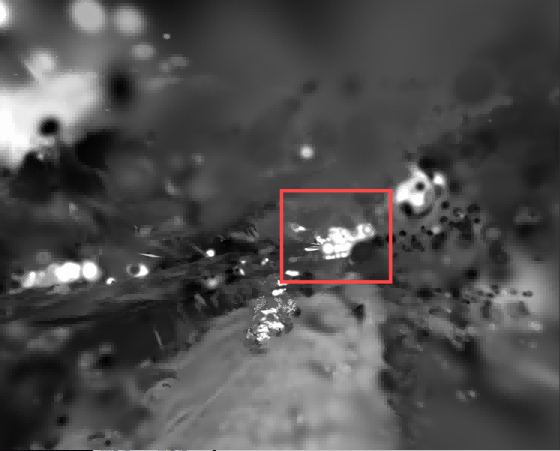}\includegraphics[width=4cm,height=2.5cm]{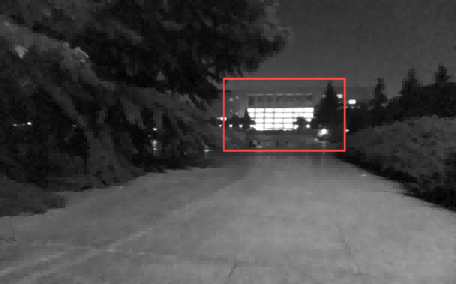}
    \includegraphics[width=4cm,height=2.5cm]{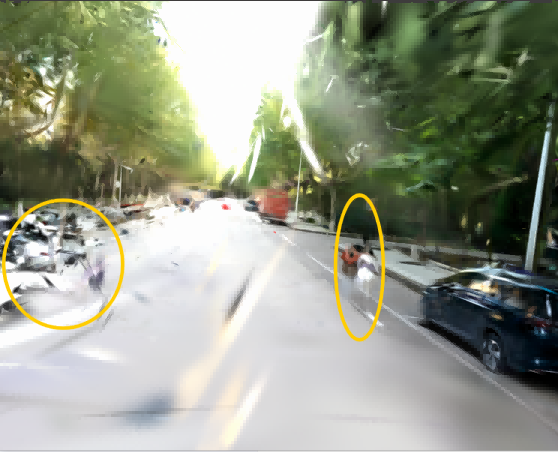}\includegraphics[width=4cm,height=2.5cm]{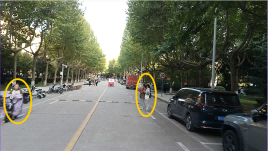}
    \includegraphics[width=4cm,height=2.5cm]{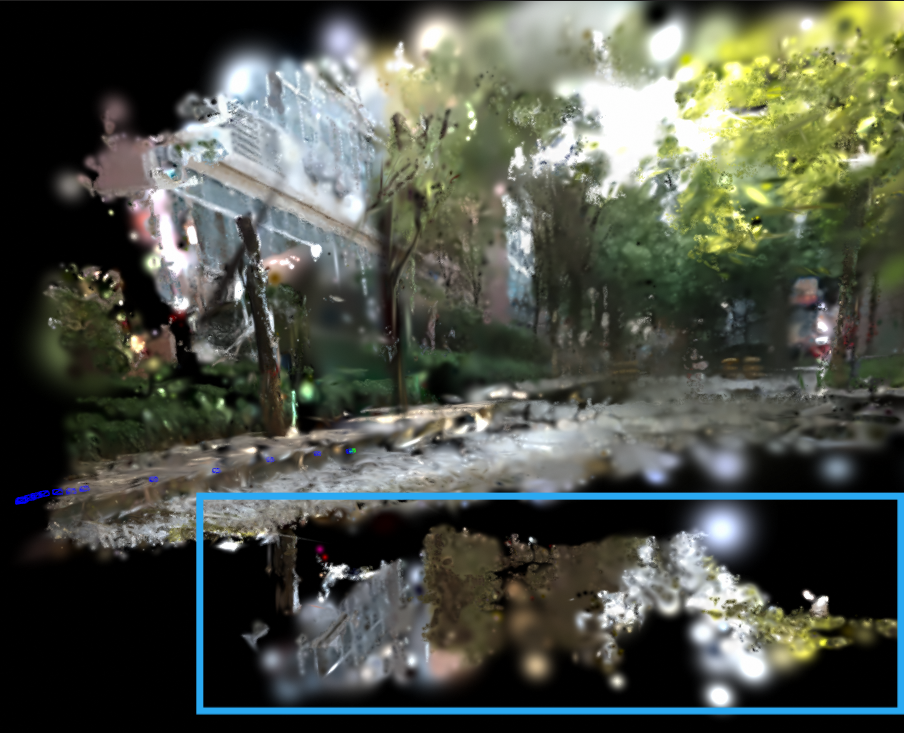}\includegraphics[width=4cm,height=2.5cm]{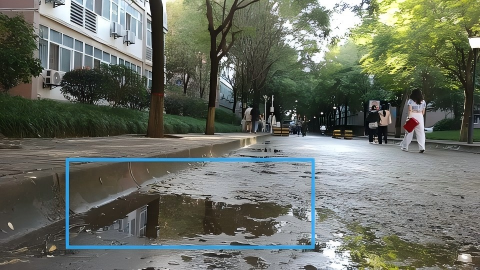}
\end{center}\vspace{-10pt}
\caption{On the left is a screenshot of the mapping result provided by monoGS, and on the right is the actual map. The landmark building (the library) is outlined in red, dynamic pedestrians are highlighted in yellow, and water accumulation is indicated in blue.}
\end{figure}

\subsection{Gaussian splatting SLAM}

\begin{table*}[h!]
\small\sf\centering
\caption{Different SLAM algorithms exhibit varying  Relative Pose Error (RPE) (unit: meters). }\vspace{-5pt}
\renewcommand{\arraystretch}{1.1}\begin{tabular}{lcccccccccccccccccccc}
\hline
Sequence &      {Livox-}&{4DRT-}& {CT-}& {ORB-}&{FAST-}& {LIO-}& {VINS-}& {FAST-} & {FAST-} &   {R3}\\
 &      {SLAM}&{SLAM}& {ICP}& {SLAM3}&{LIO2}& {SAM}& {MONO}& {LIVO}& {LIVO2}& {live}\\\hline
 sensors&      {3D}&{4D }& {3D}& {mono}&{LiDAR}& {LiDAR}& {RGB}& {LiDAR}& {LiDAR}&   {LiDAR}\\
 & {LiDAR}& {Radar}& {LiDAR}& {RGB}& {IMU}& {IMU}& {IMU}& {IMU RGB}& {IMU RGB}& {IMU RGB}\\
\hline
\texttt{L-NB-SU-D}&      {1.35}&{9.38}& {0.90}& {3.19}&{0.80}& {0.83}& {1.15}& {\textbf{0.18}}& {0.55}&   {0.81}\\
\texttt{L-NB-SU-N}&      {1.93}&{46.73}& {2.42}& {x }&{1.21}& {1.33}& {x }& {\textbf{0.27}}& {0.94}&   {1.50}\\
\texttt{L-B-SU-D}&      {1.10}&{18.87}& {0.66}& {x }&{0.73}& {0.82}& {0.67}& {\textbf{0.16}}& {0.48}&   {0.89}\\
\texttt{L-B-SU-N}&      {1.15}&{118.67}& {0.82}& {x}&{0.90}& {0.80}& {x}& {\textbf{0.16}}& {0.48}&   {0.74}\\
\texttt{H-BS-SU-D}&      {16.67}&{431.77}& {11.31}& {12.96}&{11.17}& {13.09}& {9.99}& {\textbf{2.24}}& {7.23}&   {10.28}\\
\texttt{H-BS-R-D}&      {18.67}&{334.37}& {11.94}& {x}&{11.54}& {12.60}& {x}& {\textbf{2.44}}& {7.57}&   {10.13}\\
\texttt{H-BS-SN-D}&      {12.68}&{241.33}& {10.66}& {16.45}&{12.30}& {13.78}& {10.41}& {\textbf{1.70}}& {5.21}&   {7.74}\\
\texttt{H-BS-SN-N}&      {15.01}&{295.18}& {9.25}& {17.97}&{9.51}& {10.96}& {9.30}& {\textbf{1.92}}& {5.63}&   {9.22}\\\hline
\end{tabular}\vspace{1pt}\renewcommand{\arraystretch}{1}\\\raggedright\footnotesize{\hspace{2em}x denotes failed experiments.}\vspace{-5pt}
\end{table*}

\begin{figure*}[t!]
\begin{center}
     \includegraphics[width=8cm,height=5cm]{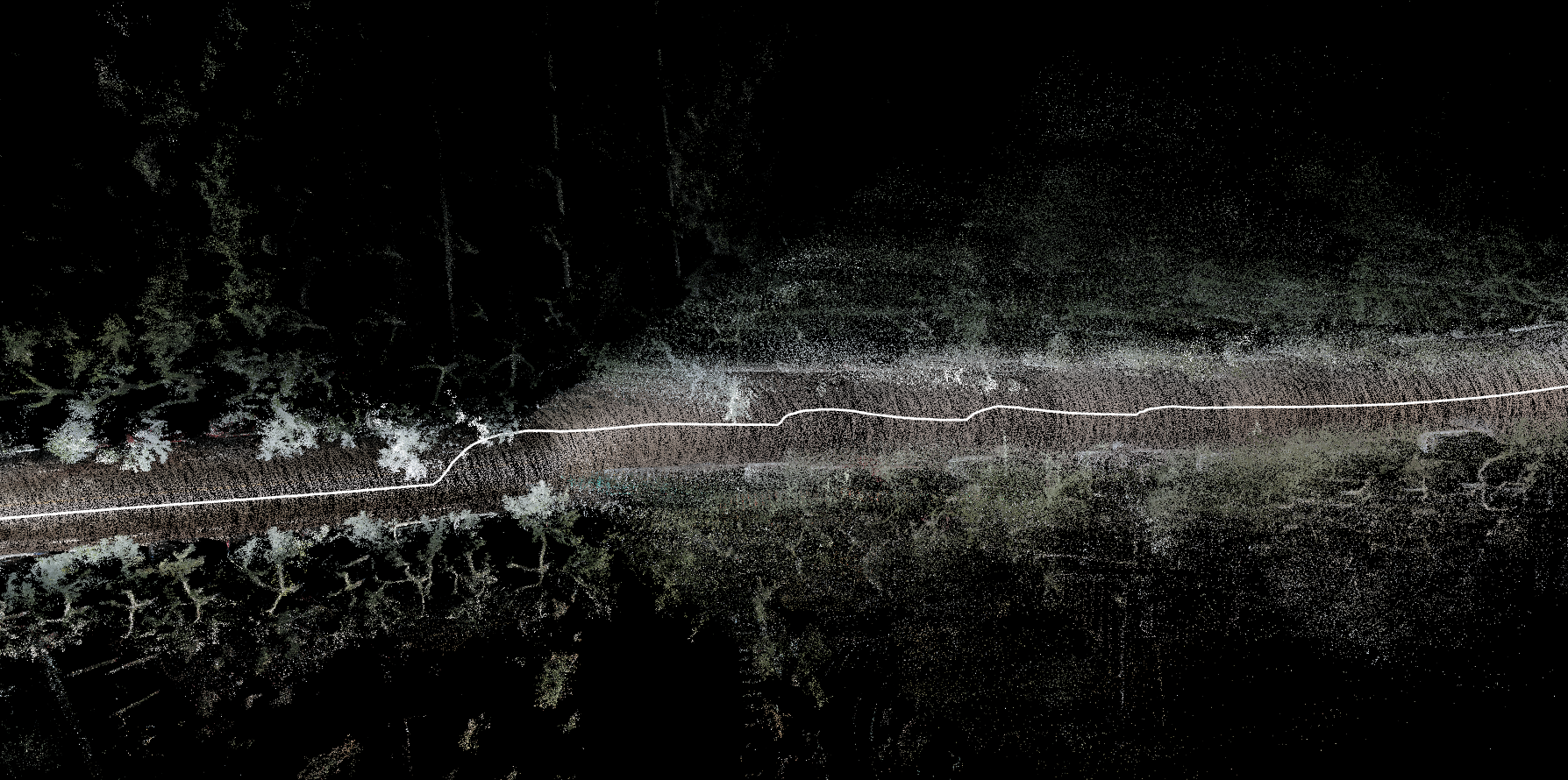}\includegraphics[width=8cm,height=5cm]{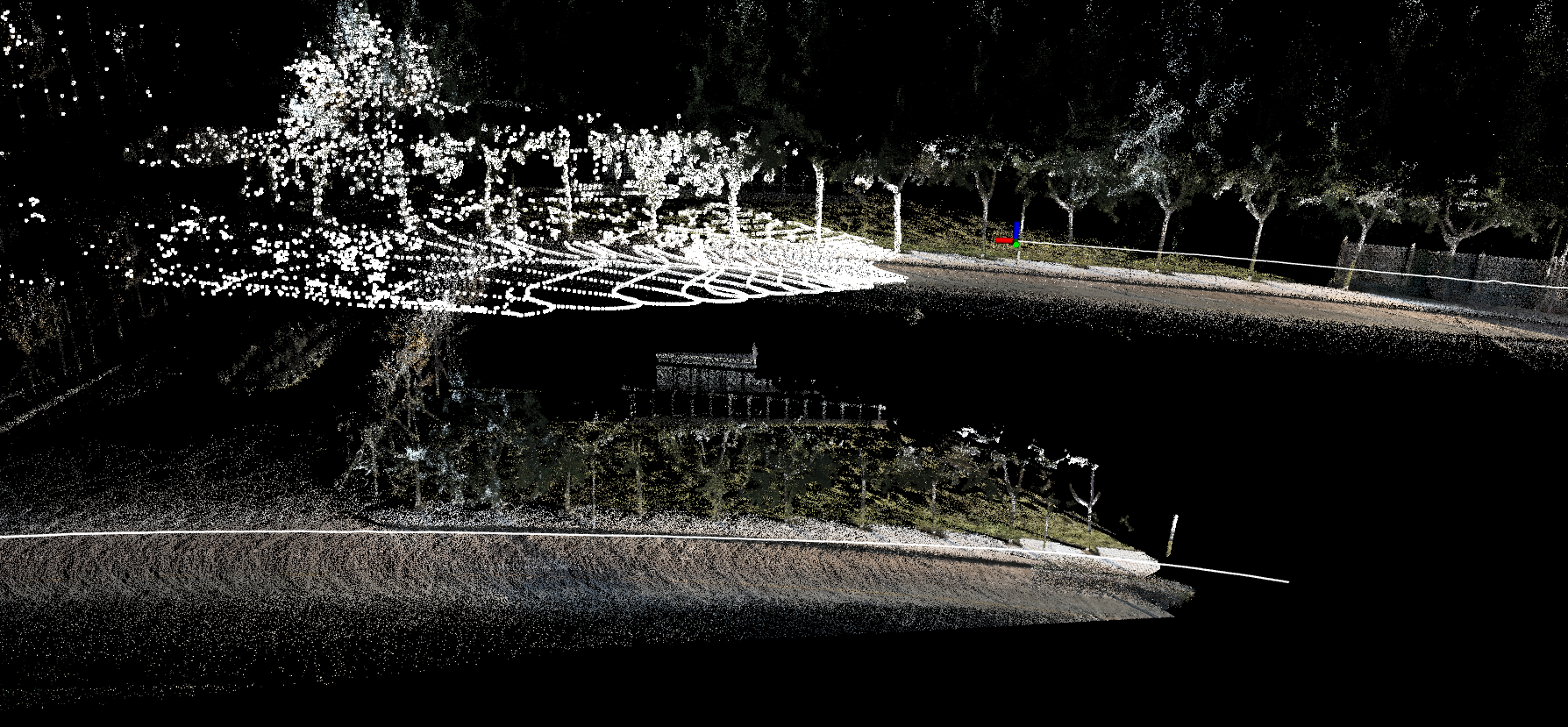}\\[5pt]
  \hspace{0.5em}(a)  \hspace{20em} (b)  \\
     
    \includegraphics[width=8cm,height=5cm]{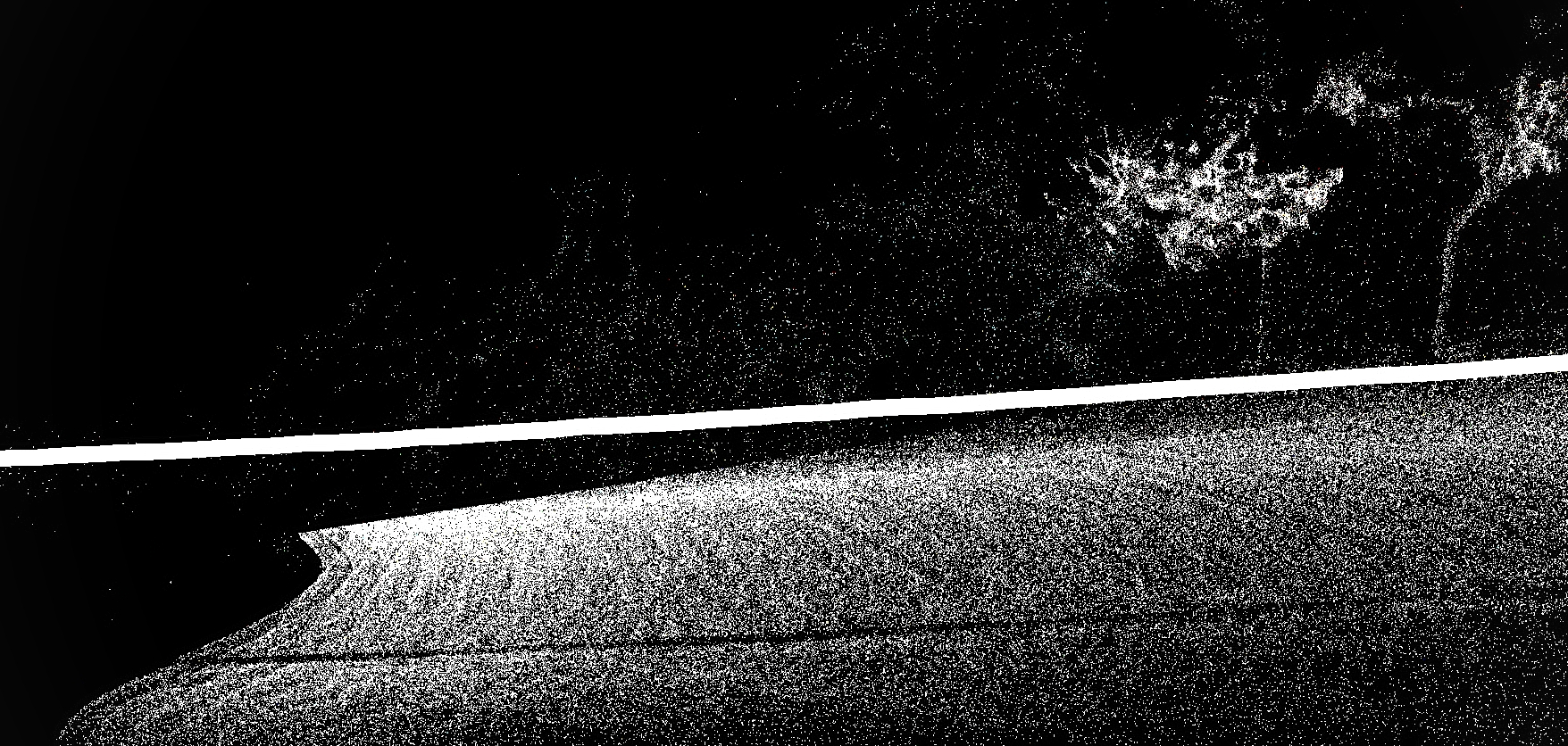}\includegraphics[width=8cm,height=5cm]{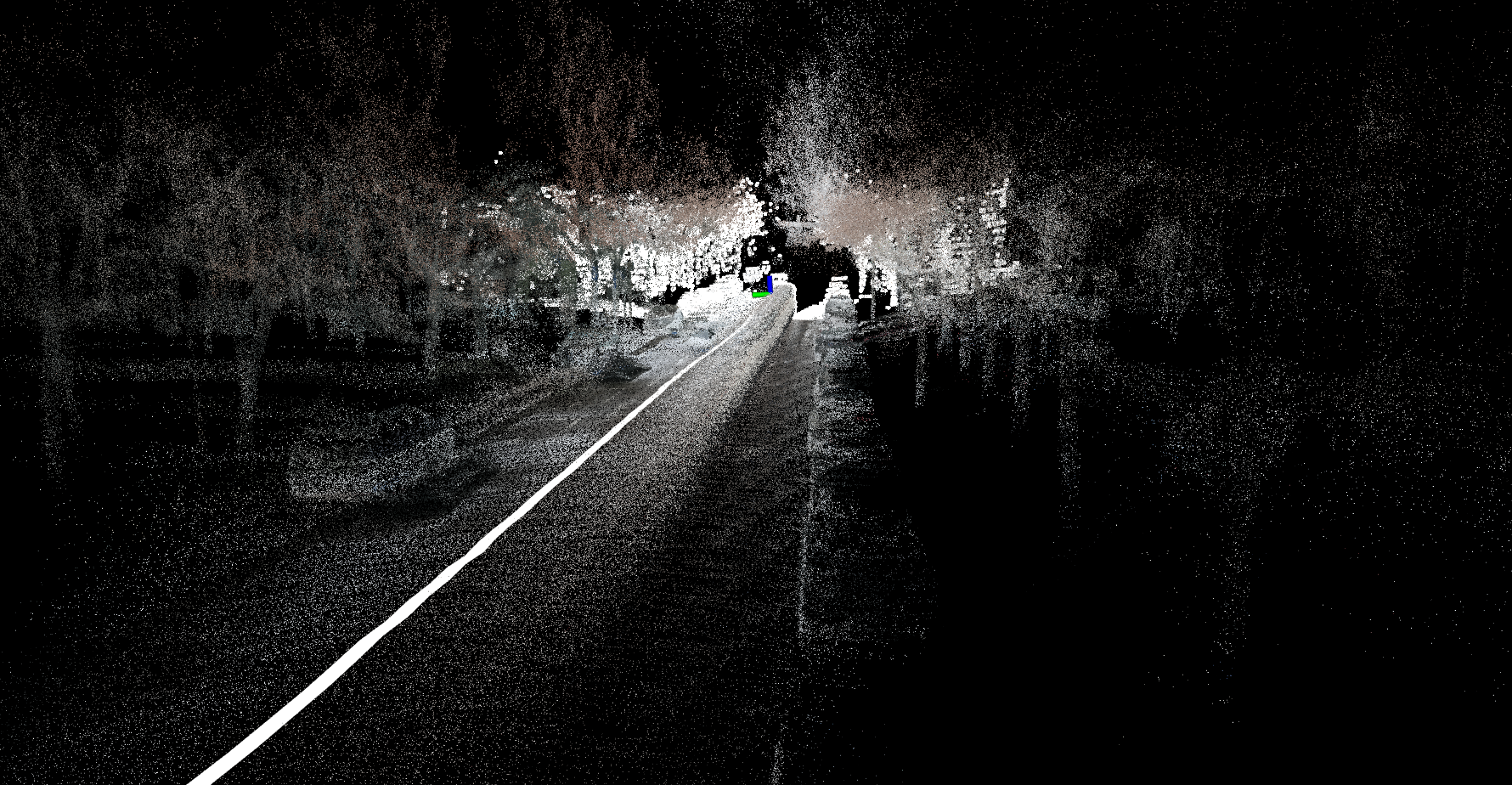}\\[5pt]
  \hspace{0.5em}(c)  \hspace{20em} (d)  
\end{center}\vspace{-10pt}
\caption{The figure illustrates the mapping issues in R3live: (a) Trajectory drift in rainy conditions,  
(b) vertical height deviation between initial and final points in sunny conditions,  
(c) noise from snowflakes in snowy conditions,  
(d) erroneous ground point clouds caused by interference from the preceding vehicle.}\vspace{-5pt}
\end{figure*}
Gaussian Splatting SLAM (2024)\cite{gs} utilizes 3D Gaussian splatting for map representation and optimization, effectively handling dynamic scenes and complex environments. It demonstrates good mapping results. We test its mapping and localization accuracy using RGB images and depth images. While it shows promise in many scenarios, there are still issues in mapping dynamic objects, as it cannot effectively remove them. Additionally, in scenes with reflective surfaces, such as water, MonoGS may produce erroneous mappings, placing the water surface below the ground. This is likely due to incorrect depth information from the depth images, combined with normal RGB coloring. When comparing the ATE, we found that MonoGS has larger localization errors in outdoor large-scale environments. In snowy conditions, the error is 100 meters greater than in sunny or rainy weather. This is probably due to the significant noise introduced in the depth images and the presence of numerous snowflakes in the RGB images. The system also falls short of real-time SLAM requirements, with computation times often exceeding ten times the data acquisition time. The detail can be seen in Fig. 10.

\subsection{Ldar\_imu\_Visual SLAM}
R3LIVE \cite{r3live} is proposed in 2022. It is a tightly-coupled LiDAR-inertial-visual fusion framework that leverages measurements from LiDAR, IMU, and visual sensors to achieve robust and accurate state estimation. FAST-LIVO (2022)\cite{fastlivo} introduces a novel outlier rejection method that discards unstable map points located on edges or occluded in the image view, thereby enhancing the robustness and accuracy of the visual-inertial odometry (VIO). FAST-LIVO2 (2024)\cite{fastlivo2} is an efficient LiDAR-inertial-visual fusion localization and mapping system that tightly integrates IMU, LiDAR, and image measurements through an error-state iterative Kalman filter (ESIKF).

\begin{table*}[htbp]
\small\sf\centering
\caption{Comparison of the average running time per frame for each algorithm (units: s)}
\label{tab:running_time_comparison}
\renewcommand{\arraystretch}{1.1}
\begin{tabular}{l|ccccccccc}\hline
& {Livox-SLAM} & {4DRT-SLAM} & {ORB-SLAM3} & {Vins-mono} & {LIO-SAM} & \multicolumn{2}{c}{R3live} & \multicolumn{2}{c}{Fast-livo2} \\
& {LO} & {RO} & {VO} & {VIO} & {LIO} & {LIO} & {VIO} & {LIO} & {VIO} \\\hline
\texttt{L-NB-SU-D} & 0.0246 & 0.0041 & 0.0245 & 0.0147 & 0.0679 & 0.0120 & 0.0664 & 0.0110 & 0.0039 \\
\texttt{L-NB-SU-N} & 0.0285 & 0.0062 & x & x & 0.0691 & 0.0108 & 0.0433 & 0.0119 & 0.0035 \\
\texttt{L-B-SU-D} & 0.0310 & 0.0067 & x & 0.0158 & 0.0632 & 0.0101 & 0.0643 & 0.0113 & 0.0034 \\
\texttt{L-B-SU-N} & 0.0307 & 0.0030 & x & x & 0.0606 & 0.0119 & 0.0277 & 0.0126 & 0.0036 \\
\texttt{H-BS-SU-D} & 0.0260 & 0.0086 & 0.0201 & 0.0195 & 0.1625 & 0.0242 & 0.0641 & 0.0162 & 0.0033 \\
\texttt{H-BS-R-D} & 0.0245 & 0.0065 & x & x & 0.0841 & 0.0267 & 0.0712 & 0.0164 & 0.0041 \\
\texttt{H-BS-SN-D} & 0.0231 & 0.0043 & 0.0225 & 0.0150 & 0.1235 & 0.0254 & 0.0440 & 0.0147 & 0.0028 \\
\texttt{H-BS-SN-N} & 0.0309 & 0.0039 & 0.0264 & 0.0171 & 0.1071 & 0.0145 & 0.0383 & 0.0150 & 0.0028 \\\hline
\end{tabular}
\vspace{1pt}\renewcommand{\arraystretch}{1}

\raggedright\footnotesize{x denotes failed experiments.}
\vspace{-5pt}
\end{table*}
These multi-sensor fusion methods offer high localization accuracy and robustness, as well as good intuitive mapping results. However, we identify some issues. In rainy conditions, R3LIVE's mapping is affected by erroneous feature points from the visual sensor due to rain interference. The system fails to recognize these errors and still incorporates images into the iterative calculations, impacting the less affected IMU and LiDAR in the rain. FAST-LIVO2 circumvents this issue by changing the operation mode of the visual module, achieving good localization and mapping accuracy. In snowy conditions, both LiDAR and images are inaccurate. The accuracy of all three algorithms is low, even comparable to FAST-LIO, which relies solely on LiDAR and IMU. This indicates that in snowy weather, the contribution of RGB images to system localization is limited, and further compensation using 4D millimeter-wave radar for LiDAR data may be necessary. We observe that FAST-LIVO performs excellently in terms of RPE, but it does not achieve the best results in ATE. Maybe this is because FAST-LIVO has strong robustness in local motion estimation, but there are errors in global optimization. How to improve global optimization and eliminate cumulative errors by utilizing multi-sensors is a direction that needs to be explored. The detail can be seen in Fig. 11.

\subsection{Computational overhead}
To quantify the computational overhead of the single-, dual-, and triple-sensor fusion SLAM algorithms, we measured the average processing times of six configurations: LiDAR-only SLAM, 4-D millimetre-wave radar-only SLAM, vision-only SLAM, LiDAR–IMU SLAM, vision–IMU SLAM, and LiDAR–IMU–vision SLAM.

In single-sensor configurations, the LiDAR-based and vision-based pipelines exhibited similar latencies. The 4-D millimetre-wave radar pipeline completed processing faster due to its sparser point cloud. The detailed results can be found in Table VII.
\begin{figure*}[t]
   \centering
       \includegraphics[width=0.5cm,height=1.4cm]{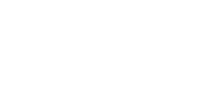}\includegraphics[width=2.23cm,height=1.2cm]{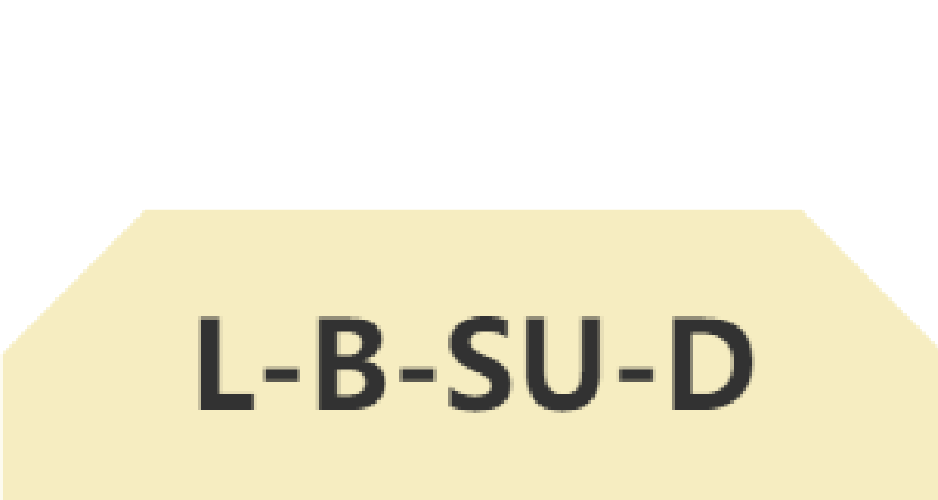}\includegraphics[width=2.23cm,height=1.2cm]{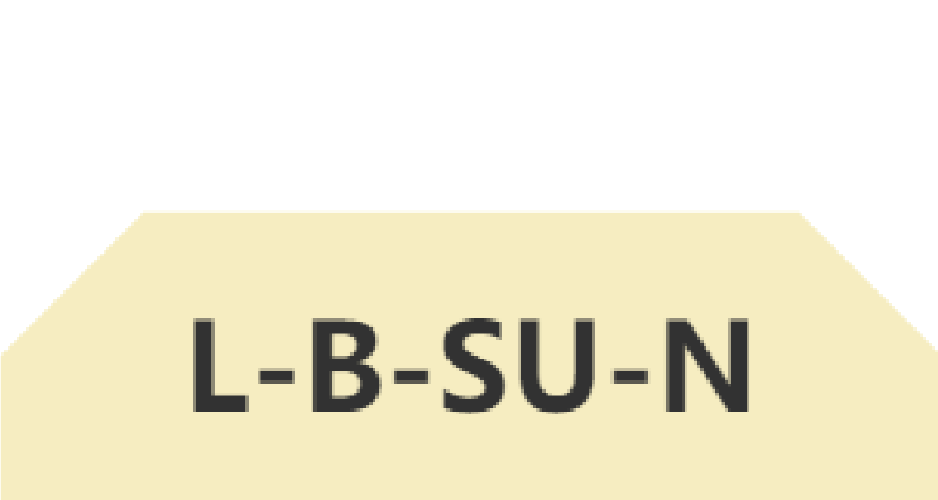}\includegraphics[width=2.23cm,height=1.2cm]{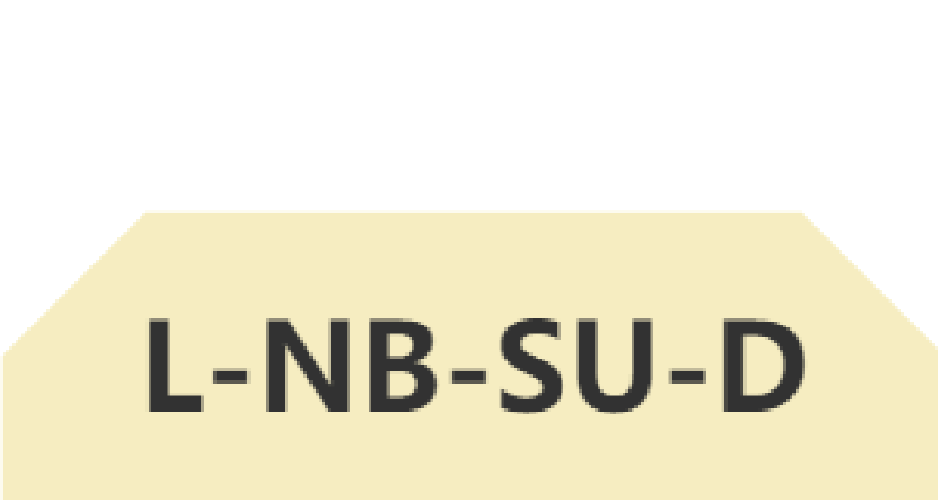}\includegraphics[width=2.23cm,height=1.2cm]{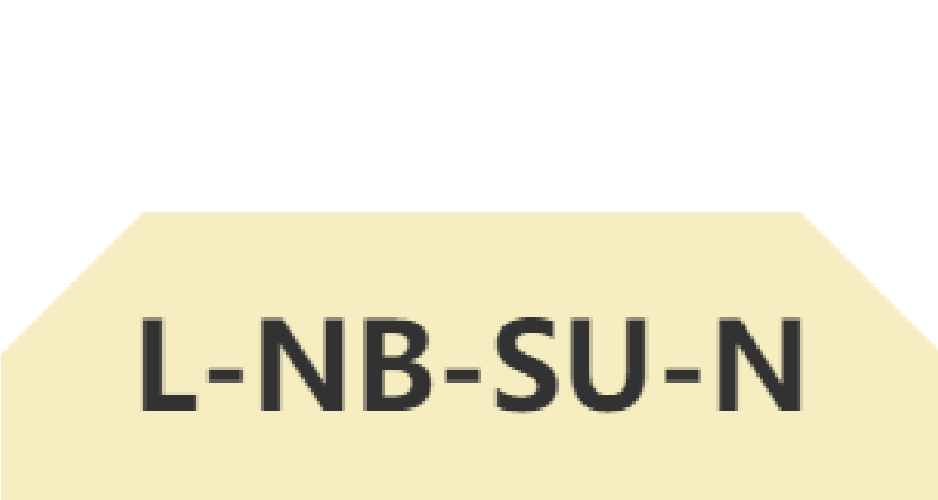}\includegraphics[width=2.23cm,height=1.2cm]{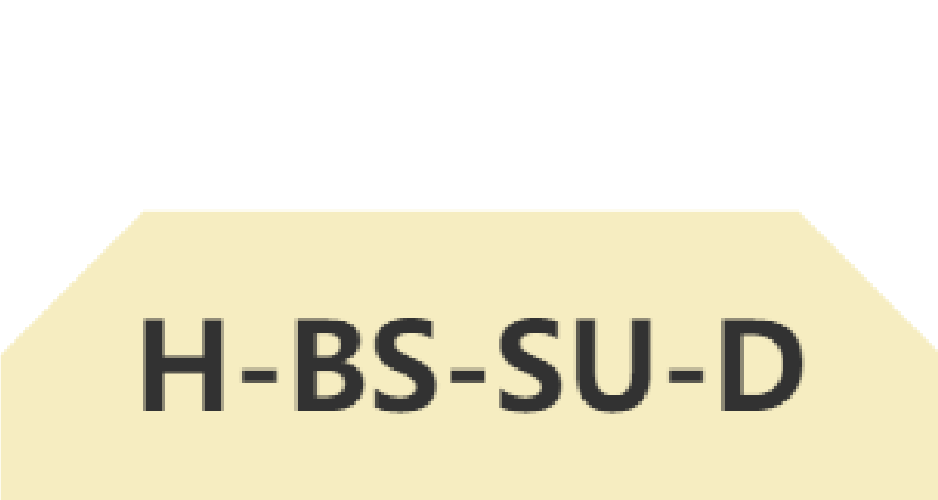}\includegraphics[width=2.23cm,height=1.2cm]{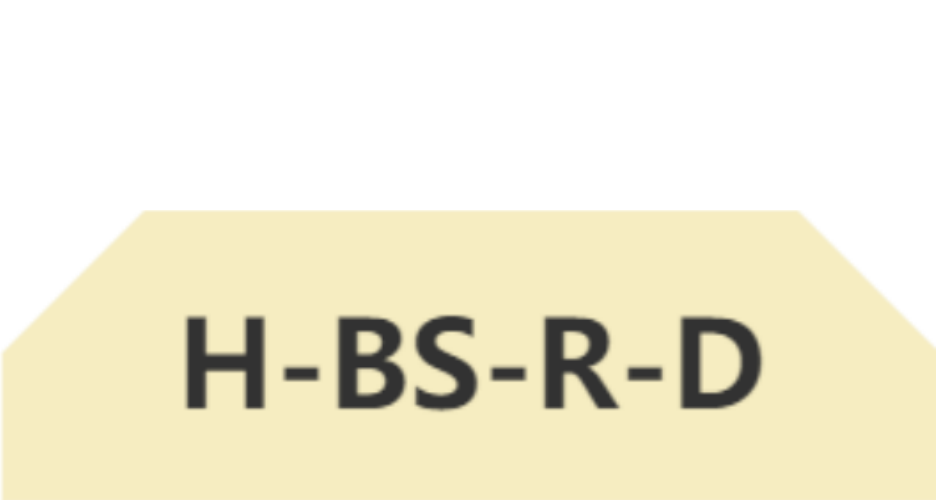}\includegraphics[width=2.23cm,height=1.2cm]{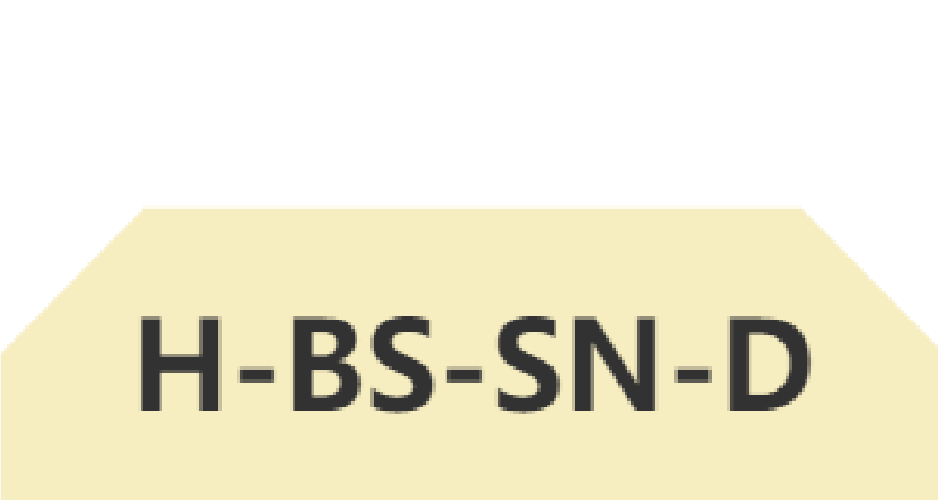}\includegraphics[width=2.23cm,height=1.2cm]{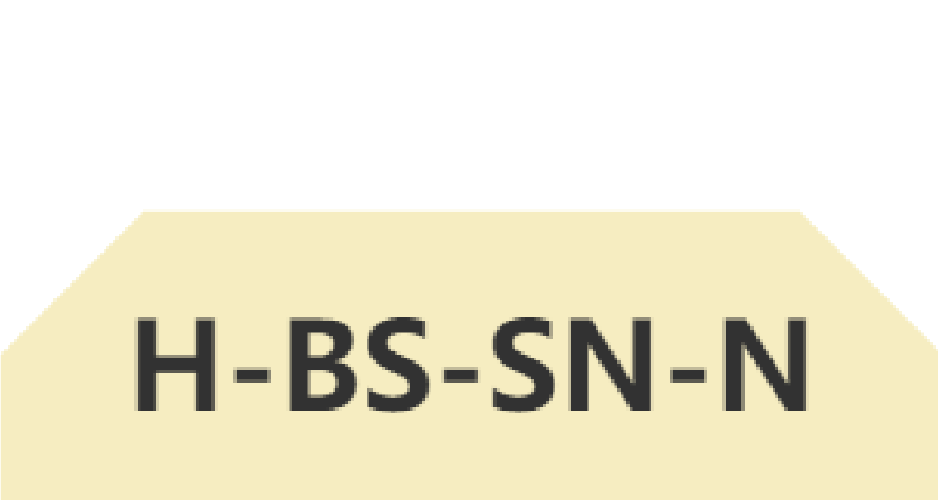}
       
    \includegraphics[width=0.5cm,height=1.5cm]{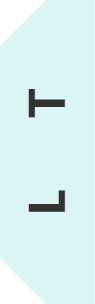}\includegraphics[width=2.23cm,height=1.5cm]{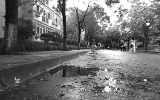}\includegraphics[width=2.23cm,height=1.5cm]{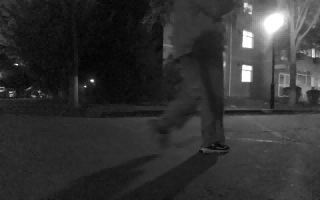}\includegraphics[width=2.23cm,height=1.5cm]{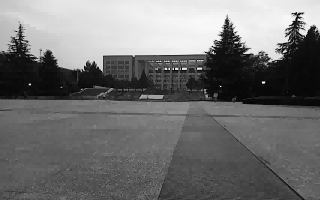}\includegraphics[width=2.23cm,height=1.5cm]{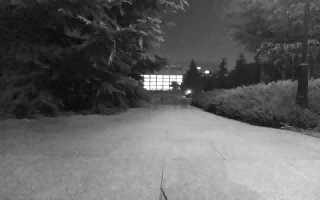}\includegraphics[width=2.23cm,height=1.5cm]{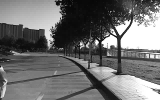}\includegraphics[width=2.23cm,height=1.5cm]{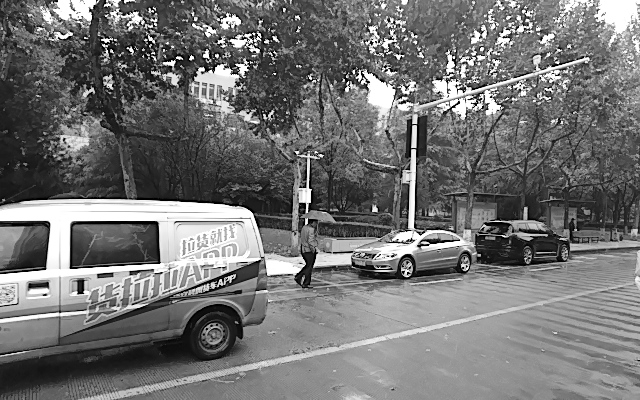}\includegraphics[width=2.23cm,height=1.5cm]{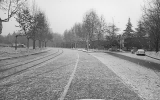}\includegraphics[width=2.23cm,height=1.5cm]{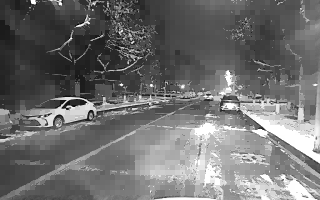}
    
    \includegraphics[width=0.5cm,height=1.5cm]{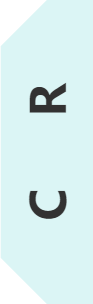}\includegraphics[width=2.23cm,height=1.5cm]{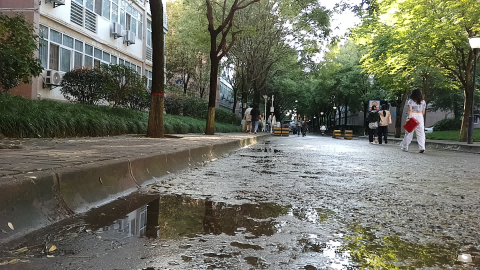}\includegraphics[width=2.23cm,height=1.5cm]{ddbqy.png}\includegraphics[width=2.23cm,height=1.5cm]{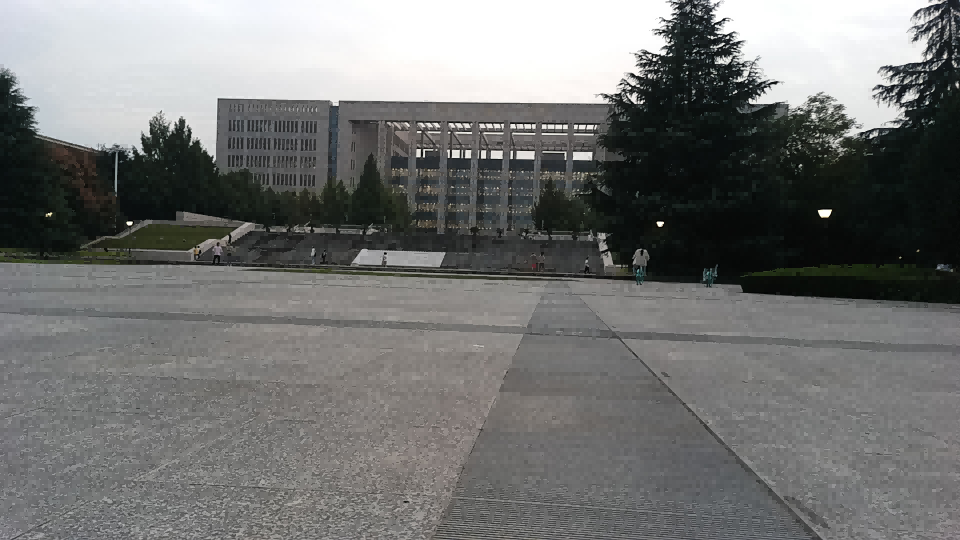}\includegraphics[width=2.23cm,height=1.5cm]{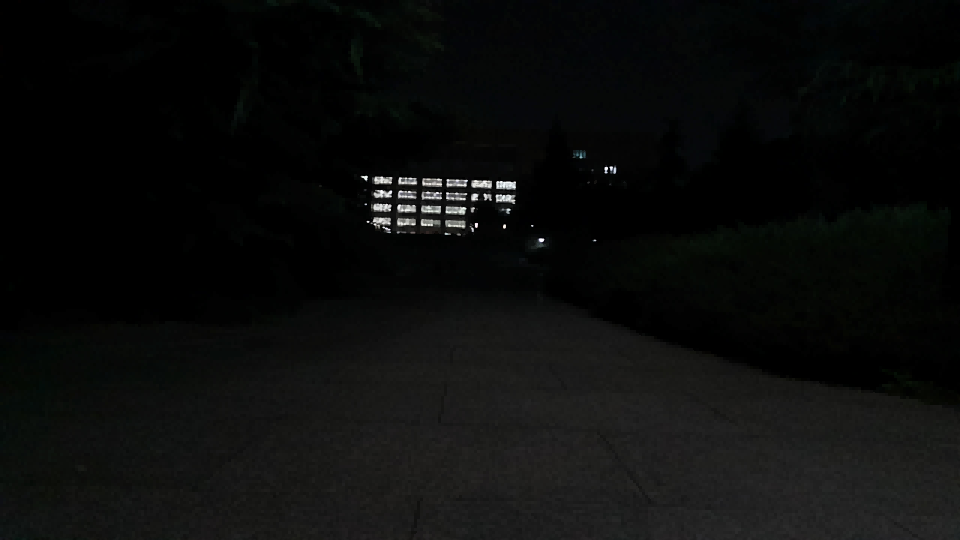}\includegraphics[width=2.23cm,height=1.5cm]{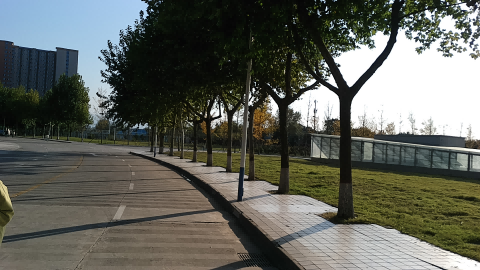}\includegraphics[width=2.23cm,height=1.5cm]{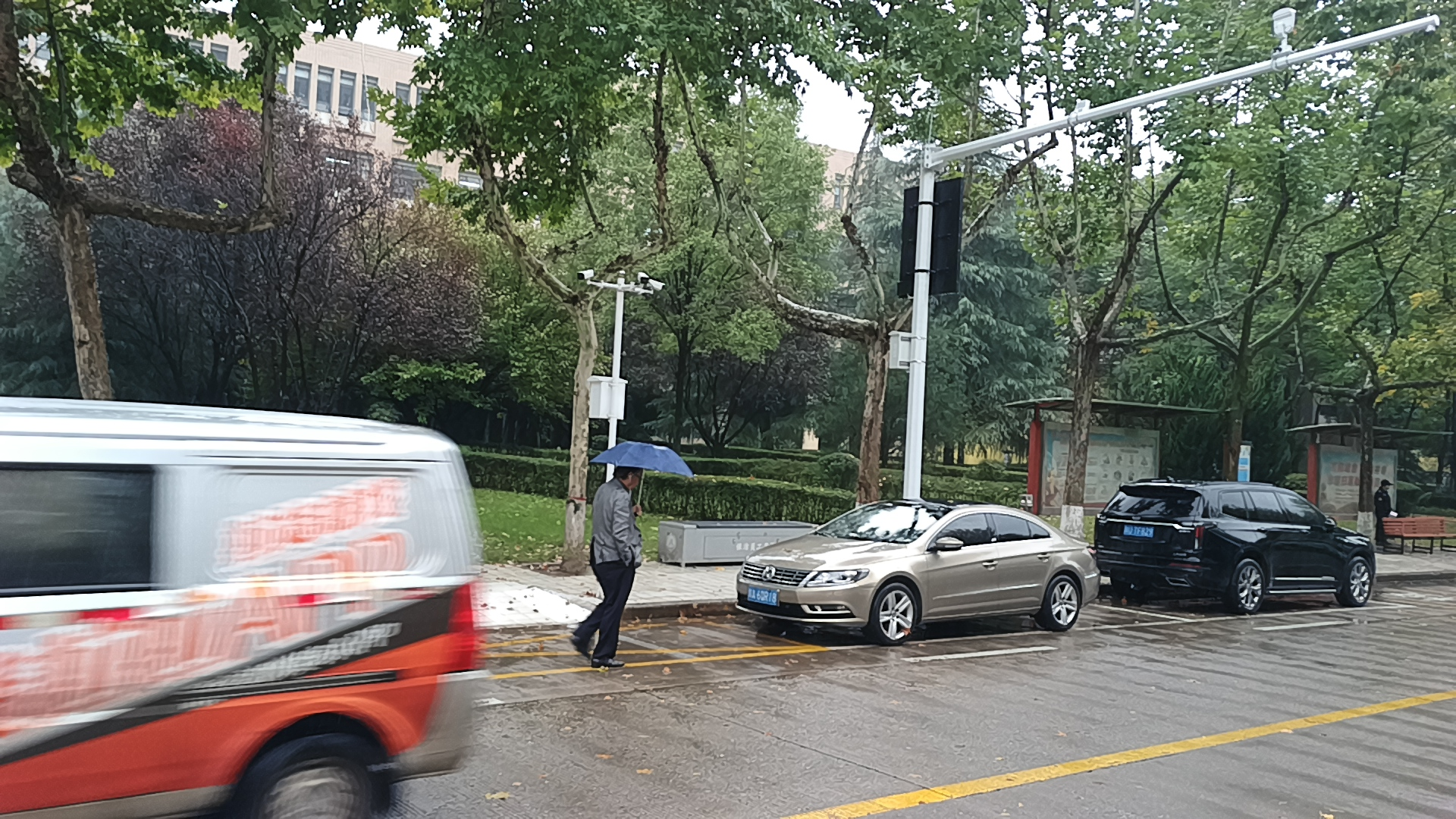}\includegraphics[width=2.23cm,height=1.5cm]{c4.png}\includegraphics[width=2.23cm,height=1.5cm]{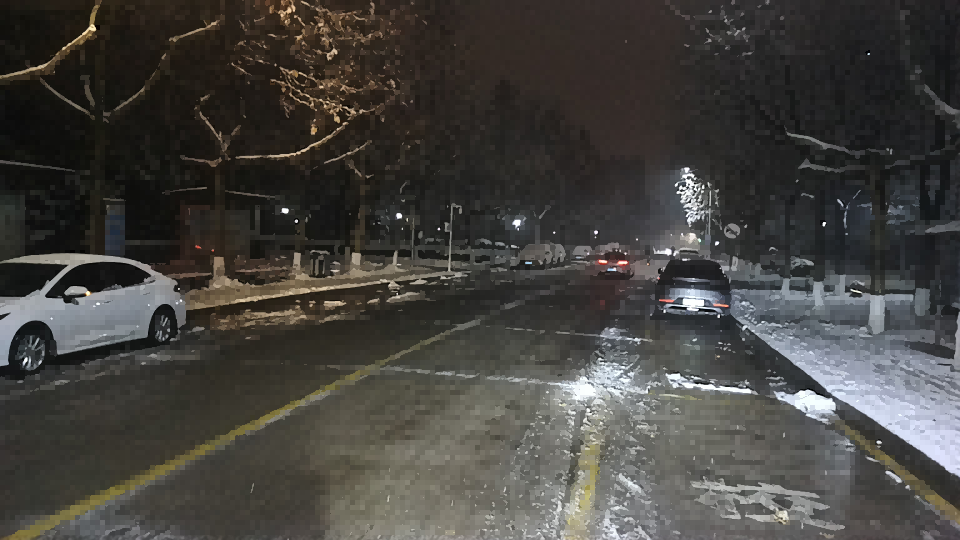}

    \includegraphics[width=0.5cm,height=1.5cm]{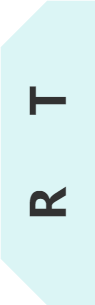}\includegraphics[width=2.23cm,height=1.5cm]{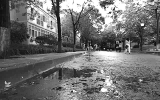}\includegraphics[width=2.23cm,height=1.5cm]{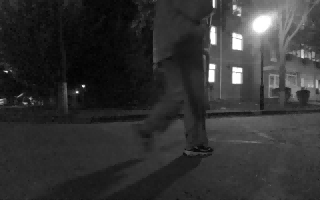}\includegraphics[width=2.23cm,height=1.5cm]{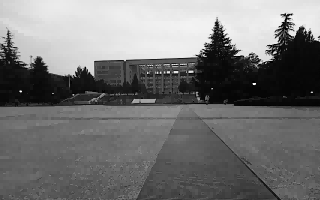}\includegraphics[width=2.23cm,height=1.5cm]{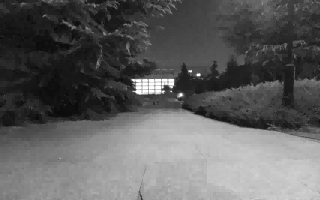}\includegraphics[width=2.23cm,height=1.5cm]{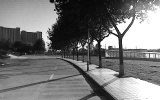}\includegraphics[width=2.23cm,height=1.5cm]{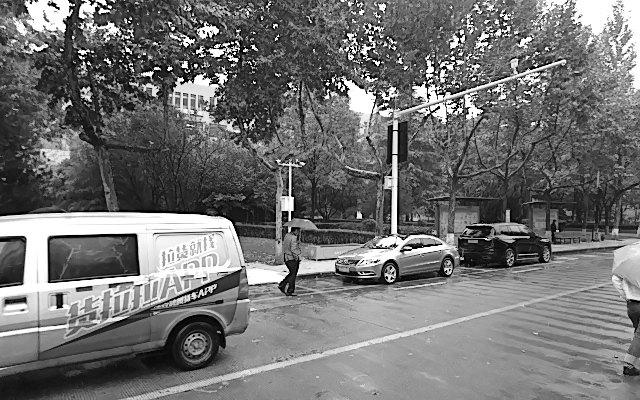}\includegraphics[width=2.23cm,height=1.5cm]{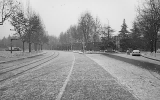}\includegraphics[width=2.23cm,height=1.5cm]{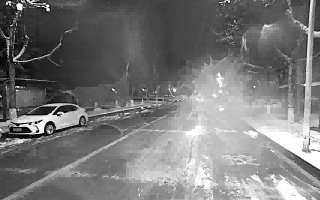}
    
   \includegraphics[width=0.5cm,height=1.5cm]{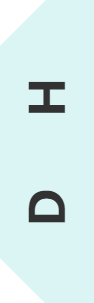}\includegraphics[width=2.23cm,height=1.5cm]{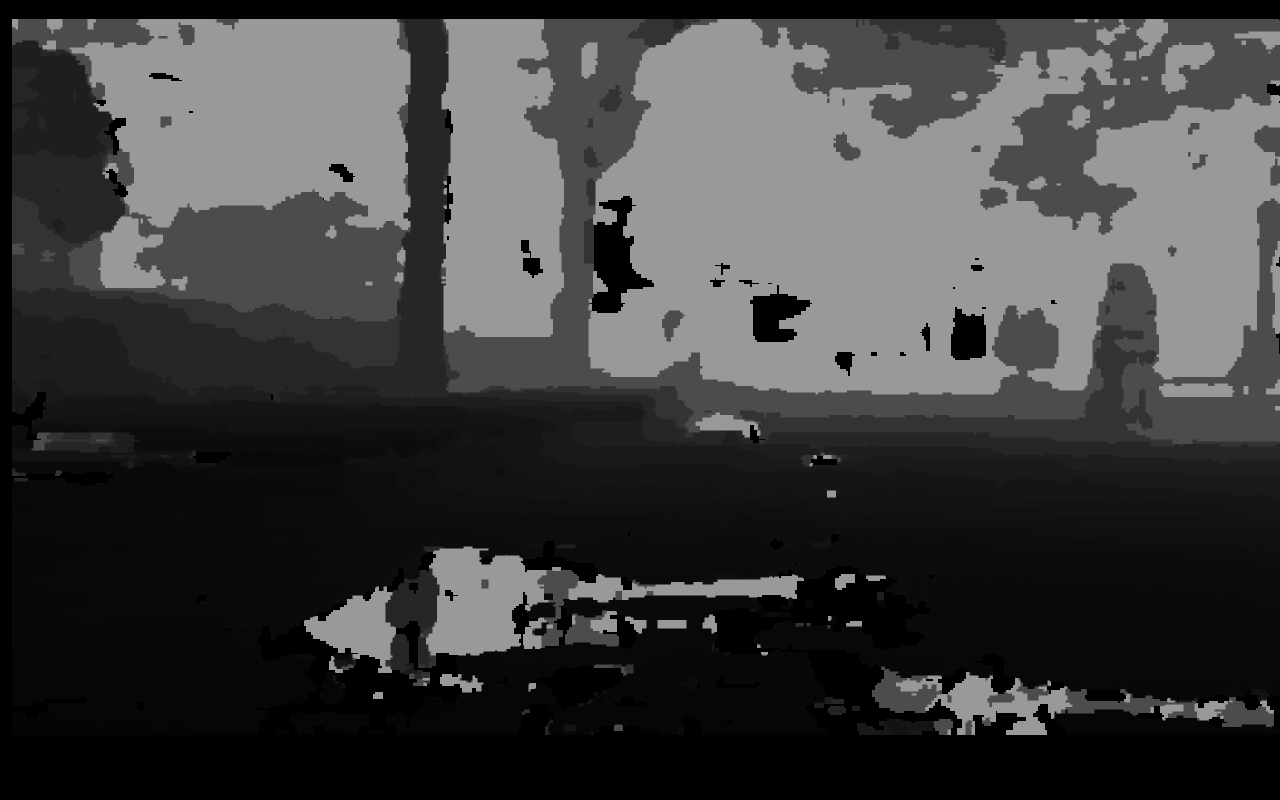}\includegraphics[width=2.23cm,height=1.5cm]{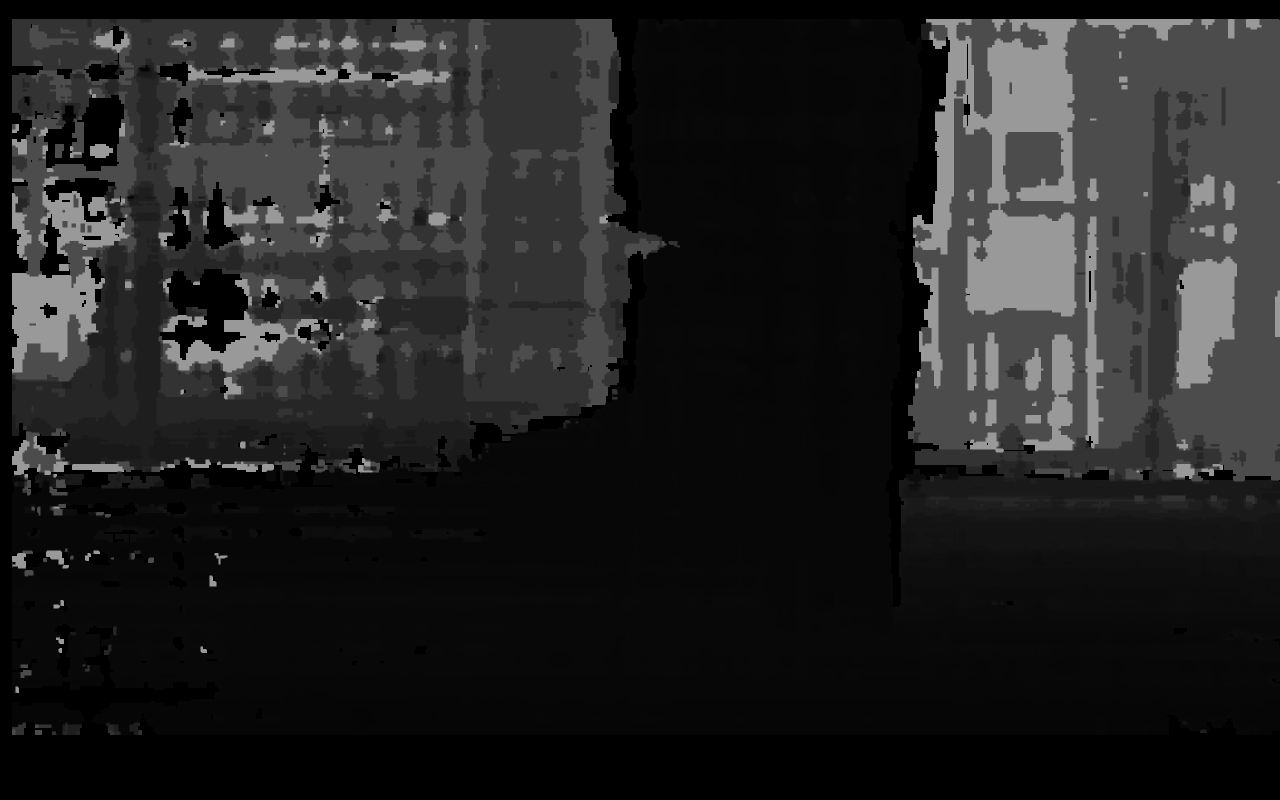}\includegraphics[width=2.23cm,height=1.5cm]{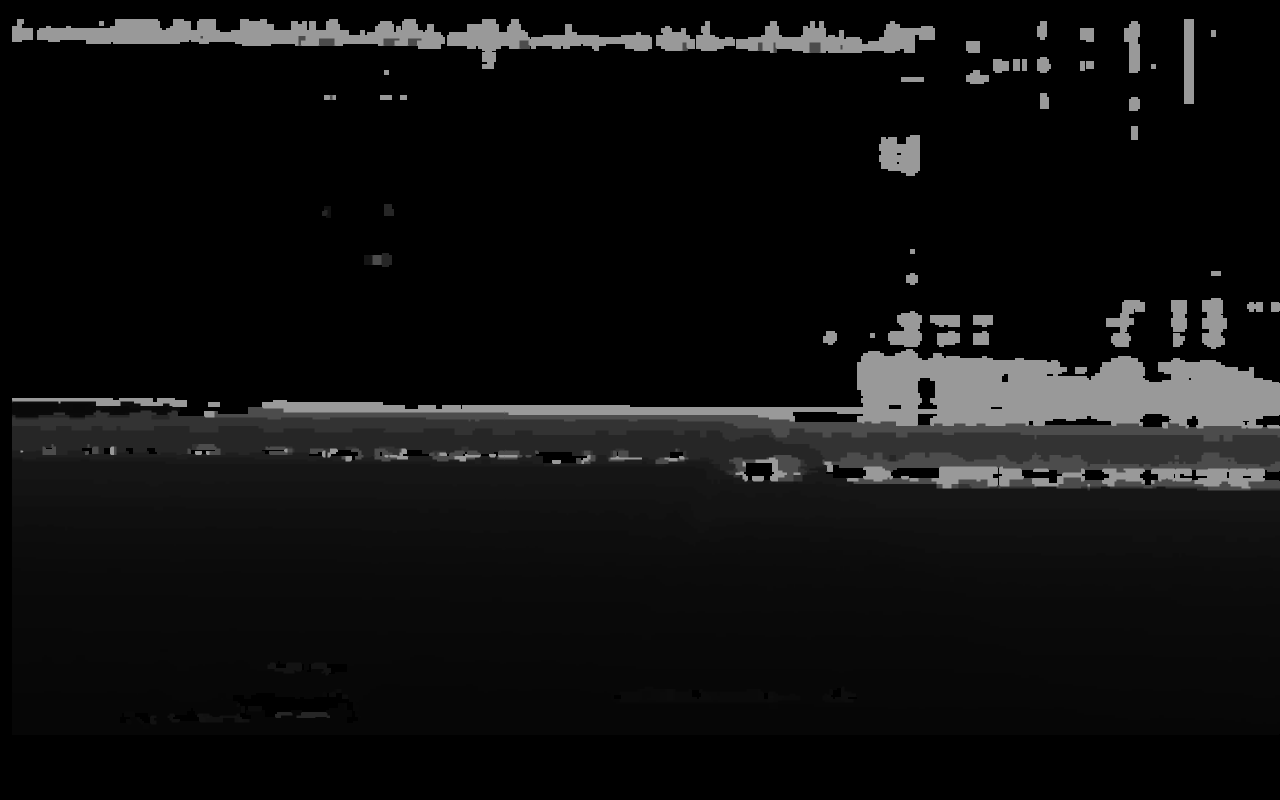}\includegraphics[width=2.23cm,height=1.5cm]{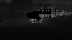}\includegraphics[width=2.23cm,height=1.5cm]{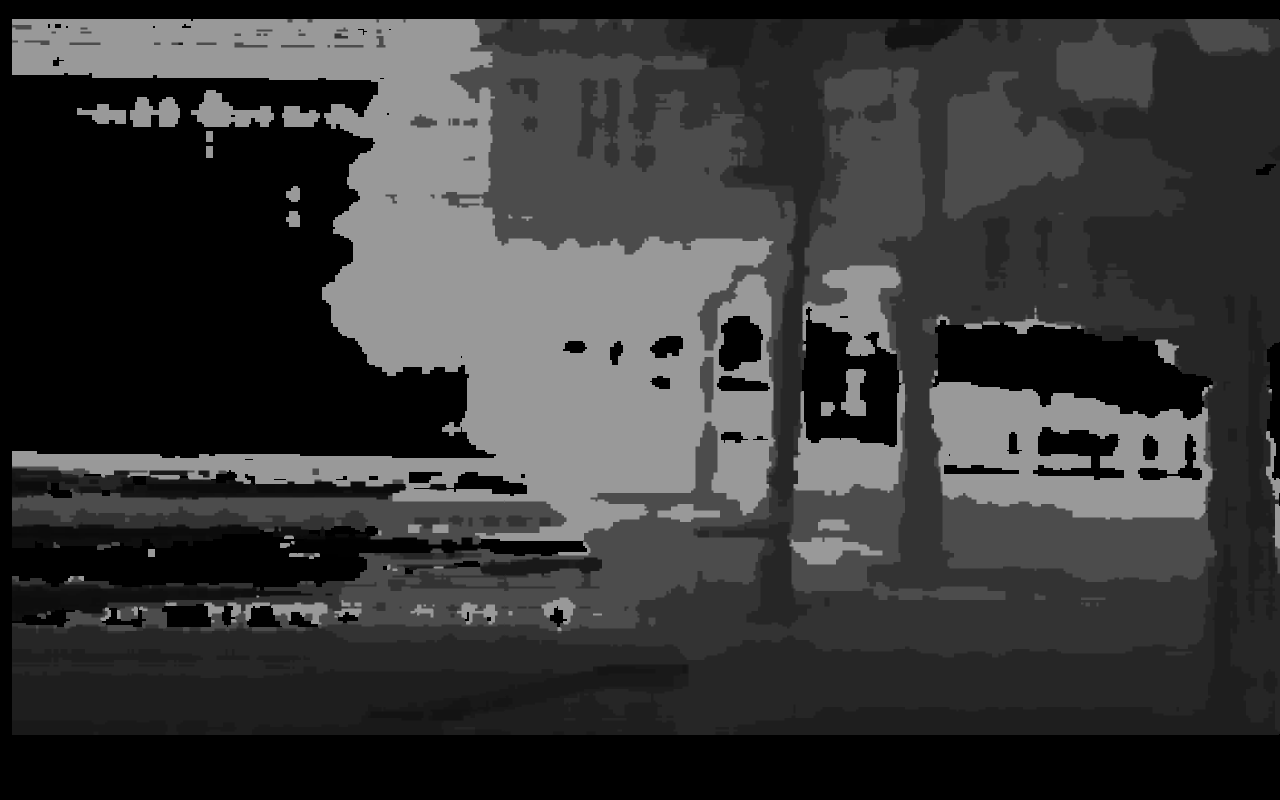}\includegraphics[width=2.23cm,height=1.5cm]{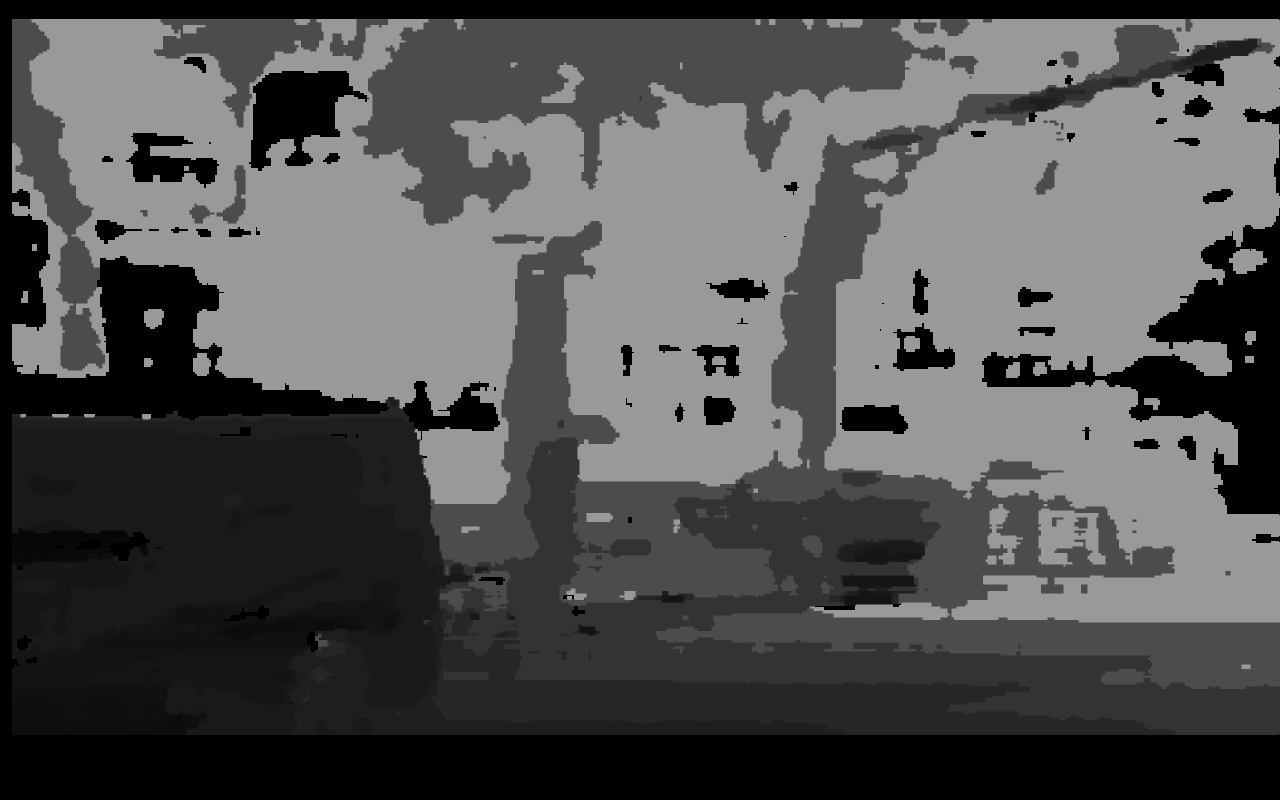}\includegraphics[width=2.23cm,height=1.5cm]{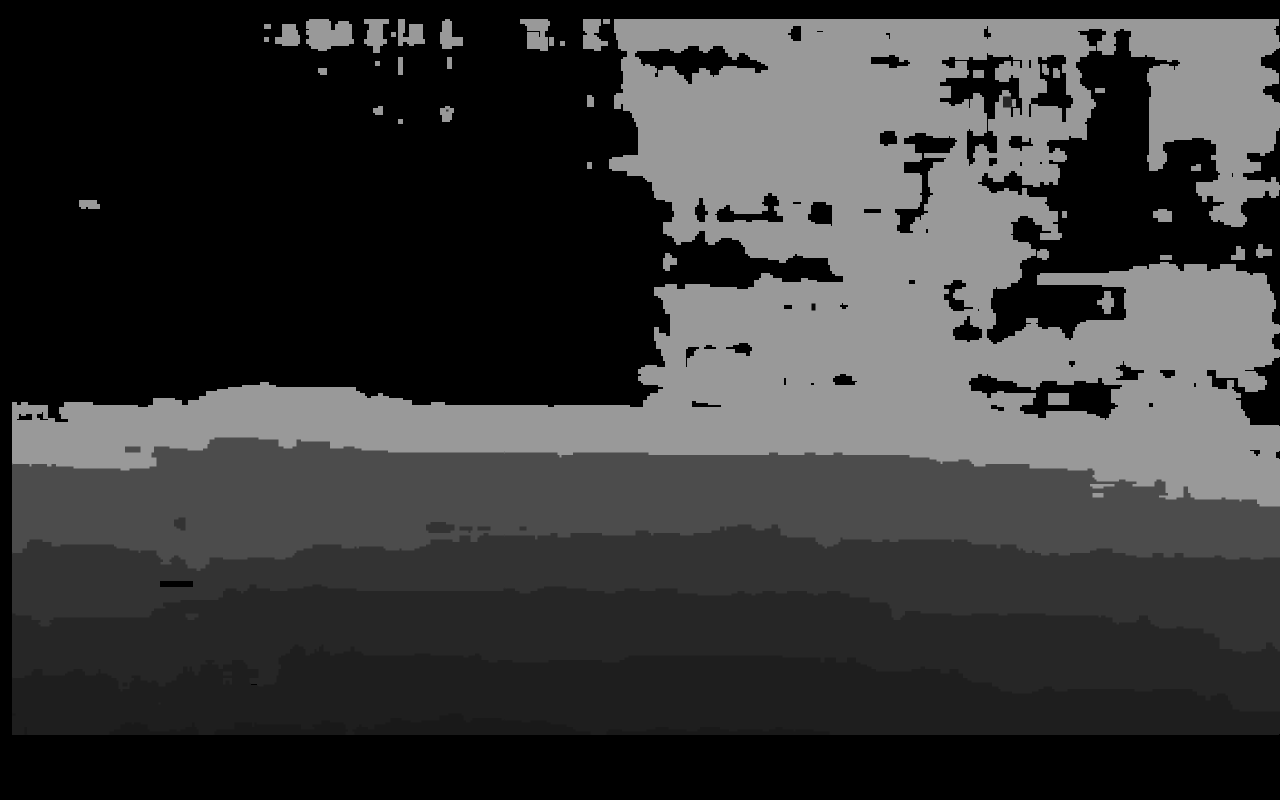}\includegraphics[width=2.23cm,height=1.5cm]{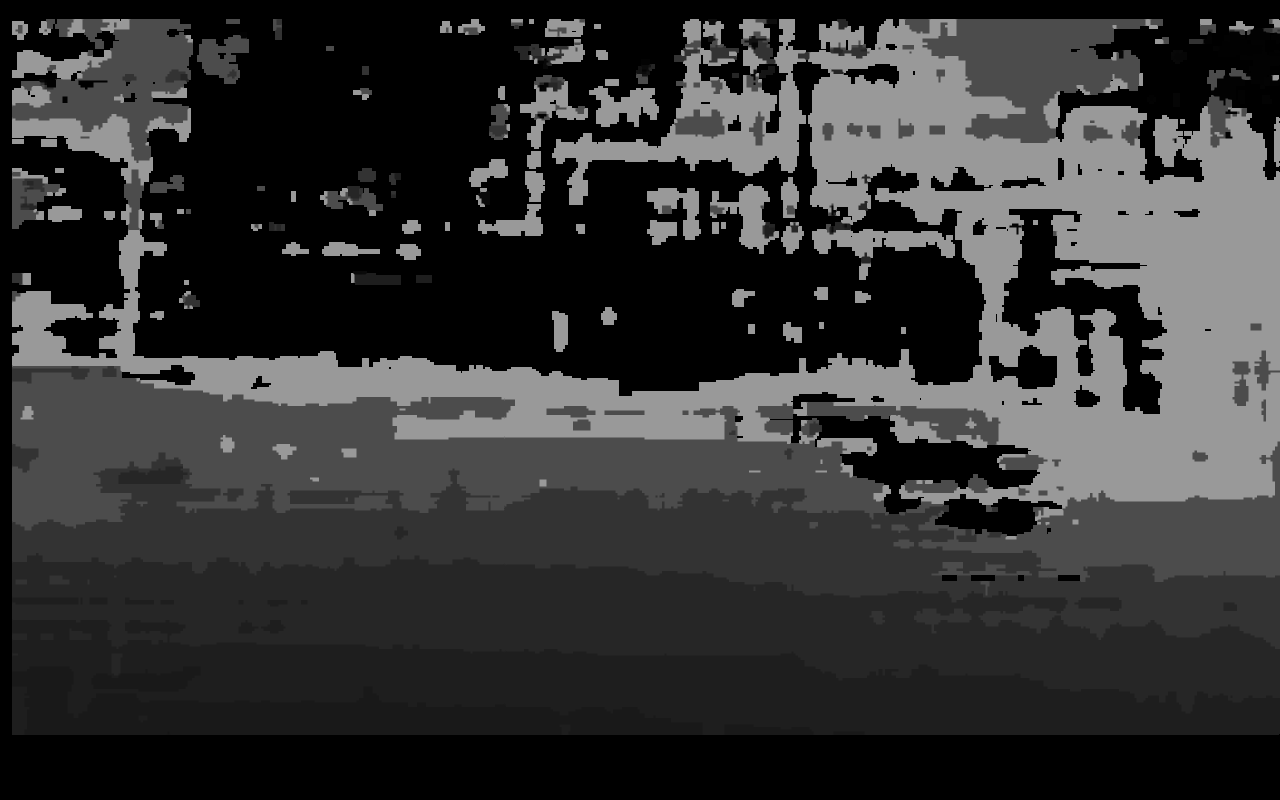}

    \includegraphics[width=0.5cm,height=1.5cm]{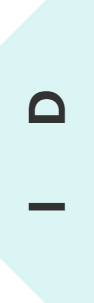}\includegraphics[width=2.23cm,height=1.5cm]{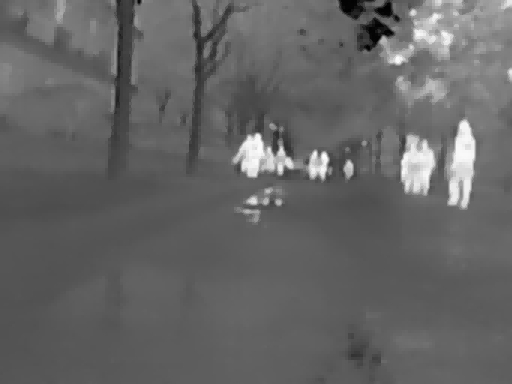}\includegraphics[width=2.23cm,height=1.5cm]{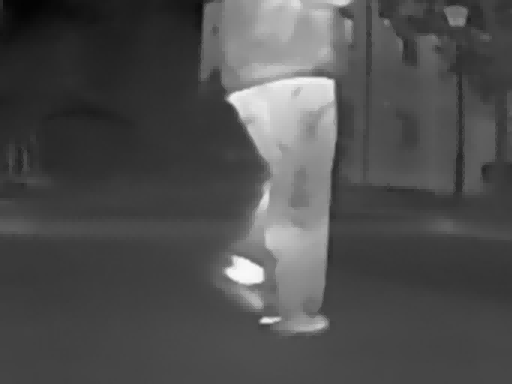}\includegraphics[width=2.23cm,height=1.5cm]{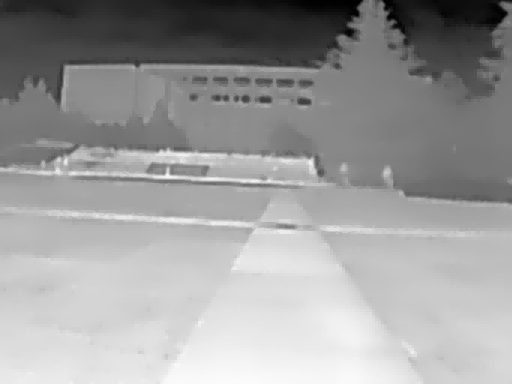}\includegraphics[width=2.23cm,height=1.5cm]{dsbdbqyi.png}\includegraphics[width=2.23cm,height=1.5cm]{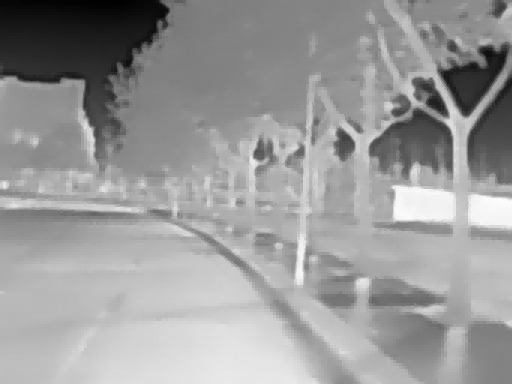}\includegraphics[width=2.23cm,height=1.5cm]{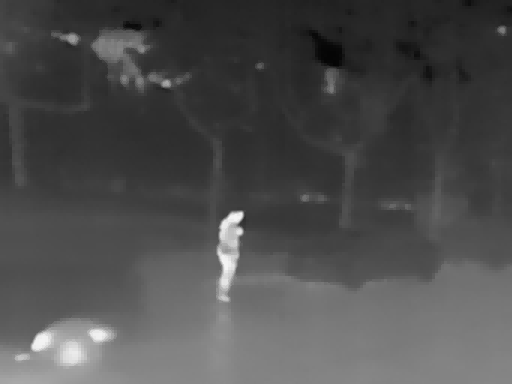}\includegraphics[width=2.23cm,height=1.5cm]{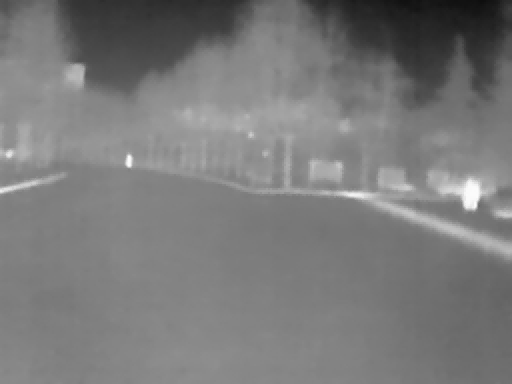}\includegraphics[width=2.23cm,height=1.5cm]{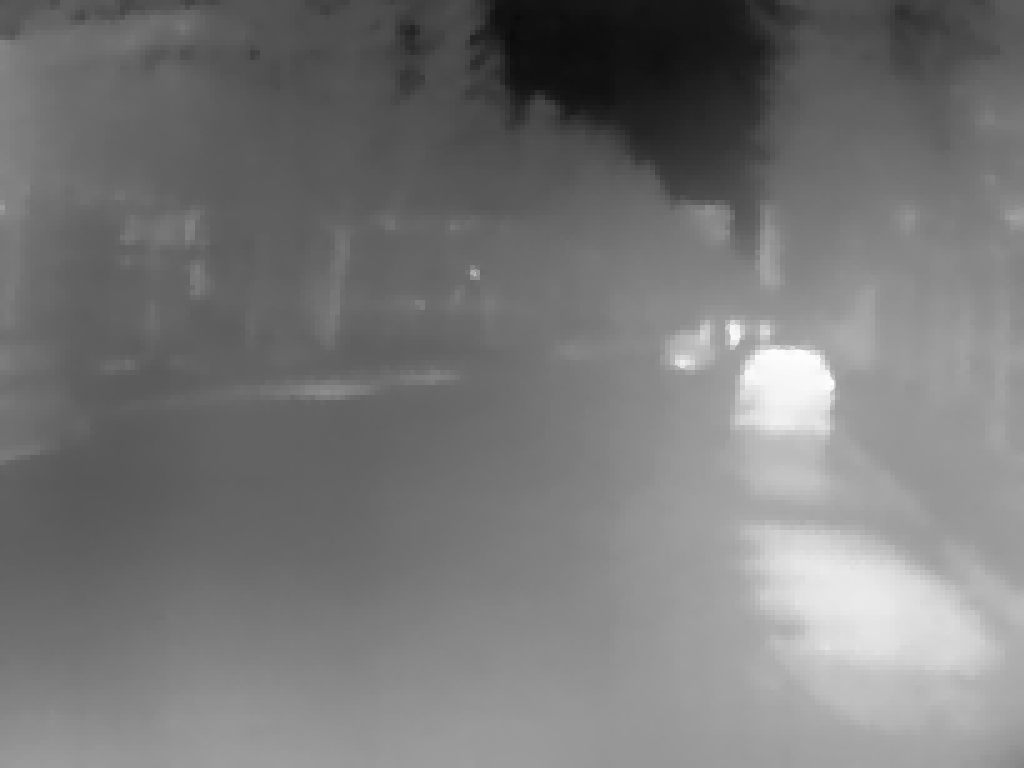}

    \includegraphics[width=0.5cm,height=1.5cm]{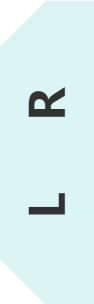}\includegraphics[width=2.23cm,height=1.5cm]{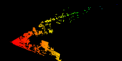}\includegraphics[width=2.23cm,height=1.5cm]{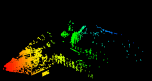}\includegraphics[width=2.23cm,height=1.5cm]{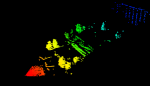}\includegraphics[width=2.23cm,height=1.5cm]{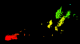}\includegraphics[width=2.23cm,height=1.5cm]{2li.png}\includegraphics[width=2.23cm,height=1.5cm]{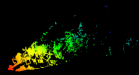}\includegraphics[width=2.23cm,height=1.5cm]{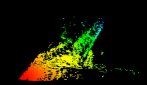}\includegraphics[width=2.23cm,height=1.5cm]{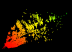}

    \includegraphics[width=0.5cm,height=1.5cm]{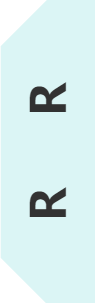}\includegraphics[width=2.23cm,height=1.5cm]{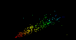}\includegraphics[width=2.23cm,height=1.5cm]{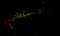}\includegraphics[width=2.23cm,height=1.5cm]{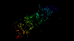}\includegraphics[width=2.23cm,height=1.5cm]{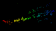}\includegraphics[width=2.23cm,height=1.5cm]{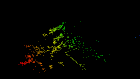}\includegraphics[width=2.23cm,height=1.5cm]{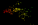}\includegraphics[width=2.23cm,height=1.5cm]{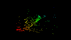}\includegraphics[width=2.23cm,height=1.5cm]{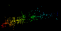}
    
  \caption{The figure provides detailed information about the collected dataset. LT: Left, CR: RGB, RT: Right, DH: Depth, ID: Infrared, LR: Lidar, RR: Radar. L: Low Speed, H: High Speed, B: Bumpy, NB: Non-Bumpy, BS: Bumps, SU: Sunny, R: Rainy, SN: Snowy, D: Day, N: Night.}
    \label{fig:data_distribution}\vspace{-5pt}
\end{figure*}
For dual-sensor fusion, the LiDAR–IMU combination showed higher latency than the radar-only baseline, as the LiDAR front end still processed dense scans. The vision–IMU SLAM (VINS-Mono) was faster than ORB-SLAM3 because of its efficient optical-flow tracking, despite inertial data integration. In the triple-sensor setup, the LiDAR–IMU modules in R3LIVE and FAST-LIVO2 had comparable runtimes. However, R3LIVE’s image-processing component increased its latency by over an order of magnitude compared to FAST-LIVO2, making it non-real-time. All computational experiments were conducted on a workstation equipped with an Intel Core i9-13900K processor (24 cores, 32 threads, boost clock up to 5.8 GHz), 62 GB system memory (RAM), and a discrete GPU with 24 GB dedicated video memory (VRAM). Except for R3LIVE, all algorithms listed in Table VII meet real-time requirements. 

\section{Conclusion}

We have collected a dataset from ground robots and autonomous vehicles under various weather conditions, different road surface conditions, and varying lighting conditions. The dataset combines novel and traditional sensors to help address accuracy challenges of SLAM in extreme conditions. By amassing substantial data and making it publicly available, we aim to provide SLAM researchers with more options for their studies. We have evaluated state-of-the-art SLAM algorithms based on various sensor types using this dataset. Experimental results demonstrate that these algorithms' positioning accuracy needs further improvement under certain extreme conditions. In future work, we plan to enhance SLAM's adaptability to various extreme environments through multi-sensor fusion methods.

\section*{License}
This dataset is licensed under CC BY 4.0.

\appendix

\begin{figure*}[t!]\vspace{0pt}
   \centering
  \includegraphics[width=8.3cm,height=4.5cm]{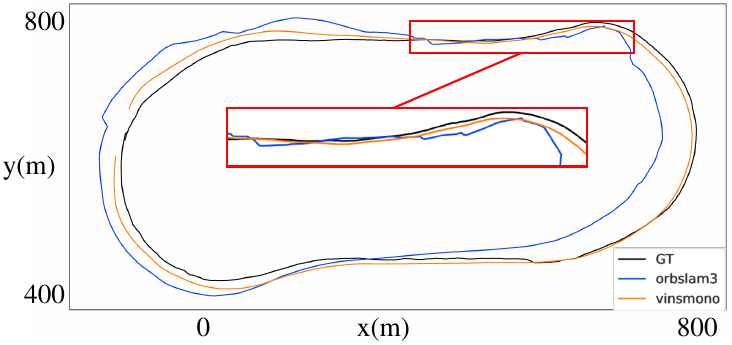}% 
  \includegraphics[width=8cm,height=4.5cm]{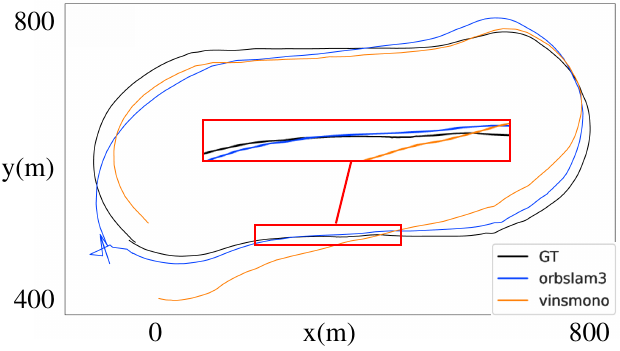}%
  \\
  \makebox[8cm]{(a) V-H-BS-SU-D}\makebox[8cm]{(b) V-H-BS-SN-D} \\
  \includegraphics[width=8cm,height=4.5cm]{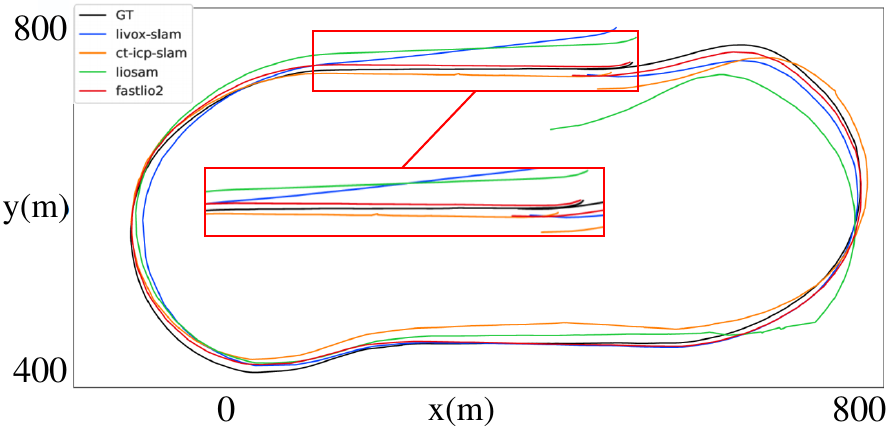}
  \includegraphics[width=8cm,height=4.5cm]{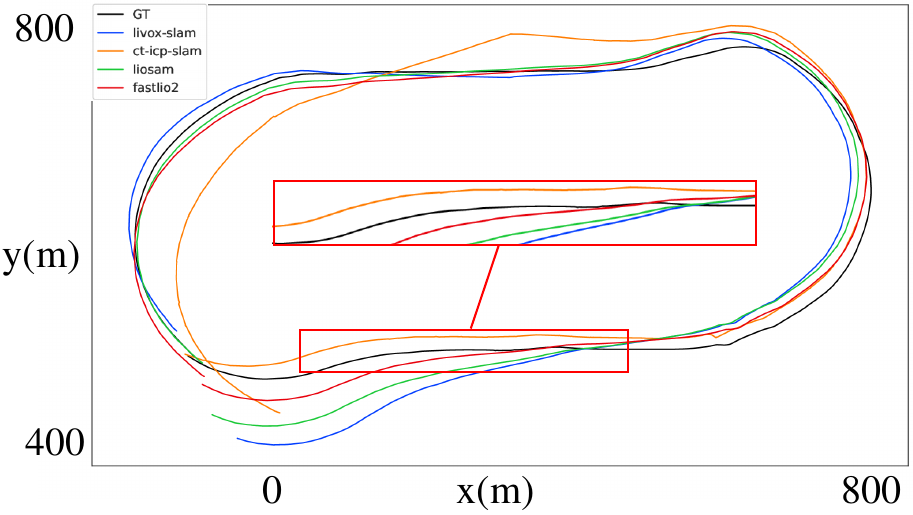}%
  \\
 \makebox[8cm]{(c) L-H-BS-R-D}\makebox[8cm]{(d) L-H-BS-SN-D}\\
  \includegraphics[width=8.1cm,height=4.5cm]{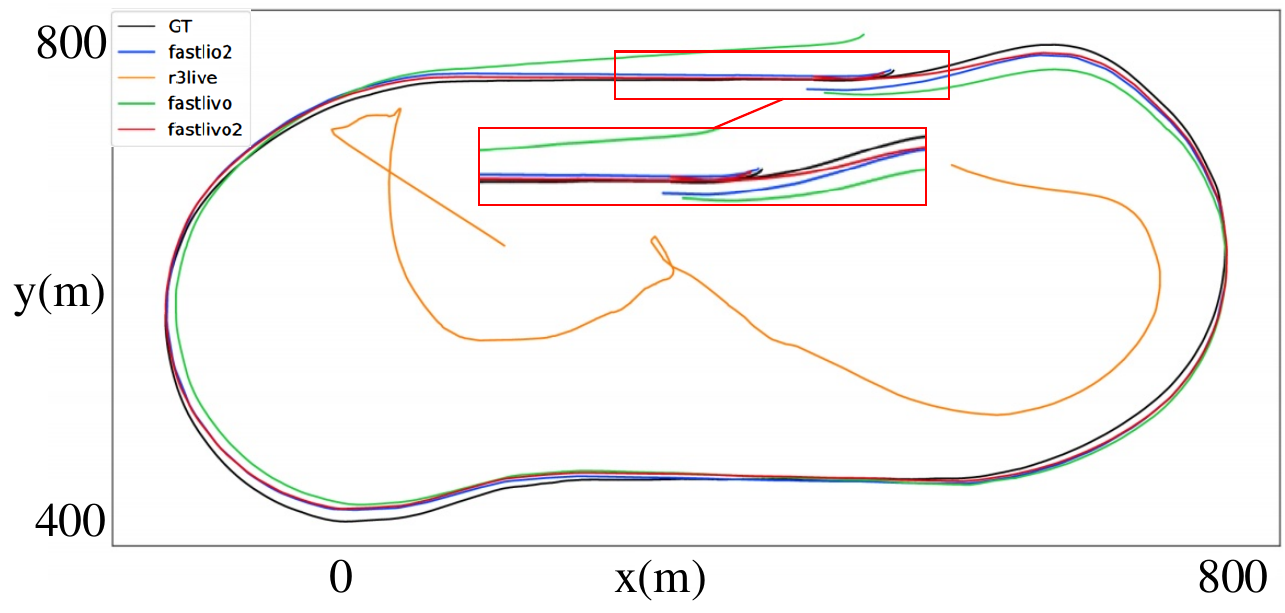}%
  \includegraphics[width=8cm,height=4.5cm]{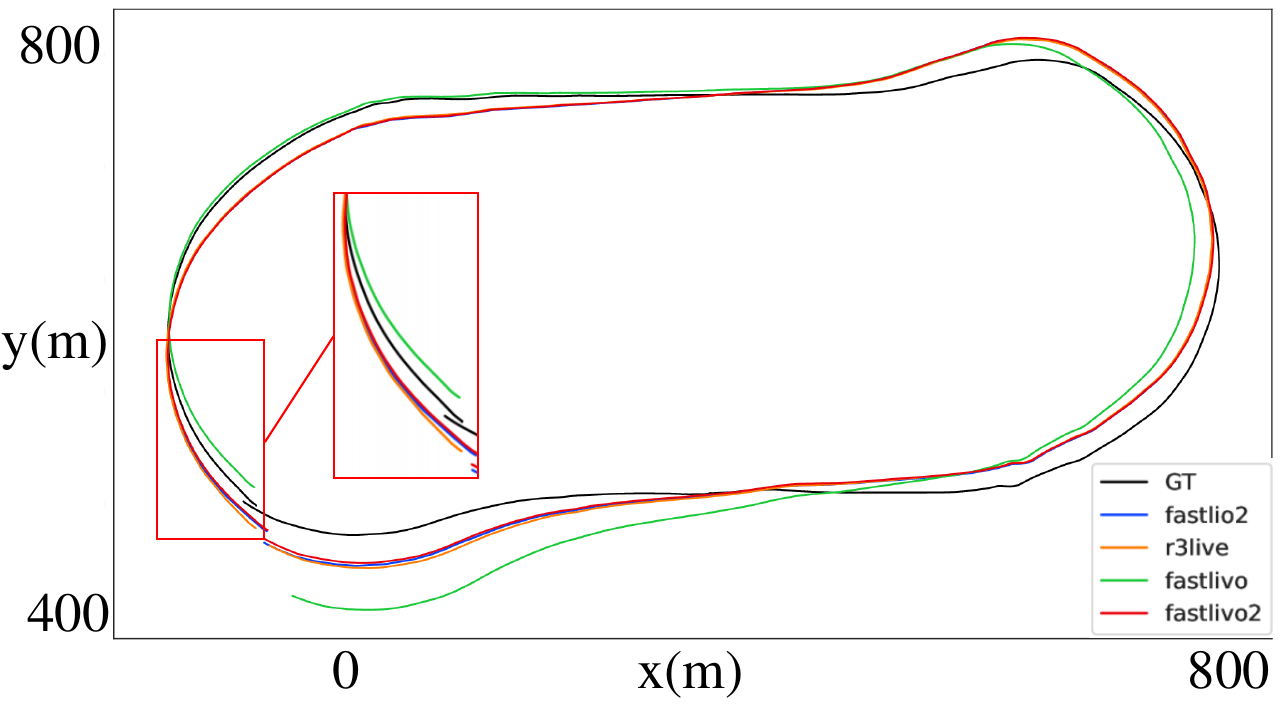}\\
  \makebox[8cm]{(f) F-H-BS-R-D}\makebox[8cm]{(g) F-H-BS-SN-D}
    \vspace{-5pt}
  \caption{The figure shows typical problem trajectory plots for Visual SLAM, LiDAR SLAM, and LiDAR-IMU-Visual Fusion SLAM.The labels “V-”, “L-”, and “F-” are used to denote Visual, LiDAR, and Fusion SLAM, respectively.}\vspace{-10pt}
\end{figure*}
\paragraph{Dataset Details}

We provide the information of all topics for some of our datasets, as shown in Fig. 12.

\paragraph{Trajectories and Analyses}

We place the overly large trajectory diagrams in the appendix. For clarity and intuitiveness, we only display meaningful trajectory diagrams in each figure.

By comparing the trajectory plots of ORB-SLAM3 and VINS-Mono under the same scenarios, it is evident that the visual SLAM systems augmented with IMU significantly enhance the accuracy of the trajectories. When comparing the trajectory plot of ORB-SLAM3 in Fig. 13 (a) with the speed bump distribution in Fig. 5, it is evident that the trajectory of ORB-SLAM3 exhibits significant deviations each time it passes through areas with speed bumps. We speculate that these deviations are caused by the jolts experienced when traversing speed bumps at high speed, which negatively impact the monocular SLAM system. In contrast, VINS-Mono, which integrates an IMU, displays a smoother trajectory. In snowy scenarios, due to the successful loop closure detection of ORB-SLAM3, the accumulated errors are remarkably reduced. In contrast, VINS-Mono fails in loop closure detection. Consequently, although the trajectory of VINS-Mono appears smoother, the Absolute Trajectory Error (ATE) actually increases. This highlights the important contribution of precise loop closure detection to SLAM. Furthermore, in both low-speed and high-speed sequences, vision-based SLAM algorithms struggle to achieve the highest level of precision.

In rainy and snowy high-speed scenarios with multiple speed bumps, LIO-SAM's robustness lags behind FastLIO2 due to algorithmic disparities. LIO-SAM uses full point cloud data and manual feature extraction, making it vulnerable to noise and discontinuities from bumps, which disrupt feature extraction and matching, leading to pose estimation errors. Its sliding window optimization cannot effectively handle accumulated errors, and its map update struggles to keep up with rapid environmental changes. Conversely, FastLIO2 forgoes traditional feature extraction, reduces noise sensitivity via point cloud downsampling and enhancement. Its incremental k-d tree enables efficient map updates for quick adaptation. Tight lidar-IMU fusion, along with high computational efficiency, ensures real-time performance and better stability in such bumpy conditions.

We also test the snowy scenario on the 4D millimeter-wave slam. The SLAM cannot meet the demand for high-quality SLAM in sparse point clouds. So better SLAM development for this sensor is still needed.

We evaluate the performance of three sensor-fusion SLAM algorithms: FastLIO, FastLIO2, and R3LIVE, which integrate LiDAR, IMU, and visual data. Our analysis cover both trajectory errors and mapping quality, including point cloud density, and color accuracy.
In rainy conditions, R3LIVE shows a significant drift in the trajectory when visual features are compromised by rain, while FastLIVO's trajectory is worse than FastLIO2's. Moreover, FastLIO2 achieve better trajectories and a mark increase in ATE.
All three algorithms effectively fuse visual data, but their varying fusion mechanisms affect system robustness differently. In snowy conditions, R3LIVE and FastLIVO2 (with visual data) show no clear trajectory-accuracy improvement over FastLIO2 (without visual data), likely because heavy snow degrades visual-pose estimation, limiting error compensation for LiDAR. This indicates that current fusion-based SLAM algorithms are insufficient for snowy conditions. Future work should focus on algorithm improvements or integrating 4D millimeter-wave radar to address this issue.
Additionally, we observe that in areas with speed bumps, the fusion algorithms effectively compensate for errors, generally preventing trajectory drift caused by jolts.

\vspace{-33pt}
\begin{IEEEbiography}[{\includegraphics[width=1in,height=1.25in,clip,keepaspectratio]{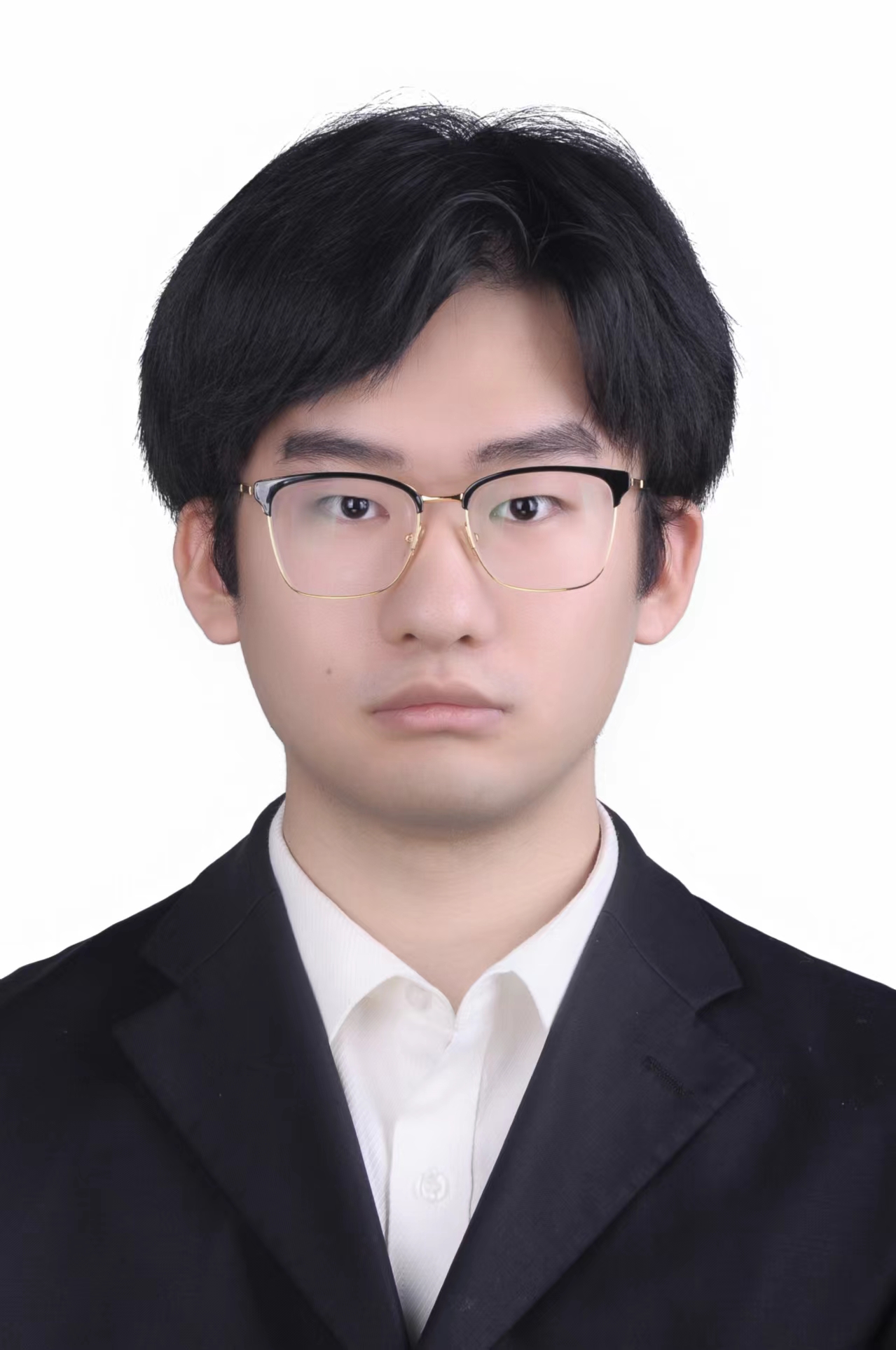}}]{Weisheng Gong} is a Ph.D. student at the School of Information Science, Northwest University. He graduated from Xi'an Jiaotong University with a Bachelor's degree in 2020. His research interests include autonomous driving and ground robot SLAM, multi-sensor fusion experimental platform construction, dataset collection, and multi-sensor fusion SLAM development.
\end{IEEEbiography}
\vspace{-33pt}
\begin{IEEEbiography}[{\includegraphics[width=1in,height=1.25in,clip,keepaspectratio]{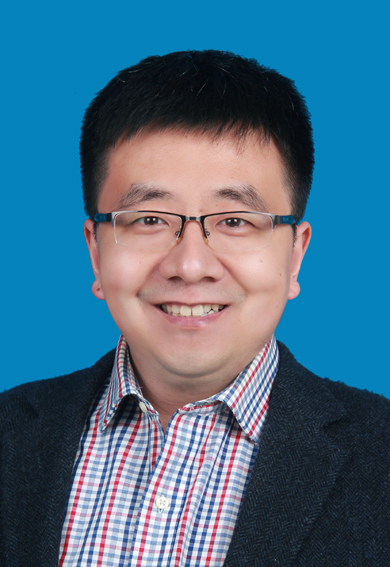}}]{Chen He} (Member, IEEE) received the B.Eng.
degree (summa cum laude) from McMaster University in 2007 and the M.A.Sc. and Ph.D. degrees
from The University of British Columbia (UBC),
Vancouver, in 2009 and 2014, respectively, all
in electrical and computer engineering. He was
a Research Engineer at Blackberry Ltd., Canada,
and a Post-Doctoral Research Fellow at UBC.
He is currently a Full Professor with Northwest
University, China. His research interests include signal processing and machine learning, with applications in wireless sensing and communications, and embodied intelligence. He is an Associate Editor of the IEEE SIGNAL
PROCESSING LETTERS.
\end{IEEEbiography}
\vspace{-33pt}
\begin{IEEEbiography}[{\includegraphics[width=1in,height=1.25in,clip,keepaspectratio]{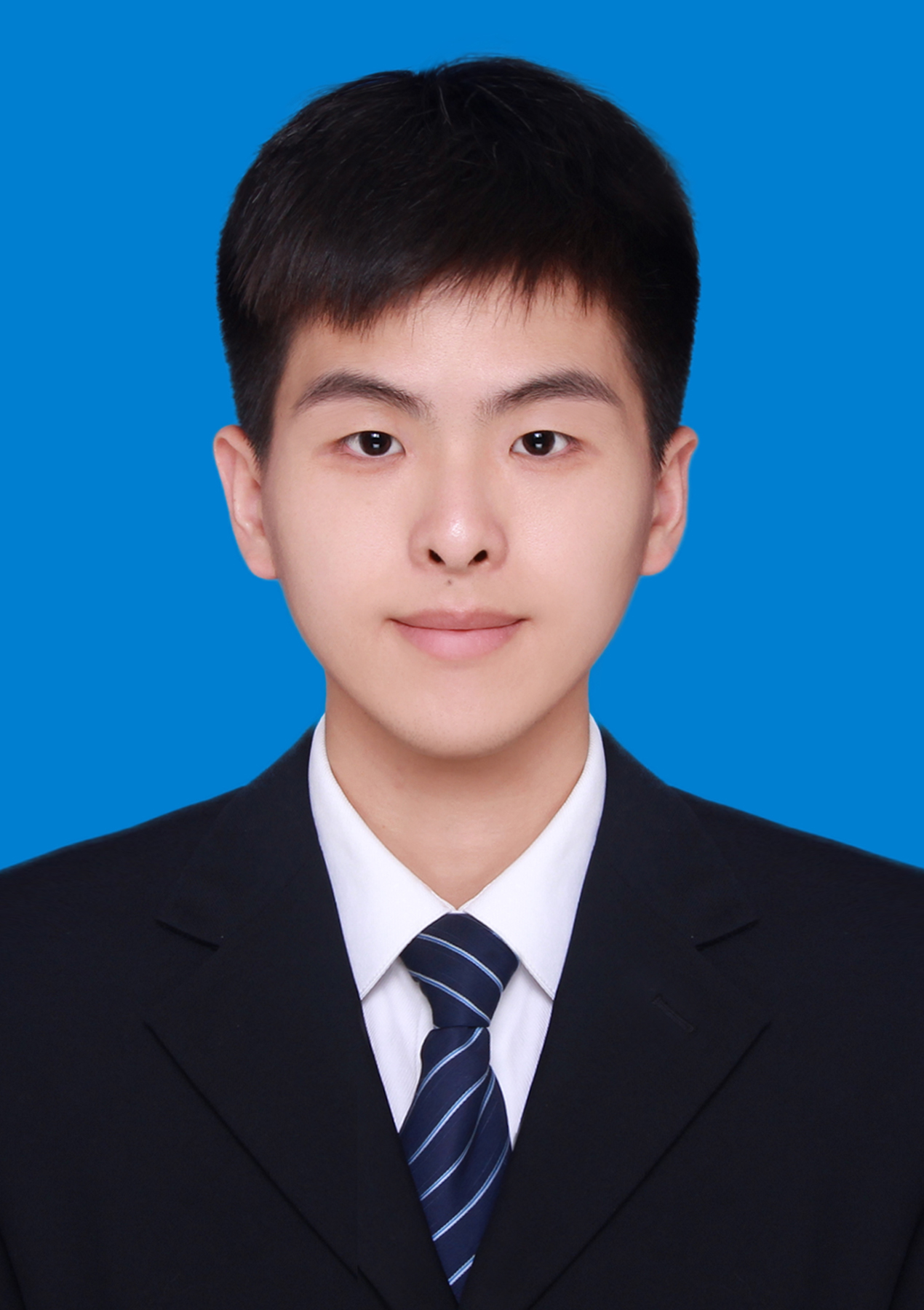}}]{Kaijie Su} is a Master's student at Northwest University. He obtained his bachelor's degree from the university in 2022, with research interests in autonomous driving and 4D millimeter-wave radar SLAM.
\end{IEEEbiography}
\vspace{-33pt}
\begin{IEEEbiography}[{\includegraphics[width=1in,height=1.25in,clip,keepaspectratio]{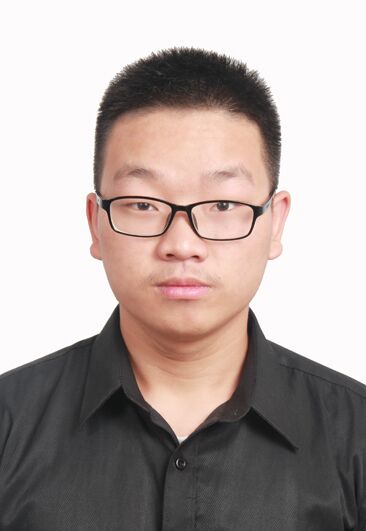}}]{Qingyong Li} is a Master's student at Northwest University. He earned his bachelor's degree there in 2022, focusing on autonomous driving and multi-sensor fusion for object detection in his research.
\end{IEEEbiography}
\vspace{-33pt}
\begin{IEEEbiography}[{\includegraphics[width=1in,height=1.25in,clip,keepaspectratio]{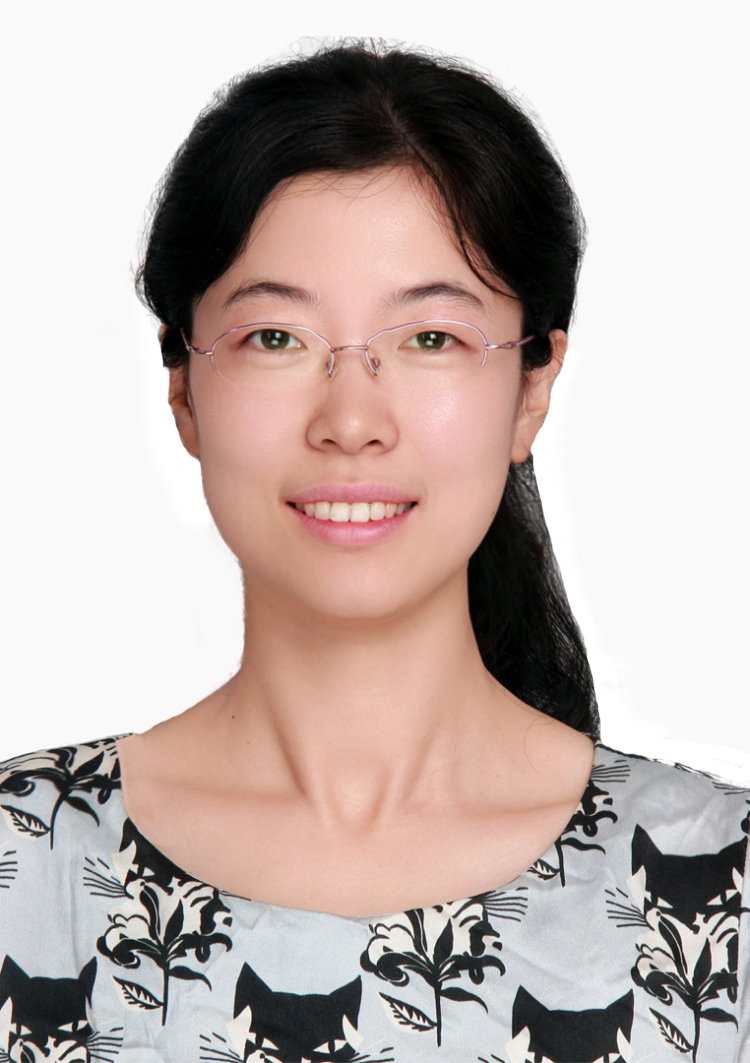}}]{Tong Wu} is an Engineer and a Ph.D student at Northwest University. She completed her undergraduate and master's degrees at Nanjing University, majoring in electronic engineering and computer science, with research directions in autonomous driving and object detection.
\end{IEEEbiography}
\vspace{-33pt}
\begin{IEEEbiography}[{\includegraphics[width=1in,height=1.25in,clip,keepaspectratio]{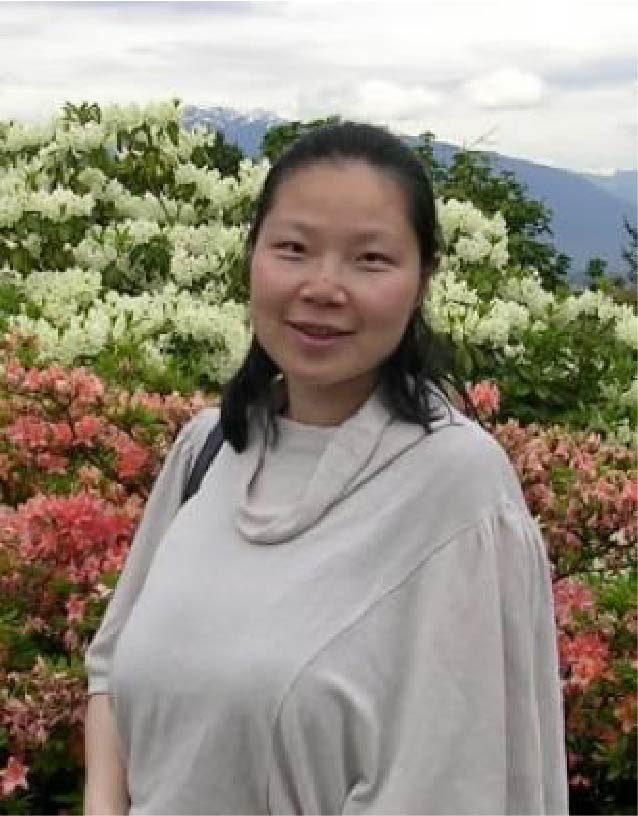}}]{Z. Jane Wang} (Fellow, IEEE) received the B.Sc.
degree in electrical engineering from Tsinghua
University, China, in 1996, and the M.Sc. and Ph.D.
degrees in electrical engineering from the University
of Connecticut, in 2000 and 2002, respectively. She
was a Research Associate at the Electrical and
Computer Engineering Department, University of
Maryland, College Park. Since 2004, she has been
with the Department of Electrical and Computer
Engineering, The University of British Columbia,
Canada, where she is currently a Professor. Her
research interests include statistical signal processing theory and applications,
with focus on multimedia security and biomedical signal processing and
modeling. She received the Outstanding Engineering Doctoral Student Award
from the University of Connecticut. She co-received the \textit{EURASIP Journal on
Applied Signal Processing} (JASP) Best Paper Award in 2004 and the IEEE
Signal Processing Society Best Paper Award in 2005. She is the Chair and
the Founder of the IEEE Signal Processing Chapter at Vancouver. She is an
Associate Editor of the IEEE TRANSACTIONS ON SIGNAL PROCESSING, the
IEEE TRANSACTIONS ON INFORMATION FORENSICS AND SECURITY, and
the IEEE TRANSACTIONS ON BIOMEDICAL ENGINEERING.
\end{IEEEbiography}


\begin{thebibliography}{99}
    \bibitem[1]{Pritsker1984} Pritsker, A. Alan B. (1984) Introduction to Simulation and SLAM II. Halsted Press.
    
    \bibitem[2]{Singandhupe2019} Singandhupe, A., La, H. M. (2019) A review of SLAM techniques and security in autonomous driving. In 2019 third IEEE international conference on robotic computing (IRC) (pp. 1-6). IEEE.
    
    \bibitem[3]{Cadena2016} Cadena, C., Carlone, L., Carrillo, H., Latif, Y., Scaramuzza, D., Neira, J., ...  Leonard, J. J. (2016) Past, present, and future of simultaneous localization and mapping: Toward the robust-perception age. IEEE Transactions on Robotics, 32(6), 1309-1332.
    
    \bibitem[4]{Geiger2012} Geiger, A., Lenz, P., Urtasun, R. (2012) Are we ready for autonomous driving? the kitti vision benchmark suite. In 2012 IEEE conference on computer vision and pattern recognition (pp. 1-8). IEEE.

    
    \bibitem[5]{oxford}Maddern W, Pascoe G, Linegar C, et al. 1 year, 1000 km: The oxford robotcar dataset[J]. The International Journal of Robotics Research, 2017, 36(1): 3-15.
    
    \bibitem[6]{Schubert2018} Schubert, D., Goll, T., Demmel, N., Usenko, V., Stückler, J., Cremers, D. (2018) The TUM VI benchmark for evaluating visual-inertial odometry. In 2018 IEEE/RSJ International Conference on Intelligent Robots and Systems (IROS) (pp. 1-6). IEEE.
    
    \bibitem[7]{Wenzel2021b} Wenzel, P., Whelan, T., Khan, Q., von Stumberg, L., Cremers, D. (2021) 4Seasons: A cross-season dataset for multi-weather SLAM in autonomous driving. Pattern Recognition: 42nd DAGM German Conference, DAGM GCPR 2020, Tübingen, Germany, September 28–October 1, 2020, Proceedings 42. Springer International Publishing.
    
    \bibitem[8]{Yin2021} Yin, J., Xie, Q., Wu, Z., Wu, Y., Xie, Q., Huang, G., ... Gao, F. (2021) M2DGR: A multi-sensor and multi-scenario SLAM dataset for ground robots. IEEE Robotics and Automation Letters, 7(2), 2266-2273.
    
    
    \bibitem[9]{Zhao2023} Zhao, T., Lu, X., Ye, T. (2023) A comprehensive implementation of road surface classification for vehicle driving assistance: Dataset, models, and deployment. IEEE Transactions on Intelligent Transportation Systems, 24(8), 8361-8370.
    
    \bibitem[10]{Zhao2024} Zhao, T., Xie, Y., Ding, M. et al. (2024) A road surface reconstruction dataset for autonomous driving. Sci Data, 11, 459.

    \bibitem[11]{urban} Wen W, Zhou Y, Zhang G, et al. UrbanLoco: A full sensor suite dataset for map** and localization in urban scenes[C]//2020 IEEE international conference on robotics and automation (ICRA). IEEE, 2020: 2310-2316. 

    \bibitem[12]{brno} Ligocki A, Jelinek A, Zalud L. Brno urban dataset-the new data for self-driving agents and map** tasks[C]//2020 IEEE International Conference on Robotics and Automation (ICRA). IEEE, 2020: 3284-3290. 

    \bibitem[13]{goose}  Mortimer P, Hagmanns R, Granero M, et al. The goose dataset for perception in unstructured environments[C]//2024 IEEE International Conference on Robotics and Automation (ICRA). IEEE, 2024: 14838-14844. 

    \bibitem[14]{kaist}Choi Y, Kim N, Hwang S, et al. KAIST multi-spectral day/night data set for autonomous and assisted driving[J]. IEEE Transactions on Intelligent Transportation Systems, 2018, 19(3): 934-948.
    
    \bibitem[15]{nuScenes}Caesar H, Bankiti V, Lang A H, et al. nuscenes: A multimodal dataset for autonomous driving[C]//Proceedings of the IEEE/CVF conference on computer vision and pattern recognition. 2020: 11621-11631.
    
    \bibitem[16]{Sheeny}Sheeny M, De Pellegrin E, Mukherjee S, et al. Radiate: A radar dataset for automotive perception in bad weather[C]//2021 IEEE International Conference on Robotics and Automation (ICRA). IEEE, 2021: 1-7.
    
    \bibitem[17]{Burnett}Burnett K, Yoon D J, Wu Y, et al. Boreas: A multi-season autonomous driving dataset[J]. The International Journal of Robotics Research, 2023, 42(1-2): 33-42.
    
    \bibitem[18]{OORD}Gadd M, De Martini D, Bartlett O, et al. Oord: The oxford offroad radar dataset[J]. IEEE Transactions on Intelligent Transportation Systems, 2024.
   
    \bibitem[19]{view}Palffy A, Pool E, Baratam S, et al. Multi-class road user detection with 3+ 1d radar in the view-of-delft dataset[J]. IEEE Robotics and Automation Letters, 2022, 7(2): 4961-4968.
    
    \bibitem[20]{dual}Zhang X, Wang L, Chen J, et al. Dual radar: A multi-modal dataset with dual 4d radar for autononous driving[J]. Scientific Data, 2025, 12(1): 439.
    
    \bibitem[21]{kradar}Paek D H, Kong S H, Wijaya K T. K-radar: 4d radar object detection for autonomous driving in various weather conditions[J]. Advances in Neural Information Processing Systems, 2022, 35: 3819-3829.
    
    \bibitem[22]{ntu}Zhang J, Zhuge H, Liu Y, et al. Ntu4dradlm: 4d radar-centric multi-modal dataset for localization and mapping[C]//2023 IEEE 26th International Conference on Intelligent Transportation Systems (ITSC). IEEE, 2023: 4291-4296.

    \bibitem[23]{imu}Gao W, Liu X, Zhang H. imu\_utils: A ROS-based IMU calibration toolbox[EB/OL]. 2016. GitHub repository. https://github.com/gaowenliang/imu\_utils.

    \bibitem[24]{livox}Livox Technology. livox\_camera\_lidar\_calibration: A Toolbox for LiDAR-Camera Extrinsic Calibration [EB/OL]. 2020. GitHub repository. https://github.com/Livox-SDK/livox\_camera\_lidar\_calibration.
    
    \bibitem[25]{kar}Furgale P, Rehder J, Siegwart R. Unified temporal and spatial calibration for multi-sensor systems[C]//2013 IEEE/RSJ International Conference on Intelligent Robots and Systems. IEEE, 2013: 1280-1286.
    
    \bibitem[26]{vin}Qin T, Li P, Shen S. Vins-mono: A robust and versatile monocular visual-inertial state estimator[J]. IEEE transactions on robotics, 2018, 34(4): 1004-1020.

    \bibitem[27]{lii}F. Zhu, Y. Ren and F. Zhang, "Robust Real-time LiDAR-inertial Initialization," 2022 IEEE/RSJ International Conference on Intelligent Robots and Systems (IROS), Kyoto, Japan, 2022, pp. 3948-3955, doi: 10.1109/IROS47612.2022.9982225.

    \bibitem[28]{fastlio2}Xu W, Cai Y, He D, et al. Fast-lio2: Fast direct lidar-inertial odometry[J]. IEEE Transactions on Robotics, 2022, 38(4): 2053-2073.
    
    \bibitem[29]{fastlivo2}Zheng C, Xu W, Zou Z, et al. Fast-livo2: Fast, direct lidar-inertial-visual odometry[J]. IEEE Transactions on Robotics, 2024.
   
    \bibitem[30]{ika}Chen S, Li X, Li S, et al. ikalibr: Unified targetless spatiotemporal calibration for resilient integrated inertial systems[J]. IEEE Transactions on Robotics, 2025.

    \bibitem[31]{liosam}Shan T, Englot B, Meyers D, et al. Lio-sam: Tightly-coupled lidar inertial odometry via smoothing and mapping[C]//2020 IEEE/RSJ international conference on intelligent robots and systems (IROS). IEEE, 2020: 5135-5142.

    \bibitem[32]{fastlivo}Zheng C, Zhu Q, Xu W, et al. Fast-livo: Fast and tightly-coupled sparse-direct lidar-inertial-visual odometry[C]//2022 IEEE/RSJ International Conference on Intelligent Robots and Systems (IROS). IEEE, 2022: 4003-4009.

    \bibitem[33]{r3live}Lin J, Zhang F. R 3 LIVE: A Robust, Real-time, RGB-colored, LiDAR-Inertial-Visual tightly-coupled state Estimation and mapping package[C]//2022 International Conference on Robotics and Automation (ICRA). IEEE, 2022: 10672-10678.
    
    \bibitem[34]{livoxslam}Lin J, Zhang F. Loam livox: A fast, robust, high-precision LiDAR odometry and mapping package for LiDARs of small FoV[C]//2020 IEEE international conference on robotics and automation (ICRA). IEEE, 2020: 3126-3131.

    \bibitem[35]{cticp}Dellenbach P, Deschaud J E, Jacquet B, et al. Ct-icp: Real-time elastic lidar odometry with loop closure[C]//2022 International Conference on Robotics and Automation (ICRA). IEEE, 2022: 5580-5586.    
    
    \bibitem[36]{orb}Campos C, Elvira R, Rodríguez J J G, et al. Orb-slam3: An accurate open-source library for visual, visual–inertial, and multimap slam[J]. IEEE transactions on robotics, 2021, 37(6): 1874-1890.
    
    \bibitem[37]{dso}Engel J, Koltun V, Cremers D. Direct sparse odometry[J]. IEEE transactions on pattern analysis and machine intelligence, 2017, 40(3): 611-625.
    
    \bibitem[38]{4dslam}Zhang J, Zhuge H, Wu Z, et al. 4dradarslam: A 4d imaging radar slam system for large-scale environments based on pose graph optimization[C]//2023 IEEE International Conference on Robotics and Automation (ICRA). IEEE, 2023: 8333-8340.
    
    \bibitem[39]{gs}Matsuki H, Murai R, Kelly P H J, et al. Gaussian splatting slam[C]//Proceedings of the IEEE/CVF Conference on Computer Vision and Pattern Recognition. 2024: 18039-18048.
    \end{thebibliography}
\end{document}